\pdfoutput=1
\documentclass[11pt, a4paper]{google}

\pdftrailerid{redacted}

\usepackage[authoryear, sort&compress, round]{natbib}

\usepackage{mathtools}
\usepackage{nicefrac}

\usepackage{multirow}
\usepackage{makecell}
\usepackage{hhline}
\usepackage{threeparttable}
\usepackage{array}

\usepackage{float}
\usepackage{placeins}
\usepackage{subcaption}
\usepackage{wrapfig}
\newcommand{\FinishWrap}{%
  \par
  \ifnum\value{WF@wrappedlines}>1
    \vspace{\dimexpr\value{WF@wrappedlines}\baselineskip-\baselineskip\relax}%
  \fi
  \WFclear}
\definecolor{mygray}{gray}{.88}
\definecolor{pyblue}{rgb}{0.0, 0.0, 0.5}
\definecolor{pygreen}{rgb}{0.0, 0.5, 0.0}
\definecolor{pyorange}{rgb}{1.0, 0.4, 0.0}
\definecolor{myblue}{HTML}{5f8fb7}
\definecolor{myred}{HTML}{cc7c7c}
\definecolor{mybrown}{HTML}{D8C6A6}
\definecolor{mossgreen}{HTML}{8B9A7B}
\definecolor{ourstext}{HTML}{9B2C4E}

\definecolor{appblue}{HTML}{1A73E8}
\definecolor{appdeep}{HTML}{174EA6}
\definecolor{apptint}{HTML}{F1F6FE}
\definecolor{nupurple}{HTML}{4E2A84}
\definecolor{nupurple80}{RGB}{104,76,150}
\definecolor{nulight}{HTML}{F2F0F7}
\hypersetup{colorlinks, citecolor=appdeep, linkcolor=appblue, urlcolor=appblue}

\definecolor{headcolor}{HTML}{C5D5ED}
\definecolor{catcolor}{HTML}{E7F6FB}
\definecolor{firstcolor}{HTML}{F2C8D9}
\definecolor{secondcolor}{HTML}{FDE9F3}
\newcommand{\hlfirst}[1]{\cellcolor{firstcolor}\textbf{#1}}
\newcommand{\hlsecond}[1]{\cellcolor{secondcolor}#1}

\newcommand{\mname}[1]{\mbox{\texttt{#1}}}
\newcommand{\hgroup}[1]{\textbf{#1}}
\newcommand{\hcol}[1]{#1}
\newcommand{\mdl}[2]{\makecell[c]{\llmname{#1}\\[2pt]{\scriptsize\colorbox{gray!70}{\textcolor{white}{#2}}}}}

\usepackage{algorithm}
\usepackage{algpseudocode}
\usepackage{needspace}
\usepackage{listings}
\usepackage{alltt}

\lstdefinestyle{pythonstyle}{
    language=Python,
    basicstyle=\footnotesize\ttfamily,
    breaklines=true,
    morekeywords={self},
    keywordstyle=\color{pyblue},
    commentstyle=\color{pygreen},
    stringstyle=\color{pyorange},
    numberstyle=\tiny\color{gray},
    numbers=left,
    numbersep=10pt,
    tabsize=4,
    showspaces=false,
    showstringspaces=false
}

\usepackage[skins,breakable,most]{tcolorbox}
\tcbuselibrary{listings}
\usepackage{seqsplit}

\newtcblisting{promptbox}[2]{%
  breakable,
  enhanced jigsaw,
  listing only,
  colback=#2!4!white,
  colframe=#2!50!white,
  coltitle=#2!60!black,
  colbacktitle=#2!18!white,
  fonttitle=\bfseries\footnotesize,
  title={#1},
  boxrule=0.6pt,
  arc=2pt,
  left=4pt,right=4pt,top=3pt,bottom=3pt,
  listing options={
    basicstyle=\ttfamily\scriptsize,
    breaklines=true,
    breakatwhitespace=true,
    breakautoindent=false,
    breakindent=0pt,
    columns=fullflexible,
    keepspaces=true,
    showstringspaces=false,
  },
}

\newtcblisting{skillbox}[2]{%
  breakable,
  enhanced jigsaw,
  listing only,
  colback=#2!4!white,
  colframe=#2!50!white,
  coltitle=#2!60!black,
  colbacktitle=#2!18!white,
  fonttitle=\bfseries\footnotesize,
  title={#1},
  boxrule=0.6pt,
  arc=2pt,
  left=4pt,right=4pt,top=3pt,bottom=3pt,
  listing options={
    basicstyle=\ttfamily\scriptsize,
    breaklines=true,
    breakatwhitespace=false,
    breakautoindent=false,
    breakindent=0pt,
    columns=fullflexible,
    keepspaces=true,
    upquote=false,
    literate={`}{{\char0}}1,
    showstringspaces=false,
  },
}

\newtcolorbox{defbox}{enhanced, breakable, colback=nulight, colframe=nulight, boxrule=0pt, arc=4pt,
  left=10pt, right=10pt, top=7pt, bottom=7pt, before skip=10pt, after skip=10pt}
\newcounter{definition}
\newenvironment{definition}[1][]{\refstepcounter{definition}\begin{defbox}\noindent
  \textbf{Definition~\thedefinition}\if\relax\detokenize{#1}\relax\else\ (#1)\fi\textbf{.}\ \ignorespaces}
  {\end{defbox}}
\newcommand{\defterm}[1]{\textbf{\textcolor{nupurple}{#1}}}

\usepackage{fontawesome5}

\usepackage{tikz}

\usetikzlibrary{decorations.pathmorphing, arrows.meta, calc}
\definecolor{cvblue}{HTML}{A9C0DE}\definecolor{cvrose}{HTML}{E8A9C0}\definecolor{cvlav}{HTML}{C8B6E2}
\definecolor{cvbluedk}{HTML}{4F6F9E}\definecolor{cvrosedk}{HTML}{A8506F}\definecolor{cvlavdk}{HTML}{6F5A9C}
\definecolor{cvsand}{HTML}{EBD7A8}\definecolor{cvsanddk}{HTML}{8F7232}\definecolor{cvmint}{HTML}{C6E6DD}
\definecolor{cvmintdk}{HTML}{2E7D6B}\definecolor{cvimp}{HTML}{D8D2EC}\definecolor{cvimpdk}{HTML}{5B4A9B}
\definecolor{cvbulb}{HTML}{FFC400}\definecolor{figred}{HTML}{FF646F}\definecolor{figgreen}{HTML}{28BC78}

\newcounter{cvseed}
\newcommand{\cvseedset}{\pgfmathsetseed{\number\numexpr\value{cvseed}*97+1842\relax}}
\newcommand{\cvhandrect}[3]{let \p1 = #1, \p2 = #2 in
  (\x1 + #3, \y1) decorate{-- (\x2 - #3, \y1)} arc[start angle=-90, end angle=0, radius=#3]
  decorate{-- (\x2, \y2 - #3)} arc[start angle=0, end angle=90, radius=#3]
  decorate{-- (\x1 + #3, \y2)} arc[start angle=90, end angle=180, radius=#3]
  decorate{-- (\x1, \y1 + #3)} arc[start angle=180, end angle=270, radius=#3]}
\newcommand{\cvhandpath}[2]{\cvhandrect{([shift={(#1,#1)}]frame.south west)}{([shift={(-#1,-#1)}]frame.north east)}{#2}}
\tikzset{cvouter/.style={decoration={random steps, segment length=4.5pt, amplitude=0.35pt}, rounded corners=0.7pt},
  cvmid/.style={decoration={random steps, segment length=4.5pt, amplitude=0.3pt}, rounded corners=0.6pt},
  cvinner/.style={decoration={random steps, segment length=4.5pt, amplitude=0.22pt}, rounded corners=0.5pt},
  cvjit/.style={decorate, decoration={random steps, segment length=4.5pt, amplitude=0.22pt}, rounded corners=0.8pt},
  cvx/.pic={\fill[figred] (0,0) circle (3.3pt); \draw[black!85, line width=0.45pt] plot[smooth cycle, tension=0.9] coordinates {(3.35pt,0.15pt) (0.12pt,3.3pt) (-3.3pt,0.05pt) (-0.08pt,-3.35pt)};
    \draw[white, line width=0.7pt, line cap=round] (-1.3pt,-1.3pt) -- (1.3pt,1.3pt) (-1.3pt,1.3pt) -- (1.3pt,-1.3pt);},
  cvok/.pic={\fill[figgreen] (0,0) circle (3.3pt); \draw[black!85, line width=0.45pt] plot[smooth cycle, tension=0.9] coordinates {(3.35pt,0.15pt) (0.12pt,3.3pt) (-3.3pt,0.05pt) (-0.08pt,-3.35pt)};
    \draw[white, line width=0.7pt, line cap=round, line join=round] (-1.5pt,0.05pt) -- (-0.45pt,-1.05pt) -- (1.6pt,1.1pt);}}
\newtcolorbox{cvpanel}[5]{enhanced, colback=#1!22, colframe=#2, boxrule=0.8pt, arc=4pt, outer arc=4pt, colbacktitle=#1!85,
  coltitle=black, fonttitle=\small\bfseries, halign title=center, toptitle=2pt, bottomtitle=2pt, left=4pt, right=4pt,
  top=4pt, bottom=4pt, title={\raisebox{0pt}[2ex][0.6ex]{#3}}, width=#4, nobeforeafter, box align=center,
  equal height group=cvp#5, code={\stepcounter{cvseed}},
  frame code={\cvseedset\begin{scope}\path[clip, cvouter] \cvhandpath{0.3pt}{4pt};
    \fill[#1!22] (frame.south west) rectangle (frame.north east);
    \fill[#1!85] (title.south west) rectangle (frame.north east);\end{scope}},
  interior code={}, title code={},
  overlay={\draw[#2, line width=0.6pt, cvjit] ([xshift=0.6pt]title.south west) -- ([xshift=-0.6pt]title.south east);
    \cvseedset\draw[#2, line width=0.8pt, cvouter, line cap=round, line join=round] \cvhandpath{0.3pt}{4pt};}}
\newtcolorbox{cvtask}{enhanced, colback=cvsand!28, colframe=cvsanddk, boxrule=0.8pt, arc=4pt, outer arc=4pt, left=6pt,
  right=6pt, top=3pt, bottom=3pt, code={\stepcounter{cvseed}},
  frame code={\cvseedset\begin{scope}\path[clip, cvouter] \cvhandpath{0.3pt}{4pt};
    \fill[cvsand!28] (frame.south west) rectangle (frame.north east);\end{scope}}, interior code={},
  overlay={\cvseedset\draw[cvsanddk, line width=0.8pt, cvouter, line cap=round, line join=round] \cvhandpath{0.3pt}{4pt};}}
\newtcolorbox{cvcard}[1]{enhanced, colback=white, colframe=black!75, boxrule=0.6pt, arc=5pt, left=4pt, right=4pt,
  top=3pt, bottom=4pt, equal height group=cvc#1, code={\stepcounter{cvseed}},
  frame code={\cvseedset\begin{scope}\path[clip, cvmid] \cvhandpath{0.3pt}{5pt};
    \fill[white] (frame.south west) rectangle (frame.north east);\end{scope}}, interior code={},
  overlay={\cvseedset\draw[black!75, line width=0.6pt, cvmid, line cap=round, line join=round] \cvhandpath{0.3pt}{5pt};}}

\input{figures/rendered/dims.tex}

\makeatletter
\newcommand{\nonfloattable}{\renewenvironment{table}[1][]{\def\@captype{table}}{}}
\makeatother

\usepackage{cleveref}

\usepackage{xspace}
\usepackage{fancyvrb}
\usepackage{fvextra}
\usepackage{titletoc}
\usepackage{CJKutf8}
\usepackage{lscape}

\titlecontents{section}
  [0em]
  {\addvspace{5pt}\small\bfseries\color{nupurple}}
  {\makebox[1.5em][l]{\thecontentslabel}}
  {}
  {\hspace{0.4em}\titlerule*[0.6pc]{\color{nupurple!30}.}\hspace{0.4em}%
   \contentspage}
\titlecontents{subsection}
  [1.9em]
  {\addvspace{1.5pt}\small\color{nupurple80}}
  {\makebox[2.2em][l]{\thecontentslabel}}
  {}
  {\hspace{0.4em}\titlerule*[0.75pc]{\color{nupurple!22}.}\hspace{0.4em}%
   \contentspage}
\newcommand\DoToC{%
  \begingroup
  \hypersetup{hidelinks}%
  \begin{tcolorbox}[
    enhanced, breakable,
    colback=nulight, colframe=nulight,
    boxrule=0pt, arc=3pt, boxsep=0pt,
    left=15pt, right=15pt, top=12pt, bottom=13pt,
  ]
  {\color{nupurple}\bfseries\footnotesize\sffamily
   \sodef\tocls{}{0.09em}{0.5em}{0.5em}%
   \tocls{\MakeUppercase{Contents of the Appendix}}}%
  \par\vspace{3pt}
  {\color{nupurple!35}\hrule height 0.6pt}%
  \vspace{7pt}
  \startcontents
  \printcontents{}{1}{\setcounter{tocdepth}{1}}
  \end{tcolorbox}
  \endgroup
}

\newcommand{\cmark}{\color{ForestGreen}\ding{51}}
\newcommand{\xmark}{\color{Red}\ding{55}}

\newcommand{\method}{{\fontfamily{lmtt}\selectfont \textcolor{ourstext}{\textbf{MedicalRSI}}}\xspace}
\newcommand{\improver}{{\fontfamily{lmtt}\selectfont\textcolor{impcolor}{\textbf{Improver}}}\xspace}
\newcommand{\solver}{{\fontfamily{lmtt}\selectfont\textcolor{solcolor}{\textbf{Solver}}}\xspace}
\definecolor{impcolor}{HTML}{5E3A9E}
\definecolor{solcolor}{HTML}{0E7C7B}
\definecolor{learncolor}{HTML}{3C84CC}
\newcommand{\Imp}{\mathcal{I}}
\newcommand{\Sol}{\mathcal{S}}
\newcommand{\llmname}[1]{{\fontfamily{lmtt}\selectfont #1}}

\newcommand{\challenge}[1]{\hyperref[ch:#1]{\textcolor{nupurple}{Challenge~#1}}}

\newcommand{\titlesize}{}
\newcommand{\titlebreak}{ }
\newcommand{\titlelogo}{}
\newcommand{\titlefirstline}[1]{#1}
\title{\titlesize\titlefirstline{\titlelogo\method{}: Medical Recursive Self-Improving Agents}\titlebreak via Clinical Model-Harness Co-Evolution}

\renewcommand\Affilfont{\normalfont\fontsize{10}{13}\selectfont}
\author[1]{Ziqing Wang}
\author[2]{Hamid Palangi}
\author[1]{Lili Zhao}
\author[1]{Kaize Ding}
\affil[1]{Northwestern University}
\affil[2]{University of Washington}

\correspondingauthor{}
\reportnumber{}

\customlogo{\includegraphics[height=14pt]{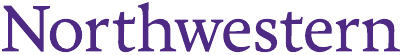}\hspace{12pt}{\color{black!30}\rule[-1pt]{0.6pt}{15pt}}\hspace{12pt}\includegraphics[height=14pt]{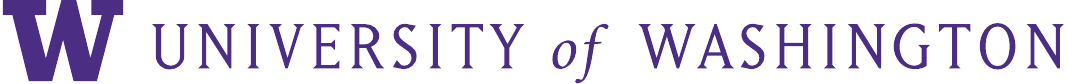}}

\newcommand{\linkpage}{https://real-lab-nu.github.io/MedicalRSI/}
\newcommand{\linkcode}{https://github.com/REAL-Lab-NU/MedicalRSI}
\newcommand{\linkmodel}{https://huggingface.co/REAL-LAB-NU/MedicalRSI-MedAgentBench-Qwen3.5-9B}
\newcommand{\hficon}{\raisebox{-0.2em}{\includegraphics[height=1.15em]{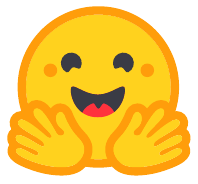}}}
\definecolor{linkink}{HTML}{2A2430}
\definecolor{pillrose}{HTML}{F9E3EC}\definecolor{pillgrey}{HTML}{ECECF1}\definecolor{pillyellow}{HTML}{FFF2C7}
\newtcbox{\linkpill}[1]{enhanced, on line, boxrule=0pt, arc=6pt, colframe=#1, colback=#1, boxsep=0pt,
  left=6pt, right=7pt, top=3pt, bottom=3pt, fontupper=\fontsize{9.5}{12}\selectfont\color{linkink}}
\newcommand{\linkrow}{%
  \href{\linkpage}{\linkpill{pillrose}{\textcolor{ourstext}{\faGlobe}~Project Page}}\hspace{7pt}%
  \href{\linkcode}{\linkpill{pillgrey}{\faGithub~Code}}\hspace{7pt}%
  \href{\linkmodel}{\linkpill{pillyellow}{\hficon~Model}}}

\usepackage{etoolbox}
\makeatletter
\patchcmd{\maketitle}{{\raggedright \@author\par}}{{\makebox[\linewidth][l]{\parbox{\textwidth}{\centering\@author}}\par}}{}{\PackageWarningNoLine{main}{Could not centre the authors}}
\makeatother

\makeatletter
\renewcommand{\abscontent}{%
  \vspace*{-9pt}{\centering\linkrow\par}\vspace{13pt}%
  \begin{tcolorbox}[enhanced, colback=nulight, colframe=nulight, boxrule=0pt, arc=4pt,
    left=12pt, right=12pt, top=6pt, bottom=8pt, before skip=0pt, after skip=0pt]
  {\centering\normalfont\bfseries\fontsize{12}{14}\selectfont\color{nupurple}Abstract\par}\vspace{3pt}
  {\normalfont\fontsize{10}{13}\selectfont\theabstract\par}
  \end{tcolorbox}}
\makeatother

\begin{abstract}
Medical agents built on large language models (LLMs) have advanced rapidly, yet most stay fixed once deployed and do not keep using their experience to improve later attempts. As a result, the same kind of error can recur across patients. To address this, existing self-improvement methods update an agent's weights or its medical harness from its own experience, but their improving strategy stays fixed and never learns which changes actually help, which caps how much the agent can improve. Recursive self-improvement (RSI) instead improves both the agent and its improving strategy. However, achieving RSI for medical agents is challenging, because clinical workflows rarely come with a reference harness, and a change that fixes early errors may not fix the problems the agent meets later. We propose \method{}, an RSI framework built on \textbf{Clinical Model-Harness Co-Evolution}, in which one model plays two roles. The \solver{} is the medical agent itself, and the \improver{} is its learnable improving strategy, which writes a medical harness from the agent's failures. Each harness is rewarded by its gain over acting without it, and both roles are trained together, so the \improver{} keeps adapting to the problems the \solver{} has not yet solved. On multiple benchmarks and LLM backbones, \method{} outperforms existing context-level and weight-level self-improvement methods and transfers well to unseen workflows, new LLMs and new databases, a step toward medical agents that keep improving.
\end{abstract}

\begin{document}

\renewcommand{\titlesize}{\fontsize{17}{21}\selectfont}
\renewcommand{\titlebreak}{\\}
\renewcommand{\titlelogo}{\texorpdfstring{\raisebox{-0.2em}{\includegraphics[height=1.1em]{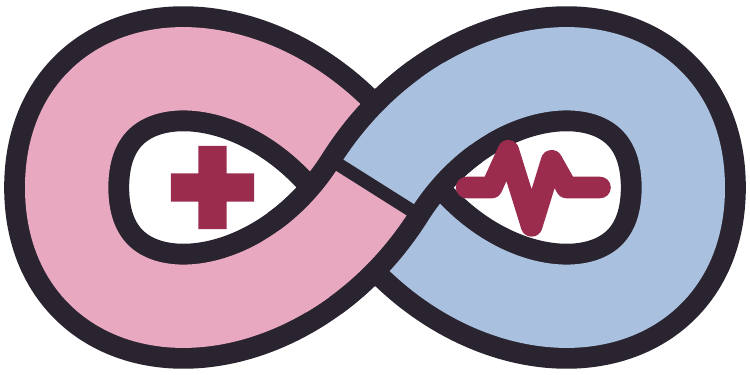}}\hspace{0.25em}}{}}
\renewcommand{\titlefirstline}[1]{\makebox[\linewidth][l]{#1}}
\maketitle
\renewcommand{\titlefirstline}[1]{#1}
\renewcommand{\titlesize}{}
\renewcommand{\titlebreak}{ }
\renewcommand{\titlelogo}{}
\vspace*{14pt}
\begin{figure}[h!]
\centering
\includegraphics[width=\textwidth]{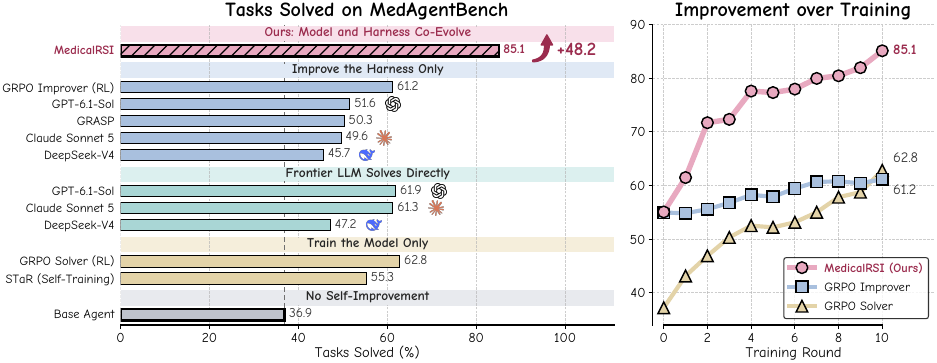}
\caption{\textbf{\method{} Outperforms Existing Methods and Keeps Improving over Training.} \emph{Left}: tasks solved on MedAgentBench with \llmname{Qwen3.5-9B}, with the methods grouped by what they change: the harness only (context-level: an RL-trained improver, GRASP, or a frontier LLM writing the harness), the model only (weight-level), or neither (a frontier LLM solving the tasks itself, or the \mname{Base Agent}, whose score the dashed line marks). \emph{Right}: \method{} keeps improving over the training rounds. It also improves faster: after two rounds, it already passes the final scores of the two strongest trained baselines.}
\label{fig:results}
\end{figure}

\section{Introduction}\label{sec:intro}

Medical agents built on large language models (LLMs) now assist clinicians in diverse settings~\citep{tang2024medagents,kim2024mdagents,wang2025amanda,wang2026latentdx,zhao2026deeprare}. On an electronic health record (EHR) system, for example, an agent can retrieve laboratory results, record patient information and place orders~\citep{jiang2025medagentbench}. In practice, clinical workflows involve recurring checks and decisions across patients, so experience gained from one attempt can guide later cases. Yet most medical agents stay fixed once deployed and do not keep using this experience to improve later attempts~\citep{feng2022clinical,fang2025comprehensive,gao2025survey,zhu2026path,zhang2026medlocomo}. As a result, the same kind of error can recur across patients~\citep{xu2026medagentgym,feng2025doctor}.

To use this experience, recent work has made medical agents improve themselves along two routes. Weight-level methods train the underlying LLM on the results of the agent's own attempts~\citep{xu2026medagentgym,feng2025doctor,chen2026cura,liu2026bat}, while context-level methods update the medical harness, including the skills, tools and memory that support clinical workflows~\citep{li2024agenthospital,shen2026evomedagent,han2026gsem,moll2026grasp,wu2026medrsi}. In both routes, an improving strategy determines how execution experience is turned into changes to the model or the harness, such as how training examples are selected or how an LLM revises the harness based on failures. However, the improving strategy in these methods stays fixed. Because this strategy never learns which of its changes actually help the agent, it caps how much the agent can improve. For example, on MedAgentBench, a 9B agent solves 51.6\% of the tasks with harnesses generated by the frontier LLM \llmname{GPT-6.1-Sol}, compared with 61.2\% when the harnesses are generated by a 9B model trained on the agent's own outcomes (Figure~\ref{fig:results}). This gap, in line with prior work~\citep{yao2024retroformer,shao2026harnessr1,ouyang2026skillos}, shows that learning how to improve can itself yield further gains. This motivates recursive self-improvement (RSI), in which \textit{an agent improves not only itself but also its improving strategy}~\citep{good1965speculations,schmidhuber2003godel,yin2025godel,zhang2026darwin}.

However, realizing RSI for medical agents is challenging. To get better, the improving strategy needs to know which of its changes actually help the agent. Clinical workflows rarely come with a reference harness that tells the agent how to organize its steps, and the details of execution also vary with each hospital's procedures and systems~\citep{finlayson2021clinician}. The improving strategy must therefore learn from the outcomes of the agent's attempts to judge whether a proposed change works. What makes this harder is that a change that fixes early errors may not fix the problems the agent meets later. For example, a magnesium replacement task requires the latest laboratory result within the last 24 hours and a replacement dose based on the magnesium level. The agent mistakenly uses a result from three days ago and so fails to place a necessary order. At this point, a skill that makes the time filter an explicit step in the harness helps the agent correct this retrieval error. Once the agent has learned to filter results correctly, however, it retrieves the correct result but selects the wrong dose, and the same change no longer helps. As a result, the improving strategy must not only learn from the agent's outcomes but also keep adapting as the agent learns, so that its changes can target the problems the agent has not yet solved. This leads to the central question of this paper:
\begin{tcolorbox}[enhanced, colback=nulight, colframe=nulight, boxrule=0pt, arc=4pt, left=10pt, right=10pt, top=6pt, bottom=6pt, before skip=5pt, after skip=8pt, halign=center]
\emph{How can a medical agent continually improve its improving strategy from its own outcomes?}
\end{tcolorbox}

To answer this question, we propose \method{}, a recursive self-improvement framework for medical agents (Figure~\ref{fig:method}). At its core is \textbf{Clinical Model-Harness Co-Evolution}, in which one model plays two roles. The \solver{} is the medical agent itself, which carries out a clinical workflow on each patient's records by following a medical harness. The \improver{} is its learnable improving strategy, which writes this harness from the agent's failures. Because no reference harness exists, each harness is rewarded by how much it improves the agent's outcomes over acting without it, so the \improver{} learns which changes actually help. Both roles are trained together on the same attempts. The \improver{} thus keeps adapting to the problems the \solver{} has not yet solved, while the \solver{} learns to carry out the workflow with the harness. On a new workflow, the same model writes a medical harness from a few failures at test time, without retraining. Our contributions are as follows:
\begin{itemize}[leftmargin=*,noitemsep,topsep=0pt]
\item \textbf{Framework.} We propose \method{}, a recursive self-improvement framework for medical agents, in which the agent improves not only itself but also its own improving strategy.
\item \textbf{Method.} We design Clinical Model-Harness Co-Evolution, which jointly trains the \improver{} to write a medical harness from failures and the \solver{} to follow it. Each harness is rewarded by its gain over acting without it, and memory keeps the best skill or tool for each workflow.
\item \textbf{Experiments.} On multiple benchmarks and LLM backbones, \method{} outperforms existing context-level and weight-level self-improvement methods (Figure~\ref{fig:results}). Extensive experiments further show its practical value, as what it learns transfers across workflows, LLMs and databases.
\end{itemize}

\section{Method}\label{sec:method}

\begin{figure*}[t!]
    \centering
    \includegraphics[width=\textwidth]{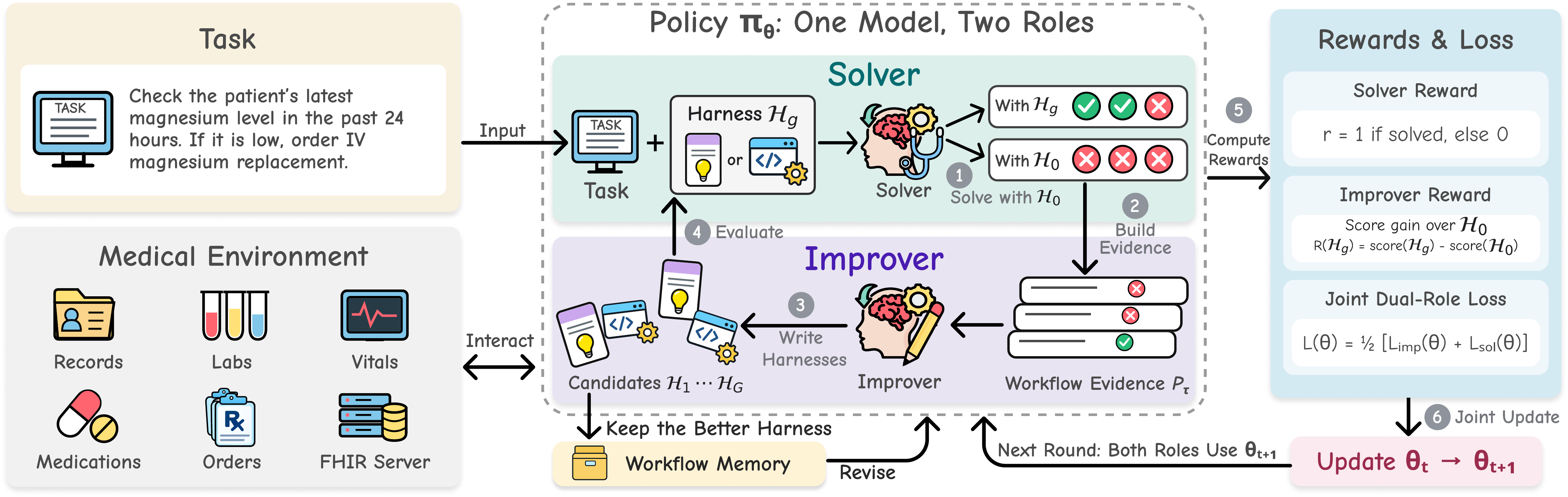}
    \vspace{-0.5cm}
\caption{\small\textbf{Overview of \method{}.}
One model plays both the \improver{} and the \solver{}. In each round, \ding{182}~the \solver{} first attempts the tasks of a workflow with $\mathcal{H}_0$, and \ding{183}~a few of its failed and successful attempts form the workflow evidence $P_\tau$. \ding{184}~The \improver{} reads $P_\tau$ with its outcome feedback and writes $G$ candidate harnesses $\mathcal{H}_1, \dots, \mathcal{H}_G$ without seeing reference answers. \ding{185}~The \solver{} then attempts the tasks with each of them. \ding{186}~The \solver{} is rewarded for solving a task, and each harness by how much it raises the \solver{}'s score over $\mathcal{H}_0$, as in Eq.~\eqref{eq:gain}. \ding{187}~One loss over both roles updates the shared weights, as in Eq.~\eqref{eq:loss}, and both roles use the new weights in the next round. Memory keeps the better harness for each workflow, and the \improver{} revises it in later rounds.}
\vspace{-0.3cm}
    \label{fig:method}
\end{figure*}

This section presents \method{} (Figure~\ref{fig:method}). Section~\ref{sec:prelim} defines clinical workflows, the medical agent and its harness, and recursive self-improvement. Section~\ref{sec:motivation} uses a preliminary experiment to motivate improving the model and its harness together. \method{} achieves this through \textbf{Clinical Model-Harness Co-Evolution}, in which one model plays both the \improver{} and the \solver{}. Section~\ref{sec:roles} describes how the \improver{} writes medical harnesses from failures without a reference harness. Section~\ref{sec:coevo} presents the gain reward, which lets the \improver{} learn which changes actually help, and the joint training of both roles, which lets the \improver{} keep adapting to the problems the \solver{} has not yet solved. Section~\ref{sec:test-time} explains how the trained model adapts to new clinical workflows at test time, without retraining.

\subsection{Preliminaries}\label{sec:prelim}

\paragraph{Clinical Tasks and Workflows.}
A medical agent acts in a clinical environment, here an EHR system that it queries and updates, for example through FHIR requests or SQL queries~\citep{mandel2016smart,jiang2025medagentbench,lee2022ehrsql,shi2024ehragent}. A task $x$ is a request about one patient, together with its context, such as the patient's identifier and the current time. A \emph{clinical workflow} $\tau$ groups the tasks that follow the same steps for different patients, for example checking a patient's latest magnesium level and ordering a replacement if it is low~\citep{jiang2025medagentbench,ryu2024ehrseqsql}. The steps are shared, but each patient's records differ, so the agent has to retrieve them and decide for each patient whether an action is due.

\paragraph{Medical Agent and Harness.}
The medical agent is a model $\pi_\theta$ that acts through a medical harness~$\mathcal{H}$, the runtime that decides what the model sees and which actions it can take, executes them in the EHR system and ends the attempt~\citep{weng2026harness,fan2026helix,shao2026harnessr1,wang2026medicalharness,wang2026openclaw,lee2026metaharness,huang2026memoharness,ouyang2026reasoningbank}. Most of its components, such as the action interface and the execution loop, stay fixed. The three that evolve, described in Section~\ref{sec:roles}, are the skills, tools and memory that guide the agent through each workflow~\citep{wang2024voyager,wang2026skillx,shi2024ehragent,wang2025awm}. The initial harness $\mathcal{H}_0$ contains no workflow-specific skill or tool, and we call the model acting with $\mathcal{H}_0$ the \solver{} alone. With a harness $\mathcal{H}$, each attempt of the agent at a task $x$ produces a trajectory ${\xi \sim \pi_\theta(\cdot \mid x, \mathcal{H})}$ of actions and environment replies. A grader checks it against the reference answer of the task, which the agent never sees~\citep{jiang2025medagentbench,lee2022ehrsql}. The grader returns an outcome $r(\xi) \in \{0, 1\}$, which marks a solved task, and a process score $\mu(\xi) = r(\xi) + \big(1 - r(\xi)\big)\, c(\xi)/2$, which gives partial credit by the share $c(\xi)$ of checks the attempt passes. The performance of the agent is the share of clinical tasks it solves,
\begin{equation}
J(\theta, \mathcal{H}) = \mathbb{E}_{x \sim \mathcal{X}}\, \mathbb{E}_{\xi \sim \pi_\theta(\cdot \mid x, \mathcal{H})} \big[r(\xi)\big].
\label{eq:perf}
\end{equation}
The agent can thus improve by training the model, which changes $\theta$, or by improving its medical harness, which changes $\mathcal{H}$.

\paragraph{Recursive Self-Improvement.}
Self-improvement proceeds in rounds. In round $t$, the agent keeps each attempt as experience $\mathcal{D}_t = \{(x_i, \xi_i, \mu_i)\}$, with its task, trajectory and process score. An improving strategy $\mathcal{U}_\psi$ turns this experience into the agent of the next round. Within this strategy, the improver is the component that proposes changes to the model or the harness, with parameters $\psi$.

\begin{definition}[Recursive Self-Improvement]\label{def:rsi}
An agent \defterm{improves itself recursively} when it uses its experience to improve both itself and its improving strategy:
\begin{equation}
(\theta_{t+1}, \mathcal{H}_{t+1}, \psi_{t+1}) = \mathcal{U}_{\psi_t}(\theta_t, \mathcal{H}_t, \psi_t, \mathcal{D}_t).
\label{eq:rsi}
\end{equation}
\end{definition}

The improving strategy aims at a higher $J(\theta_{t+1}, \mathcal{H}_{t+1})$. Under Definition~\ref{def:rsi}, the strategy also changes itself, so what the agent learns in one round shapes how its later improvements are generated, evaluated and retained~\citep{good1965speculations,duan2026last}. Fixing the improving strategy, $\psi_t = \psi_0$, reduces Eq.~\eqref{eq:rsi} to ordinary self-improvement, $(\theta_{t+1}, \mathcal{H}_{t+1}) = \mathcal{U}_{\psi_0}(\theta_t, \mathcal{H}_t, \mathcal{D}_t)$. Existing self-improvement methods for medical agents take this form. Context-level methods update the harness with a fixed LLM, writing experience into skills, tools or memories~\citep{shen2026evomedagent,moll2026grasp}, and weight-level methods update $\theta$ with a fixed training procedure~\citep{xu2026medagentgym,feng2025doctor}. In \method{}, the agent's own model plays the \improver{}, so training the agent also trains its improving strategy (Section~\ref{sec:roles}).

\subsection{Motivation: Learning from Failures in Clinical Workflows}\label{sec:motivation}

\begin{figure}[t]
\centering
\includegraphics[width=\textwidth]{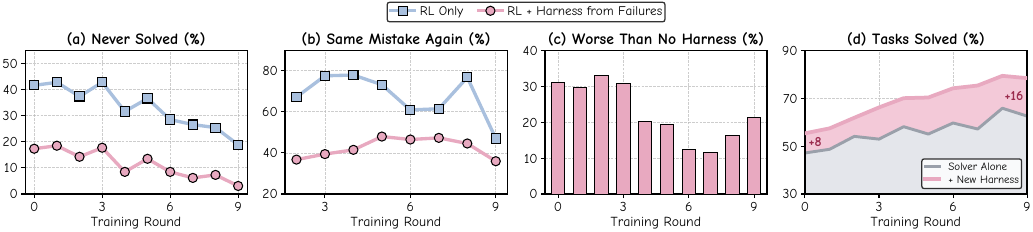}
\caption{\textbf{Learning from Failures in Clinical Workflows.} Training tasks of MedAgentBench in each round. \emph{RL only} represents training the model without a harness, and \emph{RL + harness from failures} represents also training it on attempts guided by harnesses written from its own failures. (a) Failures become a learning signal, as harnesses written from them leave far fewer tasks unsolved in all eight attempts. (b) The lessons from failures enter the model itself, which repeats the same mistake less often when acting without a harness. (c) Some harnesses make the agent worse than acting without one, but they become fewer in later rounds. (d) The harness written in each round adds more as the model improves, from 8 points over the \solver{} alone in the first round to 16 in the last.}
\label{fig:motivation}
\end{figure}

Existing self-improvement methods for medical agents improve either the model or its harness. Reinforcement learning, such as Group Relative Policy Optimization (GRPO), is a common way to improve the model: the model attempts the same task several times and learns from the attempts that score better~\citep{shao2024deepseekmath}. However, when all attempts at a task fail, the task gives no training signal. In a preliminary experiment on MedAgentBench with \llmname{Qwen3.5-9B}, we test whether a harness written from failures fills this gap (Figure~\ref{fig:motivation}). One setting trains the model with RL only, and the other also attempts each task with a harness that the agent writes from its own failures in that round. With RL only, 42\% of the training tasks are never solved in eight attempts in the first round, and 19\% still are in the last round (Figure~\ref{fig:motivation}a). With the harness, at most 19\% are in any round, and only 3\% in the last. A harness written from failures thus turns tasks the agent never solves into tasks it can learn from. To see whether the lessons from failures enter the model itself, we compare both models without a harness on the tasks they failed before. Averaged over rounds, the model trained with RL only repeats the same mistake on 68\% of these tasks, while the model trained with harnesses from its failures does so on 42\% (Figure~\ref{fig:motivation}b).

However, not every harness helps. On average, 29\% of the harnesses the agent writes in the first five rounds make it worse than acting without one. As the model improves, such harnesses become fewer, falling to 16\% in the last five rounds (Figure~\ref{fig:motivation}c). Whether a harness helps is thus known only after the agent carries it out. Meanwhile, the harness written in each round adds more over the \solver{} alone, from 8 points in the first round to 16 in the last, even as the \solver{} alone improves (Figure~\ref{fig:motivation}d). The harnesses help the model learn from its failures, and the model in turn writes more useful harnesses as training proceeds. As the model learns, a harness frozen after the first round falls behind (Figure~\ref{fig:curves}b, c). \textbf{The model and its harness should therefore learn from the latest failures together, in every round}.

\subsection{Failure-Grounded Medical Harness}\label{sec:roles}

\begin{figure*}[t]
\centering
\includegraphics[width=\textwidth]{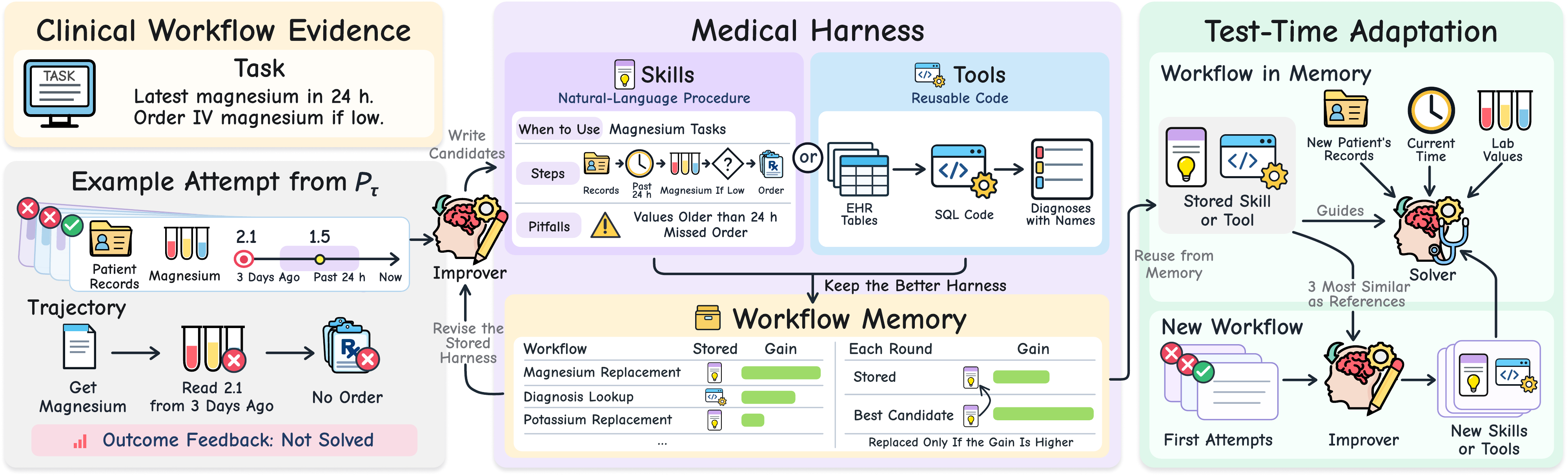}
\caption{\textbf{Details of the Failure-Grounded Medical Harness.} \emph{Left}: the clinical workflow evidence shows how the agent fails, as in this illustrative attempt where the \solver{} reads a magnesium value from three days earlier and places no order. \emph{Middle}: from such evidence, the \improver{} writes a medical harness that adds a skill or a tool, and memory keeps one per workflow, replacing it only when the best new candidate gains more. \emph{Right}: at test time, the \solver{} follows the stored skill or tool on each new patient. For a workflow without a stored skill or tool, the \improver{} writes new candidates from the first attempts, using those stored for the three most similar workflows as references.}
\label{fig:harness}
\end{figure*}

Motivated by these findings, \method{} lets the agent write its harness from its own failures in every round. Clinical workflows rarely come with a reference harness, but the latest failures show which steps the current model gets wrong, and thus what its harness should address. The paragraphs below describe what the \improver{} reads, how one model plays both roles and what each harness adds (Figure~\ref{fig:harness}).

\paragraph{Clinical Workflow Evidence.} The workflow evidence $P_\tau$ is what the \improver{} reads to write a harness for workflow $\tau$. It is drawn from the agent's latest attempts on the tasks of $\tau$, keeping failed attempts that cover different kinds of mistakes and successful ones for contrast. Together, these attempts show where the agent goes wrong in the steps that these tasks share. For each attempt, the evidence shows (1) the task instruction and its context, (2) the trajectory of actions and environment replies and (3) the outcome feedback, which tells whether the attempt solved the task and otherwise gives its process score. The evidence never shows the reference answer, the correct procedure or the grader's checks. For a workflow that orders an ECG and a repeat potassium test when the level is high, for example, the evidence shows values read outside the 24-hour window and orders that were never placed (Figure~\ref{fig:case}).

\paragraph{Dual-Role Policy.} In \method{}, the model that follows the harness also writes it. The model $\pi_\theta$ plays these two roles through two prompt templates $\rho$ (Figure~\ref{fig:method}):
\begin{align}
\Imp_\theta(\mathcal{H} \mid P_\tau) &= \pi_\theta\big(\mathcal{H} \mid \rho_{\textsc{imp}}(P_\tau)\big), \label{eq:imp}\\
\Sol_\theta(\xi \mid x, \mathcal{H}) &= \pi_\theta\big(\xi \mid \rho_{\textsc{sol}}(x, \mathcal{H})\big). \label{eq:sol}
\end{align}
The \improver{} $\Imp_\theta$ reads the workflow evidence $P_\tau$ and writes a harness $\mathcal{H}$ for workflow $\tau$, and the \solver{} $\Sol_\theta$ performs a task with this harness. We let one model play both roles for three reasons. (1)~A harness must fit the agent that uses it. Sharing the model lets the \improver{} draw on the same knowledge and capabilities that the \solver{} uses to carry out clinical workflows. (2)~Sharing weights lets both roles learn through the same update. (3)~Training is efficient, with no separate improver model to train or serve. Sharing the model also makes \method{} recursive in the sense of Definition~\ref{def:rsi}. Here, the improving strategy $\mathcal{U}_\psi$ is the whole round of Section~\ref{sec:coevo}, in which the \improver{} writes the harnesses, the \solver{} attempts the tasks with them, and one update trains both roles. Since the two roles share $\theta$, the parameters of the \improver{} are the agent's own weights, $\psi_t = \theta_t$ in Eq.~\eqref{eq:rsi}, so every update of the agent also updates its improving strategy.

\paragraph{Medical Harness Components.} The medical harness evolves through three components: skills, tools and memory. Each candidate harness adds one skill or one tool for its workflow.
\begin{itemize}[label={$\triangleright$},leftmargin=1.1em,labelsep=0.4em,itemsep=1pt,topsep=2pt,parsep=0pt,partopsep=0pt]
\item \textbf{Skills}: a short procedure in natural language with three parts. (i)~\textit{When to use} names the situation in the task that calls for it. (ii)~\textit{Steps} give the actions in order, such as which records to query, which time window and decision rule to apply, and which orders to place. (iii)~\textit{Pitfalls} name the mistakes seen in the evidence and how to avoid them.
\item \textbf{Tools}: reusable code for operations that the workflow repeats. In our experiments, a tool is a piece of SQL code, such as one that lists diagnoses together with their names. The environment loads the tools, and the \solver{} uses them by name instead of writing the same code anew in every task.
\item \textbf{Memory}: the skill or tool stored for each workflow. In each round, the \improver{} revises it and writes new ones, and the best candidate replaces it only if its gain is higher.
\end{itemize}
Appendix~\ref{app:training} gives the format of each component and how it is checked, and Appendix~\ref{app:skills} shows skills and tools that \method{} writes.

\subsection{Clinical Model-Harness Co-Evolution}\label{sec:coevo}

The \improver{} must learn to write useful harnesses for a \solver{} whose needs change during training. \method{} therefore trains both roles with reinforcement learning on the same attempts in every round, running the loop of Figure~\ref{fig:method} on a batch of training workflows (Algorithm~\ref{alg:selfrx}). Before RL, the model is fine-tuned on a small set of harnesses that a stronger LLM wrote from the failures of the \mname{Base Agent} and that raised its success. This warm start lets the \improver{} write well-formed harnesses from the first round, so RL learns faster than from scratch (Appendix~\ref{app:training}). RL then needs a reward that tells which harnesses help, and an update that trains both roles together.

\paragraph{Counterfactual Gain Reward.} The reward of a harness is what it adds to the \solver{} that uses it~\citep{wang2026huxley}. Since many harnesses that the agent writes make it worse than the \solver{} alone (Figure~\ref{fig:motivation}c), every round measures each candidate against $\mathcal{H}_0$. The \solver{} first attempts every task of a workflow with $\mathcal{H}_0$, at temperature $0$. A few of these attempts form the workflow evidence $P_\tau$, and all of them give the baseline. The \improver{} then writes $G$ candidate harnesses $\mathcal{H}_1, \dots, \mathcal{H}_G$, and the \solver{} attempts every task again with each of them. To evaluate a harness, each attempt on task $x$ is scored by the mean of its outcome and its process score, $u(\xi) = \tfrac{1}{2}\big[r(\xi) + \mu(\xi)\big]$, so partial progress also counts. Each candidate is rewarded with the gain it brings over $\mathcal{H}_0$,
\begin{equation}
R(\mathcal{H}) = \frac{1}{|\mathcal{K}_\tau(\mathcal{H})|} \sum_{k \in \mathcal{K}_\tau(\mathcal{H})} \frac{1}{|\tau_k(\mathcal{H})|} \sum_{x \in \tau_k(\mathcal{H})} \big[u(\xi_x^{\mathcal{H}}) - u(\xi_x^{\mathcal{H}_0})\big] - \lambda(\mathcal{H}),
\label{eq:gain}
\end{equation}
where $\xi_x^{\mathcal{H}}$ and $\xi_x^{\mathcal{H}_0}$ are the trajectories of the attempts on task $x$ with $\mathcal{H}$ and with $\mathcal{H}_0$. The case types $k \in \mathcal{K}_\tau(\mathcal{H})$ split the tasks of the workflow by what the patient needs, for example whether an order is due. The gain is averaged within each case type, and the case types are weighted equally. This gives the patients who need an order the same influence on the reward as those who need none, even when the latter are more common. The set $\tau_k(\mathcal{H})$ holds the tasks of case type $k$ on which both attempts were scored. The penalty $\lambda(\mathcal{H})$ charges a harness whose text is too long. Appendix~\ref{app:reward} details the reward. The reward credits a harness for the failures it fixes and charges it for the successes it breaks. It comes from the feedback of the grader alone and needs no reference harness.

\paragraph{Joint Dual-Role Optimization.} The attempts of the round train both roles at once. Besides its attempts with the candidates, the \solver{} attempts every task $N$ more times with $\mathcal{H}_0$. These attempts keep it learning to act on its own, which it needs on a new workflow, where its first attempts use $\mathcal{H}_0$ (Section~\ref{sec:test-time}). The \improver{} learns from the harness gain $R$, and the \solver{} from the task outcome $r$, both with group-relative advantages~\citep{shao2024deepseekmath},
\begin{equation}
\hat{A}^{\textsc{imp}}_{g} = R(\mathcal{H}_g) - \operatorname*{mean}_{g' \in \mathcal{G}(g)} R(\mathcal{H}_{g'}), \qquad
\hat{A}^{\textsc{sol}}_{x,j} = r_{x,j} - \operatorname*{mean}_{j'} r_{x,j'}.
\label{eq:adv}
\end{equation}
Each candidate $\mathcal{H}_g$ is compared with the candidates $\mathcal{G}(g)$ of its workflow written from the same prompt. Each attempt $j$ of the \solver{} on task $x$ is compared with the other attempts on the same task, with a harness and with $\mathcal{H}_0$. A task that the \solver{} never solves alone still yields a gradient once a candidate solves it, which is how the harness in Figure~\ref{fig:motivation} turns failures into a learning signal. One update trains both roles with one loss,
\begin{equation}
\mathcal{L}(\theta) = \tfrac{1}{2}\Big[\mathcal{L}_{\mathrm{clip}}\big(\mathcal{D}^{\textsc{imp}}\big) + \mathcal{L}_{\mathrm{clip}}\big(\mathcal{D}^{\textsc{sol}}\big)\Big],
\label{eq:loss}
\end{equation}
where $\mathcal{D}^{\textsc{imp}}$ and $\mathcal{D}^{\textsc{sol}}$ are the candidates and the \solver{}'s attempts in the round, and $\mathcal{L}_{\mathrm{clip}}$ is the clipped policy-gradient loss of GRPO with asymmetric clipping~\citep{shao2024deepseekmath,yu2025dapo}, averaged over each role's tokens so that long harnesses and short actions count equally. Because the \improver{} is trained on the failures of the current \solver{} in every round, it keeps adapting to the problems the \solver{} has not yet solved. The updated \improver{} writes the harnesses of the next round. Memory evolves with the model as well: in every round, the current \solver{} re-checks the stored skill or tool, and a candidate replaces it only if it gains more (Appendix~\ref{app:training}).

\subsection{Test-Time Adaptation to New Clinical Workflows}\label{sec:test-time}

After training, the weights $\theta$ stay fixed, and the agent adapts to each workflow only through its medical harness (Figure~\ref{fig:harness}, right). On a workflow for which memory stores a skill or tool, the \solver{} follows it. On any other workflow $\tau'$, such as one that arises after deployment, the agent writes a harness from its own failures, as in training. The \solver{} first attempts a batch of the tasks of $\tau'$ with $\mathcal{H}_0$, and the \improver{} reads these attempts with their outcome feedback, but never the reference answers. The \improver{} then writes $K$ candidate harnesses for $\tau'$, using as references the skills or tools that memory stores for the three most similar training workflows. The \solver{} follows each candidate on the same tasks, as in the protocol of Harness-R1~\citep{shao2026harnessr1}. Appendix~\ref{app:analyses} evaluates other settings, such as candidates written from other tasks or used without a grader. For evaluation, the score averages over the $K$ candidates (Section~\ref{sec:exp-setup}). The trained model can therefore write harnesses for clinical workflows it has never seen, from its own attempts and without retraining.

\section{Experiments}\label{sec:experiments}

\providecommand{\frozen}{\textcolor{myblue}{\faSnowflake}}
\providecommand{\trained}{\textcolor{pyorange}{\faFire}}
\ifcsname NC@find@Y\endcsname\else\newcolumntype{Y}{>{\centering\arraybackslash}X}\fi
\begin{table*}[t!]
\centering
\footnotesize
\setlength{\tabcolsep}{1.8pt}
\renewcommand{\arraystretch}{1}
\caption{\textbf{Main Results on MedAgentBench.} \method{} is best in every column on both backbones. Outcome is the share of tasks solved, and Process also gives partial credit for passed grader checks, both in \% with bootstrap standard deviations. Out-of-domain workflows are held out from training. \frozen{} frozen, \trained{} trained. \hlfirst{Best} and \hlsecond{second best} per column for each backbone. Baselines are described in Appendix~\ref{app:setup}.}
\label{tab:main}
\vspace{-3pt}
\begin{tabularx}{\textwidth}{cl|cc|YY|YY!{\vrule width 0.8pt}YY}
\Xhline{1.2pt}
\rowcolor{headcolor}
 & & & & \multicolumn{2}{c|}{\hgroup{In-Domain}} & \multicolumn{2}{c!{\vrule width 0.8pt}}{\hgroup{Out-of-Domain}} & \multicolumn{2}{c}{\hgroup{Overall}} \\
\rowcolor{headcolor}
\hgroup{Model} & \quad\hgroup{Method} & \hcol{Improver} & \hcol{Solver} & \hcol{Process} & \hcol{Outcome} & \hcol{Process} & \hcol{Outcome} & \hcol{Process} & \hcol{Outcome} \\
\Xhline{1pt}
\multirow{18}{*}{\mdl{Qwen3.5}{9B}}
 & \multicolumn{9}{c}{\cellcolor{catcolor}\textit{No Self-Improvement}} \\
  & \quad \mname{Base Agent} & -- & \frozen & 57.6\textsubscript{\tiny$\pm$2.2} & 40.8\textsubscript{\tiny$\pm$3.0} & 41.2\textsubscript{\tiny$\pm$4.0} & 20.0\textsubscript{\tiny$\pm$5.0} & 54.5\textsubscript{\tiny$\pm$1.9} & 36.9\textsubscript{\tiny$\pm$2.8} \\
 & \multicolumn{9}{c}{\cellcolor{catcolor}\textit{Context-Level Self-Improvement}} \\
  & \quad \mname{Reflexion} & \frozen & \frozen & 65.4\textsubscript{\tiny$\pm$2.2} & 51.2\textsubscript{\tiny$\pm$3.0} & 51.2\textsubscript{\tiny$\pm$4.5} & 31.7\textsubscript{\tiny$\pm$6.1} & 62.7\textsubscript{\tiny$\pm$1.9} & 47.5\textsubscript{\tiny$\pm$2.9} \\
  & \quad \mname{ExpeL} & \frozen & \frozen & 55.6\textsubscript{\tiny$\pm$2.1} & 36.5\textsubscript{\tiny$\pm$3.0} & 38.9\textsubscript{\tiny$\pm$3.1} & 13.3\textsubscript{\tiny$\pm$4.5} & 52.5\textsubscript{\tiny$\pm$1.9} & 32.2\textsubscript{\tiny$\pm$2.6} \\
  & \quad \mname{GEPA} & \frozen & \frozen & 63.4\textsubscript{\tiny$\pm$2.2} & 48.8\textsubscript{\tiny$\pm$3.1} & 60.0\textsubscript{\tiny$\pm$4.4} & 41.7\textsubscript{\tiny$\pm$6.5} & 62.8\textsubscript{\tiny$\pm$1.9} & 47.5\textsubscript{\tiny$\pm$2.8} \\
  & \quad \mname{EvoMedAgent} & \frozen & \frozen & 57.5\textsubscript{\tiny$\pm$2.3} & 40.4\textsubscript{\tiny$\pm$3.0} & 39.5\textsubscript{\tiny$\pm$3.1} & 13.3\textsubscript{\tiny$\pm$4.4} & 54.1\textsubscript{\tiny$\pm$1.8} & 35.3\textsubscript{\tiny$\pm$2.6} \\
  & \quad \mname{SkillX} & \frozen & \frozen & 56.0\textsubscript{\tiny$\pm$2.3} & 41.9\textsubscript{\tiny$\pm$2.9} & 47.8\textsubscript{\tiny$\pm$4.3} & 26.7\textsubscript{\tiny$\pm$5.7} & 54.4\textsubscript{\tiny$\pm$1.9} & 39.1\textsubscript{\tiny$\pm$2.7} \\
  & \quad \mname{GRASP} & \frozen & \frozen & 66.9\textsubscript{\tiny$\pm$2.3} & 54.6\textsubscript{\tiny$\pm$3.1} & 48.8\textsubscript{\tiny$\pm$4.6} & 31.7\textsubscript{\tiny$\pm$6.1} & 63.5\textsubscript{\tiny$\pm$2.0} & 50.3\textsubscript{\tiny$\pm$2.8} \\
  & \quad \mname{Fixed Improver} & \frozen & \frozen & 64.3\textsubscript{\tiny$\pm$2.0} & 49.5\textsubscript{\tiny$\pm$2.7} & 40.3\textsubscript{\tiny$\pm$3.5} & 18.3\textsubscript{\tiny$\pm$4.3} & 59.8\textsubscript{\tiny$\pm$1.7} & 43.6\textsubscript{\tiny$\pm$2.5} \\
  & \quad \mname{Fixed Improver (GPT)} & \frozen & \frozen & 67.9\textsubscript{\tiny$\pm$2.0} & 53.6\textsubscript{\tiny$\pm$2.8} & 61.9\textsubscript{\tiny$\pm$3.9} & 43.1\textsubscript{\tiny$\pm$5.6} & 66.8\textsubscript{\tiny$\pm$1.6} & 51.6\textsubscript{\tiny$\pm$2.6} \\
  & \quad \mname{SFT Improver} & \trained & \frozen & 69.9\textsubscript{\tiny$\pm$1.8} & 56.3\textsubscript{\tiny$\pm$2.6} & 64.8\textsubscript{\tiny$\pm$3.8} & 49.0\textsubscript{\tiny$\pm$5.5} & 68.9\textsubscript{\tiny$\pm$1.5} & 54.9\textsubscript{\tiny$\pm$2.4} \\
  & \quad \mname{GRPO Improver} & \trained & \frozen & 74.9\textsubscript{\tiny$\pm$1.7} & 63.1\textsubscript{\tiny$\pm$2.4} & \hlsecond{68.2\textsubscript{\tiny$\pm$3.6}} & \hlsecond{52.7\textsubscript{\tiny$\pm$5.2}} & \hlsecond{73.6\textsubscript{\tiny$\pm$1.4}} & 61.2\textsubscript{\tiny$\pm$2.2} \\
 & \multicolumn{9}{c}{\cellcolor{catcolor}\textit{Weight-Level Self-Improvement}} \\
  & \quad \mname{STaR} & -- & \trained & 71.8\textsubscript{\tiny$\pm$2.2} & 61.2\textsubscript{\tiny$\pm$3.0} & 50.4\textsubscript{\tiny$\pm$4.3} & 30.0\textsubscript{\tiny$\pm$6.0} & 67.8\textsubscript{\tiny$\pm$2.0} & 55.3\textsubscript{\tiny$\pm$2.9} \\
  & \quad \mname{GRPO Solver} & -- & \trained & \hlsecond{77.6\textsubscript{\tiny$\pm$2.1}} & \hlsecond{68.5\textsubscript{\tiny$\pm$2.9}} & 56.1\textsubscript{\tiny$\pm$4.7} & 38.3\textsubscript{\tiny$\pm$6.4} & 73.5\textsubscript{\tiny$\pm$1.9} & \hlsecond{62.8\textsubscript{\tiny$\pm$2.8}} \\
 & \multicolumn{9}{c}{\cellcolor{catcolor}\textit{Ours}} \\
  & \quad \method{} & \trained & \trained & \hlfirst{93.0\textsubscript{\tiny$\pm$1.3}} & \hlfirst{89.5\textsubscript{\tiny$\pm$1.9}} & \hlfirst{78.7\textsubscript{\tiny$\pm$3.3}} & \hlfirst{66.1\textsubscript{\tiny$\pm$5.1}} & \hlfirst{90.3\textsubscript{\tiny$\pm$1.2}} & \hlfirst{85.1\textsubscript{\tiny$\pm$1.9}} \\
\Xhline{1pt}
\multirow{18}{*}{\mdl{Gemma-4}{E4B}}
 & \multicolumn{9}{c}{\cellcolor{catcolor}\textit{No Self-Improvement}} \\
  & \quad \mname{Base Agent} & -- & \frozen & 15.7\textsubscript{\tiny$\pm$1.5} & 5.8\textsubscript{\tiny$\pm$1.5} & 23.4\textsubscript{\tiny$\pm$3.3} & 6.7\textsubscript{\tiny$\pm$3.2} & 17.1\textsubscript{\tiny$\pm$1.3} & 5.9\textsubscript{\tiny$\pm$1.4} \\
 & \multicolumn{9}{c}{\cellcolor{catcolor}\textit{Context-Level Self-Improvement}} \\
  & \quad \mname{Reflexion} & \frozen & \frozen & 52.9\textsubscript{\tiny$\pm$2.5} & 40.0\textsubscript{\tiny$\pm$3.0} & 43.1\textsubscript{\tiny$\pm$4.6} & 26.7\textsubscript{\tiny$\pm$5.9} & 51.1\textsubscript{\tiny$\pm$2.2} & 37.5\textsubscript{\tiny$\pm$2.7} \\
  & \quad \mname{ExpeL} & \frozen & \frozen & 38.9\textsubscript{\tiny$\pm$2.3} & 26.2\textsubscript{\tiny$\pm$2.8} & 33.7\textsubscript{\tiny$\pm$5.0} & 25.0\textsubscript{\tiny$\pm$5.7} & 38.0\textsubscript{\tiny$\pm$2.1} & 25.9\textsubscript{\tiny$\pm$2.5} \\
  & \quad \mname{GEPA} & \frozen & \frozen & 57.3\textsubscript{\tiny$\pm$2.3} & 40.4\textsubscript{\tiny$\pm$3.0} & 47.8\textsubscript{\tiny$\pm$4.6} & 30.0\textsubscript{\tiny$\pm$6.1} & 55.6\textsubscript{\tiny$\pm$1.9} & 38.4\textsubscript{\tiny$\pm$2.7} \\
  & \quad \mname{EvoMedAgent} & \frozen & \frozen & 23.8\textsubscript{\tiny$\pm$1.8} & 11.5\textsubscript{\tiny$\pm$2.0} & 29.0\textsubscript{\tiny$\pm$3.9} & 13.3\textsubscript{\tiny$\pm$4.4} & 24.8\textsubscript{\tiny$\pm$1.6} & 11.9\textsubscript{\tiny$\pm$1.8} \\
  & \quad \mname{SkillX} & \frozen & \frozen & 16.1\textsubscript{\tiny$\pm$1.5} & 7.7\textsubscript{\tiny$\pm$1.7} & 14.2\textsubscript{\tiny$\pm$3.0} & 6.7\textsubscript{\tiny$\pm$3.2} & 15.8\textsubscript{\tiny$\pm$1.4} & 7.5\textsubscript{\tiny$\pm$1.5} \\
  & \quad \mname{GRASP} & \frozen & \frozen & 57.4\textsubscript{\tiny$\pm$2.5} & 43.8\textsubscript{\tiny$\pm$3.0} & 47.4\textsubscript{\tiny$\pm$4.4} & 28.3\textsubscript{\tiny$\pm$6.0} & 55.5\textsubscript{\tiny$\pm$2.1} & 40.9\textsubscript{\tiny$\pm$2.8} \\
  & \quad \mname{Fixed Improver} & \frozen & \frozen & 41.5\textsubscript{\tiny$\pm$1.6} & 29.6\textsubscript{\tiny$\pm$1.8} & 38.5\textsubscript{\tiny$\pm$3.3} & 22.4\textsubscript{\tiny$\pm$4.2} & 41.0\textsubscript{\tiny$\pm$1.3} & 28.3\textsubscript{\tiny$\pm$1.8} \\
  & \quad \mname{Fixed Improver (GPT)} & \frozen & \frozen & 60.1\textsubscript{\tiny$\pm$1.9} & 44.1\textsubscript{\tiny$\pm$2.6} & \hlsecond{59.2\textsubscript{\tiny$\pm$3.9}} & \hlsecond{42.9\textsubscript{\tiny$\pm$5.5}} & 59.9\textsubscript{\tiny$\pm$1.7} & 43.9\textsubscript{\tiny$\pm$2.4} \\
  & \quad \mname{SFT Improver} & \trained & \frozen & 60.1\textsubscript{\tiny$\pm$1.9} & 43.8\textsubscript{\tiny$\pm$2.6} & 54.8\textsubscript{\tiny$\pm$3.4} & 35.3\textsubscript{\tiny$\pm$4.8} & 59.1\textsubscript{\tiny$\pm$1.6} & 42.2\textsubscript{\tiny$\pm$2.4} \\
  & \quad \mname{GRPO Improver} & \trained & \frozen & 63.0\textsubscript{\tiny$\pm$2.0} & 48.0\textsubscript{\tiny$\pm$2.7} & 58.4\textsubscript{\tiny$\pm$3.8} & 40.6\textsubscript{\tiny$\pm$5.3} & 62.1\textsubscript{\tiny$\pm$1.7} & 46.6\textsubscript{\tiny$\pm$2.5} \\
 & \multicolumn{9}{c}{\cellcolor{catcolor}\textit{Weight-Level Self-Improvement}} \\
  & \quad \mname{STaR} & -- & \trained & \hlsecond{68.1\textsubscript{\tiny$\pm$2.3}} & \hlsecond{56.2\textsubscript{\tiny$\pm$3.0}} & 51.5\textsubscript{\tiny$\pm$4.6} & 33.3\textsubscript{\tiny$\pm$6.2} & \hlsecond{65.0\textsubscript{\tiny$\pm$2.0}} & 51.9\textsubscript{\tiny$\pm$2.8} \\
  & \quad \mname{GRPO Solver} & -- & \trained & 67.6\textsubscript{\tiny$\pm$2.3} & 55.4\textsubscript{\tiny$\pm$3.1} & 54.1\textsubscript{\tiny$\pm$4.9} & 38.3\textsubscript{\tiny$\pm$6.4} & \hlsecond{65.0\textsubscript{\tiny$\pm$2.0}} & \hlsecond{52.2\textsubscript{\tiny$\pm$2.9}} \\
 & \multicolumn{9}{c}{\cellcolor{catcolor}\textit{Ours}} \\
  & \quad \method{} & \trained & \trained & \hlfirst{83.3\textsubscript{\tiny$\pm$1.7}} & \hlfirst{75.1\textsubscript{\tiny$\pm$2.7}} & \hlfirst{69.4\textsubscript{\tiny$\pm$3.6}} & \hlfirst{53.9\textsubscript{\tiny$\pm$5.2}} & \hlfirst{80.7\textsubscript{\tiny$\pm$1.6}} & \hlfirst{71.1\textsubscript{\tiny$\pm$2.4}} \\
\Xhline{1.2pt}
\end{tabularx}
\vspace{-0.3cm}
\end{table*}

Section~\ref{sec:method} lets the model and its harness evolve together. The experiments test whether this pays off on two benchmarks and three LLM backbones, asking (1)~whether \method{} improves medical agents more than context-level and weight-level self-improvement, especially on unseen workflows (Section~\ref{sec:exp-main}), (2)~whether the improver keeps improving with the agent (Section~\ref{sec:exp-coevo}), (3)~whether what it learns carries over to new solvers, improvers and databases (Section~\ref{sec:exp-transfer}), (4)~what each component adds (Section~\ref{sec:exp-ablation}), and (5)~what the harness fixes and what it costs (Section~\ref{sec:exp-case}).

\subsection{Experimental Setup}\label{sec:exp-setup}

\paragraph{Benchmarks.}
Our main benchmark is MedAgentBench~\citep{jiang2025medagentbench}, where the agent operates an EHR system through FHIR requests. We use its 10 task types, the 10 of its second version~\citep{chen2025medagentbenchv2} and 12 new ones, and treat each as one clinical workflow, following the benchmark's own division, 32 workflows in total. Twenty-six workflows are used for training, and six are held out as unseen workflows, which the tables report as out-of-domain. As a second benchmark, we use MedAgentGym~\citep{xu2026medagentgym}, specifically its EHR-SeqSQL subset~\citep{ryu2024ehrseqsql}, where the agent answers clinical questions by querying a MIMIC-III database~\citep{johnson2016mimic}. We re-split it in the same way, with each structure of the gold SQL query as one workflow, 26 for training and six held out as unseen workflows. On both benchmarks, the unseen workflows are drawn at random and differ from every training workflow. Appendix~\ref{app:data} gives the details of both benchmarks.

\paragraph{Baselines.}
We compare with three groups of methods. \ding{182}~\textbf{No self-improvement.} The \mname{Base Agent} is never trained. \ding{183}~\textbf{Context-level.} \mname{Reflexion}~\citep{shinn2023reflexion}, \mname{ExpeL}~\citep{zhao2024expel}, \mname{GEPA}~\citep{agrawal2026gepa}, \mname{EvoMedAgent}~\citep{shen2026evomedagent}, \mname{SkillX}~\citep{wang2026skillx} and \mname{GRASP}~\citep{moll2026grasp} keep the solver frozen and improve what it receives. We add four improvers that write skills from the same workflow evidence. \mname{Fixed Improver} and \mname{Fixed Improver (GPT)} use the base model and \llmname{GPT-6.1-Sol} without training. \mname{SFT Improver} and \mname{GRPO Improver} train the improver for the frozen solver. \mname{SFT Improver} is our warm start, and \mname{GRPO Improver} re-implements Harness-R1~\citep{shao2026harnessr1} for clinical workflows. \ding{184}~\textbf{Weight-level.} \mname{STaR}~\citep{zelikman2022star} and the \mname{GRPO Solver} train the solver. The \mname{GRPO Solver} uses the same multi-turn GRPO update as \method{}~\citep{wang2025ragen}. Appendix~\ref{app:setup} gives further details of the baselines.

\paragraph{Evaluation.}
We follow the evaluation protocol of Harness-R1~\citep{shao2026harnessr1}. For the improver baselines, the solver first attempts the test tasks of each workflow without a skill, and the improver writes 16 skills from these attempts and their outcome feedback, without the reference answers. The solver then attempts every test task once with each skill, and we report the mean over the 16 skills. \method{} is deployed as in Section~\ref{sec:test-time}: each workflow with a stored skill or tool uses it once, and on the other workflows we report the mean over its $K$ candidates. Each method averages over its candidates instead of keeping the best one. When every harness is written from training tasks instead of test tasks, no method changes by more than 1.2 points overall (Table~\ref{tab:train} in Appendix~\ref{app:analyses}). We report Outcome, the share of tasks solved, and Process, which also gives partial credit for passed grader checks. Trained methods are tested after ten rounds, and Appendix~\ref{app:training} gives their training details.

\subsection{Main Results}\label{sec:exp-main}

\begin{figure}[t]
\centering
\includegraphics[width=\textwidth]{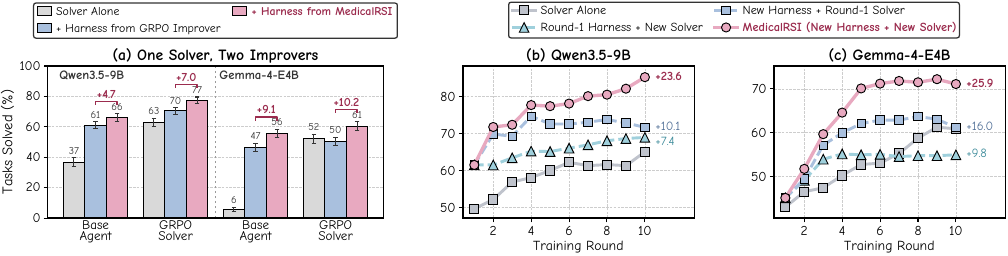}
\caption{\textbf{Analysis of Co-Evolution.} (a) With the same solver, \method{} writes a better harness than the \mname{GRPO Improver}. Each solver runs alone, with the harness of the \mname{GRPO Improver}, and with that of \method{}. Numbers above the bars give \method{} minus the \mname{GRPO Improver}. (b, c) Both roles improve, and together they improve the most. Each line starts from the first round and upgrades the harness, the solver or both to those of later rounds. Numbers give the gain from the first round.}
\label{fig:curves}
\end{figure}

Table~\ref{tab:main} reports the main results. \ding{182}~\textbf{\method{} outperforms both routes of self-improvement.} It gives the best Process and Outcome in every column on both backbones (\llmname{Qwen3.5-4B} in Appendix~\ref{app:mab4b}). On \llmname{Qwen3.5-9B}, it raises the overall Outcome of the \mname{Base Agent} from 36.9\% to 85.1\%, 22.3 points above the strongest baseline. Appendix~\ref{app:runs} reports three training runs. \ding{183}~\textbf{The advantage holds on a second benchmark.} On MedAgentGym (Table~\ref{tab:ehrseqsql} in Appendix~\ref{app:ehrseqsql}), \method{} reaches 89.0\% overall with \llmname{Qwen3.5-9B}, 19.5 points above the strongest baseline, and is best on both the training workflows and the unseen ones. \ding{184}~\textbf{The gains carry over to unseen workflows.} \method{} more than triples the Outcome of the \mname{Base Agent} on unseen workflows on both backbones, and on \llmname{Gemma-4-E4B} it leads every baseline there by 10.9 points. On fifteen additional unseen workflows with 450 tasks, \method{} solves 74.1\% against 59.8\% for the \mname{GRPO Improver} with \llmname{Qwen3.5-9B} (Table~\ref{tab:unseen15} in Appendix~\ref{app:analyses}). \ding{185}~\textbf{Each route alone falls short in its own way.} Context-level methods are held back by what they keep fixed. Even \llmname{GPT-6.1-Sol} writing the harness stays 33.5 points below \method{} with \llmname{Qwen3.5-9B} on MedAgentBench, and Section~\ref{sec:exp-coevo} examines this gap. Weight-level methods cannot adapt the agent to a new workflow at test time. On the unseen workflows, the \mname{GRPO Solver} solves 38.3\% against 66.1\% for \method{}, and Section~\ref{sec:exp-case} looks at what the harness fixes there.

\begin{figure*}[t]
\centering
\includegraphics[width=\textwidth]{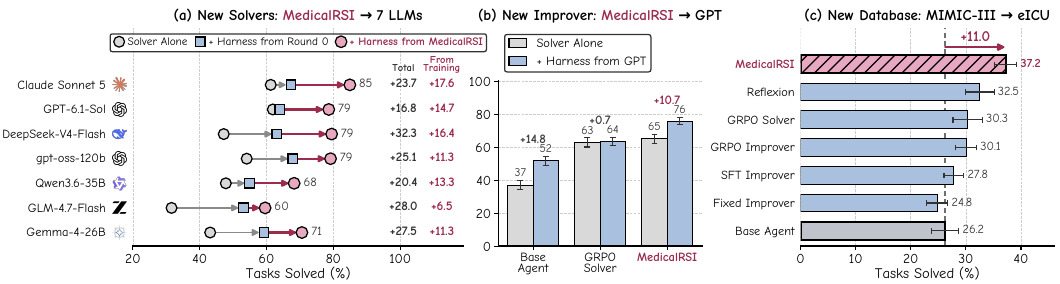}
\vspace{-19pt}
\caption{\textbf{Analysis of Transfer Beyond Training.} (a) The \improver{} of \method{} helps every solver, even frontier models, and RL training adds much of this gain. Each solver runs alone, with the harness that this \improver{} writes before RL training, and with the one it writes after. \emph{Total} gives the gain over the solver alone, and \emph{From Training} the part that RL training adds. (b) The \solver{} of \method{} makes strong use of a harness from another model. Each solver runs alone and with the harness that \llmname{GPT-6.1-Sol} writes from its failures. (c) \method{} carries over to a new database. The methods are trained on MedAgentGym over MIMIC-III and tested on eICU without further weight updates. The dashed line marks the \mname{Base Agent}, and error bars show the bootstrap standard deviation over test tasks.}
\label{fig:transfer}
\vspace{-6pt}
\end{figure*}

\subsection{Co-Evolution Analysis}\label{sec:exp-coevo}

This section asks whether the improver itself improves and whether the two roles improve each other (Figure~\ref{fig:curves}). \ding{182}~\textbf{With the same solver, the harness of \method{} is better.} We fix four solvers and let either the \mname{GRPO Improver} or the \improver{} of \method{} write their harness. \method{} leads by 4.7 to 10.2 points with all four solvers, including the \mname{Base Agent} of each backbone, which the \mname{GRPO Improver} was trained with (Figure~\ref{fig:curves}a). Its gains therefore come from a better improver and not only from a better solver. \ding{183}~\textbf{Both roles improve, and together they improve the most.} From the first round, a new harness with the round-1 \solver{} gains 10.1 points on \llmname{Qwen3.5-9B}, and the round-1 harness with a new \solver{} gains 7.4 (Figure~\ref{fig:curves}b). Upgrading both gains 23.6 points, even more than the sum of the two, since each harness is written for the \solver{} of its round. On \llmname{Gemma-4-E4B}, the two roles gain 16.0 and 9.8 points, and both together 25.9 (Figure~\ref{fig:curves}c). \ding{184}~\textbf{A frozen harness falls behind as the agent learns.} In the last round, the harness frozen after round 1 adds only 3.9 points over the \solver{} alone on \llmname{Qwen3.5-9B} and falls below it on \llmname{Gemma-4-E4B}, against 20.1 and 10.2 points for the co-evolved harness. Over the ten rounds, \method{} gains 30 and 29 points on the two backbones, against 6 and 4 for training only the improver (Figure~\ref{fig:curves-appx} in Appendix~\ref{app:analyses}).

\needspace{6\baselineskip}
\subsection{Transfer Beyond Training}\label{sec:exp-transfer}

We ask whether what \method{} learns carries over to settings it was not trained in (Figure~\ref{fig:transfer}). \ding{182}~\textbf{New solvers.} In practice, a very large model is hard to train, yet it can still lack the skills of a clinical workflow. The \improver{} of \method{}, a 9B model, therefore writes a harness from the failures of seven other models, five open-weight models from 26B to 284B parameters, \llmname{GPT-6.1-Sol} and \llmname{Claude Sonnet 5}. Every solver improves, by 16.8 to 32.3 points (Figure~\ref{fig:transfer}a). Even the frontier models gain, \llmname{Claude Sonnet 5} by 23.7 points and \llmname{GPT-6.1-Sol} by 16.8. Of these gains, 6.5 to 17.6 points come from training the \improver{}, measured against the harness it writes before RL training. A small trained \improver{} can therefore supply clinical workflow skills to these models without updating their weights. \ding{183}~\textbf{New improver.} Conversely, the \solver{} of \method{} also makes strong use of a harness that another model writes. When \llmname{GPT-6.1-Sol} replaces the \improver{} and writes the harness from each solver's own failures, the \solver{} of \method{} gains 10.7 points and reaches the highest score of the three solvers (Figure~\ref{fig:transfer}b). The \mname{GRPO Solver} starts from almost the same score without a harness, 62.8 against 65.0, but gains only 0.7. The \mname{Base Agent} gains 14.8 points from a much lower start. \ding{184}~\textbf{New database.} The methods trained on MedAgentGym over MIMIC-III are tested without further weight updates on eICU, which has records from 160 US hospitals and a different schema. Because the \improver{} writes the harness from the failures of the agent on the new database, \method{} adapts to it without retraining. It reaches 37.2\%, 11.0 points above the \mname{Base Agent} and 4.7 above the next method (Figure~\ref{fig:transfer}c).

\begin{figure*}[t]
\centering
\begin{minipage}[t]{2.985in}
\centering
\includegraphics[width=\linewidth]{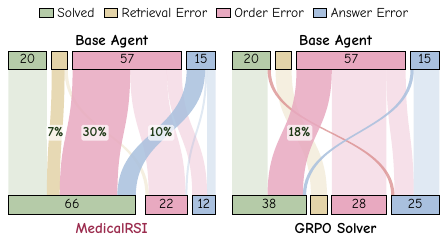}
\caption{\textbf{What the Harness Fixes on Unseen Workflows.} \method{} fixes errors of all three kinds and turns no solved task into a failure. A retrieval error means the agent does not look up the records the task needs. An order error means its orders are wrong, missing or not needed. An answer error means it reports a wrong final value.}
\label{fig:failures}
\end{minipage}\hfill
\begin{minipage}[t]{3.35in}
\centering
\includegraphics[width=\linewidth]{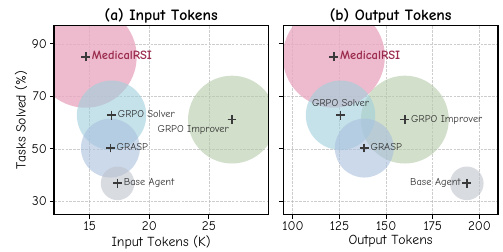}
\caption{\textbf{Cost per Attempt.} \method{} solves the most tasks while spending the fewest input and output tokens of the five methods. Tasks solved against the input (a) and output (b) tokens the solver spends per test attempt, where the upper left is better. Bubble area shows the share of tasks solved on unseen workflows, scaled within the five methods.}
\label{fig:cost}
\end{minipage}
\end{figure*}

\Needspace*{18\baselineskip}
\subsection{Ablation Study}\label{sec:exp-ablation}

\begin{wraptable}{r}{0.485\textwidth}
\vspace{-13.5pt}
\setlength{\columnwidth}{\linewidth}\nonfloattable
\begin{table}[t!]
\centering
\small
\setlength{\tabcolsep}{3pt}
\renewcommand{\arraystretch}{1.18}
\caption{\textbf{Ablations of \method{}.} \mname{w/o RL} keeps the SFT warm start, \mname{w/o SFT} runs RL from a fresh LoRA, \mname{w/o sharing} gives the two roles separate LoRAs, \mname{w/o memory} trains and tests without memory, \mname{w/o feedback} writes test-time harnesses without the outcome feedback, and \mname{w/o harness} runs the trained \solver{} alone. \hlfirst{Best} per column.}
\label{tab:ablation}
\vspace{-3pt}
\resizebox{\columnwidth}{!}{%
\begin{tabular}{l|cc|cc!{\vrule width 0.8pt}cc}
\Xhline{1.2pt}
\rowcolor{headcolor}
 & \multicolumn{2}{c|}{\hgroup{In-Domain}} & \multicolumn{2}{c!{\vrule width 0.8pt}}{\hgroup{Out-of-Domain}} & \multicolumn{2}{c}{\hgroup{Overall}} \\
\rowcolor{headcolor}
\hgroup{Variant} & \hcol{Process} & \hcol{Outcome} & \hcol{Process} & \hcol{Outcome} & \hcol{Process} & \hcol{Outcome} \\
\Xhline{1pt}
\multicolumn{7}{c}{\cellcolor{catcolor}\textit{Full System}} \\
\method{} & \hlfirst{93.0\textsubscript{\tiny$\pm$1.3}} & \hlfirst{89.5\textsubscript{\tiny$\pm$1.9}} & \hlfirst{78.7\textsubscript{\tiny$\pm$3.3}} & \hlfirst{66.1\textsubscript{\tiny$\pm$5.1}} & \hlfirst{90.3\textsubscript{\tiny$\pm$1.2}} & \hlfirst{85.1\textsubscript{\tiny$\pm$1.9}} \\
\Xhline{0.6pt}
\multicolumn{7}{c}{\cellcolor{catcolor}\textit{Training}} \\
\mname{w/o RL} & 69.5\textsubscript{\tiny$\pm$1.9} & 55.5\textsubscript{\tiny$\pm$2.7} & 69.2\textsubscript{\tiny$\pm$3.5} & 53.1\textsubscript{\tiny$\pm$5.3} & 69.4\textsubscript{\tiny$\pm$1.6} & 55.1\textsubscript{\tiny$\pm$2.4} \\
\mname{w/o SFT} & 83.4\textsubscript{\tiny$\pm$1.9} & 76.2\textsubscript{\tiny$\pm$2.6} & 57.2\textsubscript{\tiny$\pm$4.3} & 40.2\textsubscript{\tiny$\pm$5.7} & 78.5\textsubscript{\tiny$\pm$1.7} & 69.5\textsubscript{\tiny$\pm$2.6} \\
\Xhline{0.6pt}
\multicolumn{7}{c}{\cellcolor{catcolor}\textit{Architecture}} \\
\mname{w/o sharing} & 90.4\textsubscript{\tiny$\pm$1.4} & 85.1\textsubscript{\tiny$\pm$2.2} & 72.4\textsubscript{\tiny$\pm$3.9} & 58.1\textsubscript{\tiny$\pm$5.7} & 87.0\textsubscript{\tiny$\pm$1.3} & 80.0\textsubscript{\tiny$\pm$2.1} \\
\mname{w/o memory} & 87.0\textsubscript{\tiny$\pm$1.4} & 80.9\textsubscript{\tiny$\pm$2.1} & 76.0\textsubscript{\tiny$\pm$3.3} & 62.1\textsubscript{\tiny$\pm$5.1} & 85.0\textsubscript{\tiny$\pm$1.3} & 77.3\textsubscript{\tiny$\pm$2.0} \\
\Xhline{0.6pt}
\multicolumn{7}{c}{\cellcolor{catcolor}\textit{Inference}} \\
\mname{w/o feedback} & 92.9\textsubscript{\tiny$\pm$1.3} & 89.4\textsubscript{\tiny$\pm$1.9} & 76.5\textsubscript{\tiny$\pm$3.3} & 63.0\textsubscript{\tiny$\pm$5.1} & 89.8\textsubscript{\tiny$\pm$1.3} & 84.5\textsubscript{\tiny$\pm$1.9} \\
\mname{w/o harness} & 79.8\textsubscript{\tiny$\pm$1.9} & 70.0\textsubscript{\tiny$\pm$2.8} & 61.3\textsubscript{\tiny$\pm$4.5} & 43.3\textsubscript{\tiny$\pm$6.4} & 76.3\textsubscript{\tiny$\pm$1.8} & 65.0\textsubscript{\tiny$\pm$2.7} \\
\Xhline{1.2pt}
\end{tabular}}
\vspace{-12pt}
\end{table}

\end{wraptable}
\leavevmode \method{} combines RL on both roles, an SFT warm start, one model shared by the two roles, memory that keeps the best harness of each workflow, the harness at test time and the outcome feedback used to write it. To see what each contributes, Table~\ref{tab:ablation} removes one of them at a time on \llmname{Qwen3.5-9B}, and every removal lowers every score. \ding{182}~\textbf{Co-evolution brings most of the gain, and the warm start lets RL learn faster.} The warm start alone, without RL, reaches 55.1\%, 30.0 points below the full \method{} system, so most of the gain comes from training the two roles together with RL. The warm start still helps, as it lets RL learn effective harnesses faster. Starting RL from a fresh adapter instead leaves the system 15.6 points lower after the same ten rounds (Appendix~\ref{app:training}). \ding{183}~\textbf{One shared model beats two separate ones.} Giving the \improver{} and the \solver{} separate adapters loses 5.1 points overall. Sharing one set of weights thus helps, as the \improver{} draws on what the \solver{} learns (Section~\ref{sec:roles}). \ding{184}~\textbf{Memory keeps what works.} Memory keeps the best harness of each workflow across rounds. Without it, the \improver{} must write every harness from scratch in each round, and \method{} loses 7.8 points overall. \ding{185}~\textbf{The harness carries the gain at test time.} Training raises the \solver{} alone from 36.9\% to 65.0\%, and the harness adds another 20.1 points on top of it (Section~\ref{sec:exp-coevo}). \ding{186}~\textbf{\method{} stays strong without test-time outcome feedback.} At test time, the \improver{} reads the outcome feedback of the first attempts on a workflow. Without this feedback, it writes harnesses from the trajectories alone, and \method{} still reaches 84.5\% overall, above every \llmname{Qwen3.5-9B} baseline in Table~\ref{tab:main}, and 63.0\% on unseen workflows. Table~\ref{tab:nofeedback} in Appendix~\ref{app:analyses} removes this feedback from every method, and \method{} loses only 0.6 points, the least among the methods that use this feedback, while the others lose 1.6 to 10.0 points.

\subsection{What the Harness Fixes and What It Costs}\label{sec:exp-case}

\begin{figure*}[t]
\centering
\prerendered{case_main}{\includegraphics{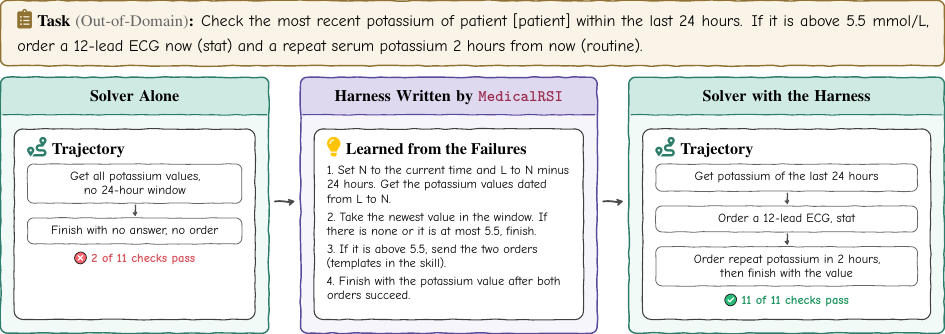}}
\vspace{-8pt}
\caption{\textbf{Case Study on an Unseen Workflow.} Without a harness, the \solver{} misses the 24-hour window and places neither order. From these failures, the same model writes a harness that fixes the window, the decision rule and both orders, and the \solver{} then passes every check. A few failures are thus enough to fix a workflow the model never saw in training. Full trajectories are in Figure~\ref{fig:case-v3-task8}, and more cases in Appendix~\ref{app:cases}.}
\label{fig:case}
\vspace{-10pt}
\end{figure*}

On MedAgentBench with \llmname{Qwen3.5-9B}, we look at what the harness fixes on unseen workflows and what each attempt costs. \ding{182}~\textbf{The harness fixes failures of every kind on unseen workflows.} Figure~\ref{fig:failures} follows each of the 60 test tasks of the unseen workflows from where the \mname{Base Agent} ends to where each method ends. \method{} turns 46.2\% of them from failed to solved, most of them order errors, and none from solved to failed. The \mname{GRPO Solver} turns only 20.0\% from failed to solved and 1.7\% from solved to failed. In training, such fixes leave fewer tasks that fail in every attempt, so RL gets a learning signal from more tasks (Section~\ref{sec:motivation}). \ding{183}~\textbf{\method{} also costs less per attempt.} The harness adds text to the prompt, yet \method{} reads the fewest input tokens per test attempt of all methods in Table~\ref{tab:main}, even fewer than the \mname{Base Agent} with no harness (Figure~\ref{fig:cost}). This is because the harness from the \improver{} makes clear to the \solver{} what to do next. The \solver{} thus finishes in fewer turns on average than the \mname{Base Agent}. Among the five methods of Figure~\ref{fig:cost}, \method{} also generates the fewest tokens, while solving 22.3 points more tasks than the \mname{GRPO Solver}. With the one-time cost of a new workflow included, \method{} spends fewer tokens in total than the \mname{GRPO Solver} after about 140 tasks of the workflow (Table~\ref{tab:cost}). \ding{184}~\textbf{A harness written from a few failures fixes an unseen workflow.} Figure~\ref{fig:case} shows an unseen potassium check workflow. Without a harness, the \solver{} reads all potassium values rather than those of the last 24 hours and places no order, passing 2 of the 11 checks. From these failures, the same model writes a harness, here a single skill. It sets the window, the decision rule and both orders, and the \solver{} then passes all 11 checks with no change to its weights.

\FloatBarrier

\section{Related Work}\label{sec:related_work}

\paragraph{Context-level self-improvement of medical agents.}
Many medical agents improve without updating their weights. EHR and tool-use agents reuse solved cases~\citep{shi2024ehragent,liao2025reflectool}, self-evolving medical agents grow case bases, memories and skill libraries from past attempts~\citep{li2024agenthospital,he2026evoclinician,shen2026evomedagent,han2026gsem,sun2026experience}, and GRASP keeps a proposed skill only if it helps on a held-out set of tasks without making any of them worse~\citep{moll2026grasp}. In all of these methods, the LLM that writes these cases, memories and skills stays fixed and is not trained on the agent's outcomes. \method{} instead trains the \improver{} with RL and rewards each harness it writes by how much it improves the \solver{} over acting without it.

\paragraph{Weight-level self-improvement of medical agents.}
Other work updates the weights of medical models from their own outcomes. RL with verifiable rewards trains medical LLMs and vision-language models for clinical reasoning~\citep{chen2024huatuogpto1,liu2025alphamed,lai2025medr1,pan2025medvlmr1} and ICD coding~\citep{wang2026icd}, multi-turn RL trains agents for biomedical code generation, patient consultation and medical imaging~\citep{xu2026medagentgym,feng2025doctor,lai2025doctorr1,fan2026evolving}, and a self-evolving system for clinical decision support fine-tunes its model on its own successful cases~\citep{zhang2026vibe}. In these methods, the training procedure is designed by hand and stays fixed. A task that every attempt fails gives them no learning signal, and the resulting improvements rarely transfer to other settings~\citep{xi2026rlgeneralization}. \method{} instead learns from such failures through the harness that its own \improver{} writes from them, and trains the \improver{} together with the agent. Section~\ref{sec:exp-transfer} shows that what its \improver{} learns also carries over to other solvers and to a new database.

\paragraph{Recursive self-improvement.}
Recursive self-improvement requires a system to improve not only itself but also the mechanism that improves it~\citep{good1965speculations,schmidhuber2003godel}. Prior work trains the improver itself, such as a reflection model, critics, teachers and a harness engineer, each rewarded by how much it helps another model~\citep{yao2024retroformer,xie2025teaching,cetin2025rlt,shao2026harnessr1}, but the model they help stays frozen, so the improver never has to keep pace with a changing agent. Self-play methods co-evolve a judge or a task proposer with the solver~\citep{yuan2024selfrewarding,zhao2025absolute,huang2026rzero,xia2026agent0}, but this role only shapes what the solver trains on and plays no part when the solver acts on a task. Recent medical work that targets RSI or self-evolution accumulates tools or synthesizes training data with a fixed LLM or pipeline~\citep{wu2026medrsi,chen2026cura,liu2026bat}. Since this LLM or pipeline never improves, such work remains self-improvement rather than RSI. \method{} instead trains one model as both \improver{} and \solver{} on shared attempts, so the \improver{} learns from a changing \solver{} and the two co-evolve. The harness from the \improver{} also guides the \solver{} at test time. Appendix~\ref{app:compare} compares \method{} with more methods.

\section{Conclusion}\label{sec:conclusion}

We presented \method{}, a recursive self-improvement framework in which a medical agent improves not only itself but also its own improving strategy. At its core, Clinical Model-Harness Co-Evolution trains one model to write a medical harness from its failures as the \improver{} and to follow it on each patient as the \solver{}. Each harness is rewarded by how much it improves the agent's outcomes, which needs only outcome feedback and no reference harness. Training both roles together lets the \improver{} keep adapting to the problems the \solver{} has not yet solved. After training, the same model adapts to a new workflow by writing a harness from a few failures, without retraining.

On MedAgentBench and MedAgentGym, \method{} outperforms both context-level and weight-level self-improvement, including on unseen workflows. Our analysis shows that most of the gain comes from training the two roles together, and that the harness must evolve with the model. A harness frozen after the first round falls behind as the model improves. What \method{} learns also carries over to new solvers, improvers and databases. The harness of \method{} fixes errors of every kind on unseen workflows, and its \solver{} reads the fewest input tokens per attempt of all methods. These results suggest that, through Clinical Model-Harness Co-Evolution, medical agents can achieve recursive self-improvement.

\bibliographystyle{abbrvnat}
\bibliography{main}

\appendix
\newpage
\DoToC

\section{Comparison with Self-Improvement Methods}\label{app:compare}

Table~\ref{tab:compare} compares \method{} with representative self-improvement methods from other domains and with methods for clinical agents. Since Definition~\ref{def:rsi} asks whether the improving strategy itself improves, rows within each group go from a fixed improver that changes text, trains the weights or does both to an improver that rewrites its own code, a separately trained improver and an improver co-trained with the solver. Each cell follows the description in the cited paper.

\paragraph{Domain.} The tasks of the evaluation. Clinical covers EHR, consultation and medical imaging tasks. Agents covers interactive household, web, tool-use and coding environments. Reasoning covers math, question answering and code generation, and General covers open-ended instructions.

\paragraph{Object.} What the method changes. Context covers text placed in the prompt other than skills, namely memories, reflections, critiques and prompts. Skills are reusable procedures for a kind of task. Code covers agent, harness and tool code. Text that exists only during training, such as hints that are distilled into the weights and then dropped, counts as weights.

\paragraph{Learns At.} Train methods learn on training tasks and keep what they learn fixed at test time. Test methods improve on the test tasks from their own attempts on them, without updating weights. Both combines the two. Retrieving what was learned on training tasks does not count as learning at test time. A dagger marks a method that evolves its harness and trains its weights on its test tasks.

\paragraph{Trained.} Whether the weights of the solver, the model that acts on the tasks, and of the improver, the component that proposes changes to the agent, are updated, and whether the two share one set of weights, so that the improver changes with the solver.

\paragraph{Transfer.} Whether the paper tests what the method learns, or its trained improver, with a solver built on another base model.

The last row shows how \method{} differs from these methods. (1)~Its harness is designed for clinical workflows. The skills state which records to query, which time window and decision rule to apply and which orders to place, the tools are reusable EHR queries, and memory keeps the best skill or tool of each workflow. The \improver{} writes them from the agent's failures on different patients. (2)~Since clinical workflows often lack a reference harness, each harness is judged only by the agent's outcomes, and patients who need an order, for example, weigh as much in the reward as patients who need none. (3)~One model is trained as both \improver{} and \solver{}. The weights of the \improver{} are trained on the gains its harnesses bring to the \solver{}, and the updated \improver{} writes the harnesses of later rounds, so the ability to improve the agent is itself learned from the agent's outcomes. The trained \improver{} also writes useful harnesses for other solvers.

\begin{table*}[tp]
\centering
\small
\setlength{\tabcolsep}{2.6pt}
\renewcommand{\arraystretch}{1.08}
\caption{\textbf{Comparison of Self-Improvement Methods.} Among clinical agents, only \method{} trains its improver. It trains one model as both solver and improver and shows that this improver transfers to other solvers. Appendix~\ref{app:compare} defines each column and the mark \dag{}.}
\label{tab:compare}
\resizebox{\textwidth}{!}{%
\begin{tabular}{l|l|l|c|ccc|c}
\Xhline{1.2pt}
\rowcolor{headcolor}
 & & & & \multicolumn{3}{c|}{\hgroup{Trained}} & \\
\rowcolor{headcolor}
\hgroup{Method} & \hcol{Domain} & \hcol{Object} & \hcol{\makecell{Learns\\At}} & \hcol{Solver} & \hcol{Improver} & \hcol{\makecell{Shared\\Weights}} & \hcol{Transfer} \\
\Xhline{1pt}
\multicolumn{8}{c}{\cellcolor{catcolor}\textit{Other Domains}} \\
Reflexion~\citep{shinn2023reflexion} & Agents, Reasoning & Context & Test & \xmark & \xmark & \xmark & \xmark \\
ExpeL~\citep{zhao2024expel} & Agents & Context & Train & \xmark & \xmark & \xmark & \xmark \\
AWM~\citep{wang2025awm} & Agents & Skills & Both & \xmark & \xmark & \xmark & \xmark \\
DC~\citep{suzgun2026dynamic} & Reasoning & Context & Test & \xmark & \xmark & \xmark & \xmark \\
GEPA~\citep{agrawal2026gepa} & Reasoning & Context & Train & \xmark & \xmark & \xmark & \cmark \\
ACE~\citep{zhang2026ace} & Agents, Reasoning & Context & Both & \xmark & \xmark & \xmark & \xmark \\
Harness$^2$~\citep{xu2026harness2} & Agents & Code & Test & \xmark & \xmark & \xmark & \cmark \\
\arrayrulecolor{black!25}\hline\arrayrulecolor{black}
STaR~\citep{zelikman2022star} & Reasoning & Weights & Train & \cmark & \xmark & \xmark & \xmark \\
GRPO~\citep{shao2024deepseekmath} & Reasoning & Weights & Train & \cmark & \xmark & \xmark & \xmark \\
SkillRL~\citep{xia2026skillrl} & Agents & Skills\,+\,Weights & Train & \cmark & \xmark & \xmark & \xmark \\
SkillForge~\citep{yang2026skillforge} & Agents & Skills\,+\,Weights & Train & \cmark & \xmark & \xmark & \cmark \\
\arrayrulecolor{black!25}\hline\arrayrulecolor{black}
SIA~\citep{hebbar2026sia} & Agents & Code\,+\,Weights & Test$^\dagger$ & \cmark & \xmark & \xmark & \xmark \\
HarnessForge~\citep{chen2026harnessforge} & Agents, Reasoning & Code\,+\,Weights & Train & \cmark & \xmark & \xmark & \xmark \\
HarnessX~\citep{chen2026harnessx} & Agents & Code\,+\,Weights & Test$^\dagger$ & \cmark & \xmark & \xmark & \xmark \\
WHALE~\citep{kim2026whale} & Reasoning & Code\,+\,Weights & Train & \cmark & \xmark & \xmark & \xmark \\
SafeEvolve~\citep{mao2026safeevolve} & Agents & Skills\,+\,Weights & Train & \cmark & \xmark & \xmark & \cmark \\
VACE~\citep{qi2026vace} & Agents, Reasoning & Skills\,+\,Weights & Train & \cmark & \xmark & \xmark & \xmark \\
\arrayrulecolor{black!25}\hline\arrayrulecolor{black}
STOP~\citep{zelikman2024stop} & Reasoning & Code & Train & \xmark & \xmark & \xmark & \xmark \\
DGM~\citep{zhang2026darwin} & Agents & Code & Train & \xmark & \xmark & \xmark & \cmark \\
\arrayrulecolor{black!25}\hline\arrayrulecolor{black}
CTRL~\citep{xie2025teaching} & Reasoning & Context & Both & \xmark & \cmark & \xmark & \cmark \\
RLT~\citep{cetin2025rlt} & Reasoning & Weights & Train & \cmark & \cmark & \xmark & \cmark \\
Harness-R1~\citep{shao2026harnessr1} & Agents & Code & Both & \xmark & \cmark & \xmark & \cmark \\
Skill-$\alpha$~\citep{shen2026skillalpha} & Agents & Skills & Train & \xmark & \cmark & \xmark & \cmark \\
SkillOS~\citep{ouyang2026skillos} & Agents, Reasoning & Skills & Both & \xmark & \cmark & \xmark & \cmark \\
\arrayrulecolor{black!25}\hline\arrayrulecolor{black}
SRLM~\citep{yuan2024selfrewarding} & General & Weights & Train & \cmark & \cmark & \cmark & \xmark \\
Absolute Zero~\citep{zhao2025absolute} & Reasoning & Weights & Train & \cmark & \cmark & \cmark & \xmark \\
\Xhline{0.6pt}
\multicolumn{8}{c}{\cellcolor{catcolor}\textit{Clinical Agents}} \\
EHRAgent~\citep{shi2024ehragent} & Clinical & Context & Test & \xmark & \xmark & \xmark & \xmark \\
Evo-MedAgent~\citep{shen2026evomedagent} & Clinical & Context & Test & \xmark & \xmark & \xmark & \xmark \\
EvoClinician~\citep{he2026evoclinician} & Clinical & Context & Test & \xmark & \xmark & \xmark & \xmark \\
GRASP~\citep{moll2026grasp} & Clinical, Agents & Skills & Train & \xmark & \xmark & \xmark & \cmark \\
SkeMex~\citep{sun2026experience} & Clinical & Skills & Both & \xmark & \xmark & \xmark & \cmark \\
MedRSI~\citep{wu2026medrsi} & Clinical & Code & Train & \xmark & \xmark & \xmark & \xmark \\
MedAgentGym~\citep{xu2026medagentgym} & Clinical & Weights & Train & \cmark & \xmark & \xmark & \xmark \\
Cura 1T~\citep{chen2026cura} & Clinical & Weights & Train & \cmark & \xmark & \xmark & \xmark \\
\Xhline{0.6pt}
\method{} (ours) & Clinical & Skills\,+\,Code\,+\,Weights & Both & \cmark & \cmark & \cmark & \cmark \\
\Xhline{1.2pt}
\end{tabular}}
\end{table*}

\section{Training Details}\label{app:training}

\paragraph{Warm start.}
The warm start mainly gives the \improver{} a well-formed way to write harnesses from the start, so that RL learns faster. On each benchmark, a stronger LLM (\llmname{GPT-6.1-Sol}) writes several candidate harnesses for every training workflow from the attempts of the \mname{Base Agent} at its training tasks. The \mname{Base Agent} attempts these tasks with each candidate, and we keep the candidates that solve at least two more tasks than the \mname{Base Agent} and come within three tasks of the best one, up to twelve per workflow. If no candidate reaches this bar, we keep up to four that solve no fewer tasks than the \mname{Base Agent}, so that every training workflow has examples. The kept harnesses, paired with workflow evidence that contains no reference answers, train a LoRA adapter for three epochs with a learning rate of $10^{-4}$ and a cosine schedule, and RL starts from this adapter. Without the warm start, RL first has to learn this format: in its first two rounds on MedAgentBench, 11\% of the harnesses written from a fresh adapter do not parse and almost half are too long, against 2\% and 13\% with the warm start.

\paragraph{Workflow evidence.}
The workflow evidence $P_\tau$ holds at most five failed attempts, chosen to cover the different kinds of failure seen in the trajectories, such as an invalid action or a rejected order, and at most two successful ones, or three when no attempt fails. Each attempt also shows how it ended, for example with an answer or at the turn limit.

\paragraph{Harness components.}
As in Section~\ref{sec:roles}, each candidate harness adds one skill or one tool.
\begin{itemize}[label={$\triangleright$},leftmargin=1.1em,labelsep=0.4em,itemsep=1pt,topsep=2pt,parsep=0pt,partopsep=0pt]
\item \textbf{Skills}: a short markdown procedure with the three parts of Section~\ref{sec:roles}, \textit{When to use}, \textit{Steps} and \textit{Pitfalls}. It is placed in the first prompt of the \solver{}, and a skill that names a specific patient is discarded. On MedAgentBench, every candidate is a skill.
\item \textbf{Tools}: on MedAgentGym and eICU, each candidate is either a skill or a tool, which is a set of SQL views with short documentation. The views are created before every query, and the \solver{} sees only their names, columns and documentation.
\item \textbf{Memory}: one skill or tool per workflow, updated in three steps in each round. (1)~Since the \solver{} changes from round to round, the current \solver{} first attempts the training tasks of the workflow again with the stored skill or tool at temperature~$0$. (2)~Half of the candidates revise the stored skill or tool from these attempts, and the other half are written from scratch from the workflow evidence. The stored one is also attempted once on every task together with the candidates. (3)~The best candidate of the round is attempted once more at temperature~$0$. It replaces the stored one only if its gain is higher by at least $\delta = 0.03$, both in the sampled attempts of the round and in this extra attempt. A workflow with nothing stored gets $G$ new candidates, and its best one is stored if its gain is positive in both kinds of attempts.
\item \textbf{New workflows at test time}: the \improver{} writes candidates from the first attempts on the new workflow, with the skills or tools stored for the three most similar training workflows as references. Similarity is the cosine similarity between the TF-IDF vectors of their task texts.
\end{itemize}

\paragraph{Co-evolution.}
The \solver{}'s first attempts, which form the workflow evidence and the baseline of the reward (Appendix~\ref{app:reward}), run at temperature $0$, and all attempts used for training at temperature $1$. For the \solver{}, each task forms one group from $G$ of its attempts with a harness and its $N$ attempts with $\mathcal{H}_0$. When memory stores a skill or tool for the workflow, the task has $G + 1$ attempts with a harness, and $G$ of them are drawn at random, so that every group has the same size. Algorithm~\ref{alg:selfrx} lists one round, and Table~\ref{tab:hparams} lists the hyperparameters.

\begin{algorithm}[t]
\caption{One Round of \method{}}
\label{alg:selfrx}
\small
\begin{algorithmic}[1]
\Require weights $\theta$, memory $\mathcal{M}$, training workflows $\mathcal{T}_t$ of round $t$, $G$ candidates per workflow, $N$ attempts per task with $\mathcal{H}_0$, margin $\delta$
\For{each workflow $\tau \in \mathcal{T}_t$}
  \State \textcolor{solcolor}{\textbf{Solve:}} $\xi_x \sim \Sol_\theta(\cdot \mid x, \mathcal{H}_0)$ at temperature $0$, $x \in \tau$
  \State $P_\tau \gets \textsc{WorkflowEvidence}(\{\xi_x\}_{x \in \tau})$
  \If{memory stores a skill or tool $\mathcal{M}_\tau$ for $\tau$}
    \State \textcolor{solcolor}{\textbf{Re-check:}} $\xi'_x \sim \Sol_\theta(\cdot \mid x, \mathcal{M}_\tau)$ at temperature $0$, $x \in \tau$
    \State $P'_\tau \gets \textsc{WorkflowEvidence}(\{\xi'_x\}_{x \in \tau})$
    \State \textcolor{impcolor}{\textbf{Write:}} $\mathcal{H}_1, \dots, \mathcal{H}_{G/2} \sim \Imp_\theta(\cdot \mid \mathcal{M}_\tau, P'_\tau)$ \Comment{revisions of $\mathcal{M}_\tau$}
    \State \phantom{\textbf{Write:}} $\mathcal{H}_{G/2+1}, \dots, \mathcal{H}_G \sim \Imp_\theta(\cdot \mid P_\tau)$ \Comment{written from scratch}
  \Else
    \State \textcolor{impcolor}{\textbf{Write:}} $\mathcal{H}_1, \dots, \mathcal{H}_G \sim \Imp_\theta(\cdot \mid P_\tau)$
  \EndIf
  \For{each task $x \in \tau$}
    \State \textcolor{solcolor}{\textbf{Evaluate:}} $\xi_{x,g} \sim \Sol_\theta(\cdot \mid x, \mathcal{H}_g)$, $g = 1, \dots, G$, and one attempt with $\mathcal{M}_\tau$ if it is stored
    \State \phantom{\textbf{Evaluate:}} $N$ trajectories $\sim \Sol_\theta(\cdot \mid x, \mathcal{H}_0)$
  \EndFor
  \State \textbf{Reward:} $R(\mathcal{H}_g)$ by Eq.~\eqref{eq:gain}, $r_{x,j} = \mathbb{I}[\mu(\xi_{x,j}) = 1]$
  \State \textbf{Keep:} $\mathcal{H}^\star \gets \arg\max_g R(\mathcal{H}_g)$, attempted once more at temperature $0$
  \State \phantom{\textbf{Keep:}} $\mathcal{M}_\tau \gets \mathcal{H}^\star$ if its gain beats that of $\mathcal{M}_\tau$ by at least $\delta$ at temperatures $1$ and $0$
  \State \phantom{\textbf{Keep:}} (if nothing is stored for $\tau$: if its gain is positive at both temperatures)
\EndFor
\State $\hat{A}^{\textsc{imp}}, \hat{A}^{\textsc{sol}}$ by Eq.~\eqref{eq:adv}, the latter over $G$ of the attempts with a harness, drawn at random when there are $G+1$, and the $N$ with $\mathcal{H}_0$ of each task
\State \textcolor{learncolor}{\textbf{Update:}} two AdamW steps on $\mathcal{L}(\theta)$, Eq.~\eqref{eq:loss}
\State \Return $\theta$, $\mathcal{M}$
\end{algorithmic}
\end{algorithm}

\begin{table}[t]
\centering
\small
\setlength{\tabcolsep}{4pt}
\caption{\textbf{Training Hyperparameters of \method{}.} \llmname{Gemma-4-E4B} and \llmname{Qwen3.5-4B} use the same settings.}
\label{tab:hparams}
\begin{tabularx}{\columnwidth}{l>{\raggedright\arraybackslash}X}
\Xhline{1.2pt}
\rowcolor{headcolor}
\hgroup{Setting} & \hgroup{Value} \\
\Xhline{1pt}
Base model & \llmname{Qwen3.5-9B}, thinking off \\
LoRA rank / $\alpha$ / dropout & 64 / 128 / 0.05 \\
Rounds & 10 \\
Training workflows per round & 13 of 26, all 20 tasks \\
Candidates per workflow ($G$) & 8 \\
\solver attempts per task with $\mathcal{H}_0$ ($N$) & 2 \\
Sampling in training & temperature 1.0, top-$p$ 0.95 \\
Agent limits & 8 turns, 512 tokens per turn \\
Optimizer & AdamW, $\beta = (0.9, 0.99)$ \\
Learning rate & $10^{-5}$ \\
Optimizer steps per round & 2 \\
Clipping $\epsilon_{\mathrm{low}}$ / $\epsilon_{\mathrm{high}}$ & 0.2 / 0.28 \\
Gradient-norm clipping & 1.0 \\
\improver reward & mean of outcome and process-score gains \\
Unparseable candidate & reward $-0.5$ \\
Length penalty & $5{\times}10^{-4}$ per character beyond 1{,}500, at most 0.1 \\
Memory replacement margin ($\delta$) & 0.03 \\
Harness sampling at test time & temperature 0.7, top-$p$ 0.8 \\
Candidates per workflow at test time ($K$) & 16 \\
\Xhline{1.2pt}
\end{tabularx}
\end{table}

\section{Reward Computation}\label{app:reward}

Both rewards of Section~\ref{sec:coevo} come from the outcome feedback of the grader. Neither role sees the reference answers or the checks of the grader.

\paragraph{Solver reward.} Each attempt of the \solver{} receives its outcome $r$, which is 1 if it solves the task and 0 otherwise. Its advantage in Eq.~\eqref{eq:adv} compares it with the other attempts in the group of its task (Appendix~\ref{app:training}), so a group whose attempts all solve the task, or all fail, gives no gradient.

\paragraph{Attempt score.} The reward of a harness builds on the score $u = \tfrac{1}{2}(r + \mu)$ of each attempt, with $\mu$ as in Section~\ref{sec:prelim}, so a failed attempt scores at most 0.25. On MedAgentBench, $c$ is the mean of the shares of query checks and action checks passed. An attempt that exceeds the time limit counts as failed with no check passed, and one that ends in an environment error is not scored.

\paragraph{Gain.} The baseline of each task is the attempt of the \solver{} with $\mathcal{H}_0$ at temperature $0$, which also supplies the workflow evidence. The gain of a candidate on a task is its score minus the baseline score, counted on the tasks where both attempts were scored. On MedAgentBench, the grader records whether each task expects an order. In a workflow where some tasks expect an order and others do not, the two kinds form two case types. The gains are averaged within each case type and then across the case types, which count equally however many tasks each holds. Every other workflow, and every workflow of MedAgentGym, forms one case type. Like the outcome, the case type comes from the grader and is never shown to either role. A candidate scored on fewer than 15 tasks, or 10 on MedAgentGym, receives no reward and joins no group.

\paragraph{Penalties.} The length penalty is $\lambda(\mathcal{H}) = \min\{0.1,\ 5{\times}10^{-4} \cdot \max(0,\ n - 1{,}500)\}$, where $n$ is the length of the harness in characters, not counting its header with a name and a description. Only the part of a tool's SQL code beyond 2{,}000 characters counts toward $n$. A candidate that does not follow the format, for example one without this header, is not attempted and receives a reward of $-0.5$.

\paragraph{Groups.} The candidates of a workflow written from the same prompt form one group, so new candidates are never compared with revisions of the stored skill or tool. The advantage of each candidate is its reward minus the mean reward of its group, not divided by the standard deviation.

\section{Data and Task Construction}\label{app:data}

\paragraph{MedAgentBench.} Our main benchmark combines MedAgentBench~\citep{jiang2025medagentbench}, its second version~\citep{chen2025medagentbenchv2} and 12 new task types.
\begin{itemize}[label={$\triangleright$},leftmargin=1.1em,labelsep=0.4em,itemsep=1pt,topsep=2pt,parsep=0pt,partopsep=0pt]
\item \textbf{Environment}: the agent operates a FHIR server with one action per turn, a GET or POST request or \texttt{FINISH} with the answer, and at most eight turns.
\item \textbf{Workflows}: following the benchmark's own division, each of the 10 task types of the first version, the 10 of the second and the 12 new ones is one clinical workflow. The new ones come from draft templates that clinicians helped write in the repository of the second version and that it did not adopt, such as removing duplicate medication orders or ordering a test when a recent lab value crosses a threshold. We wrote their task context and graders in the style of the second version.
\item \textbf{Split}: each workflow has 20 training and 10 test tasks. Twenty-six workflows are used for training, and six, two from each source drawn at random, are held out as unseen workflows. This gives 520 training and 320 test tasks, and the tables report the 60 from unseen workflows as out-of-domain. The six unseen workflows also have 20 training tasks each in the original split. \method{} never trains on them, and Table~\ref{tab:train} writes harnesses from them.
\end{itemize}

\paragraph{MedAgentGym.} As a second benchmark, we use the EHR-SeqSQL subset~\citep{ryu2024ehrseqsql} of MedAgentGym~\citep{xu2026medagentgym}, text-to-SQL follow-up questions on the MIMIC-III database of EHRSQL~\citep{lee2022ehrsql,johnson2016mimic} with 1,000 patients.
\begin{itemize}[label={$\triangleright$},leftmargin=1.1em,labelsep=0.4em,itemsep=1pt,topsep=2pt,parsep=0pt,partopsep=0pt]
\item \textbf{Environment}: the agent answers a clinical question by querying a read-only SQLite copy of the database, with one action per turn, \texttt{SQL} with a query or \texttt{FINISH} with the answer, and at most eight turns, the same budget as on MedAgentBench. Each question follows an earlier question of the same conversation, and the prompt quotes that question because it names the patient, the admission or the item.
\item \textbf{Workflows}: in MedAgentGym's own split, almost every question template appears only once, so a workflow could not have both training and test tasks. We therefore re-split it in the same way as MedAgentBench. (1)~We combine its training and test questions. (2)~We keep one question per distinct gold SQL query that runs and returns one to ten rows. (3)~Each query structure, the gold SQL with every value masked, is a workflow. The tasks of a workflow thus differ only in patients, items, dates and numbers, as those of a MedAgentBench workflow differ in the patient.
\item \textbf{Split}: the training workflows are 26 structures drawn at random from those with at least 30 questions, with 20 training and 10 test tasks each. The unseen workflows are 6 structures drawn at random from the remaining ones with at least 10 questions, with 10 test tasks each. So that no unseen workflow nearly repeats a training workflow, its query pattern, with table and column names also masked, must have a token Jaccard similarity below 0.8 to that of every training workflow. This also yields 520 training and 320 test tasks, 60 of them from unseen workflows. The SFT warm start follows Appendix~\ref{app:training} on the 26 training workflows.
\item \textbf{Grading}: MedAgentGym computed the answers with the system clock, so we recompute them with the current time fixed at 2105-12-31 23:59:00, as in the EHRSQL scorer. Answers are compared as sets of values rounded to three decimals. For all methods, an answer also counts as correct when it differs from the gold answer only in repeated whitespace or in a midnight time after a date, such as \texttt{2105-07-27} for \texttt{2105-07-27 00:00:00}.
\end{itemize}

\paragraph{eICU.} For the transfer to a new database (Section~\ref{sec:exp-transfer}), we use the eICU questions of EHRSQL~\citep{lee2022ehrsql} on its eICU database~\citep{pollard2018eicu}, 1,000 patients from 160 US hospitals with a schema different from MIMIC-III. We draw 32 templates at random from those with at least 20 training and 10 test or validation questions, treat each as a workflow and take 10 test tasks from each, 320 in total. The models trained on MedAgentGym are tested without further weight updates in the same environment, with the current time again fixed at 2105-12-31 23:59:00 and answers compared as sets of values rounded to three decimals.

\section{Experimental Setup}\label{app:setup}

\paragraph{Metrics.} Outcome is the share of tasks solved. Process averages the process score $\mu$ of Section~\ref{sec:prelim} over the test tasks, so a solved task counts one point and a failed task up to half a point by the share of grader checks it passes. Both are in \%, with bootstrap standard deviations over test tasks.

\paragraph{Baselines.} We group the baselines as in Section~\ref{sec:exp-setup}.
\begin{itemize}[label={$\triangleright$},leftmargin=1.1em,labelsep=0.4em,itemsep=1pt,topsep=2pt,parsep=0pt,partopsep=0pt]
\item \textbf{Context-level}: methods that learn over several epochs use the epoch that their authors' code selects on validation tasks drawn from the training split. \mname{SFT Improver} is the SFT warm start that \mname{GRPO Improver} and \method{} both start from, used as the improver for the frozen base model. \mname{GRPO Improver} re-implements Harness-R1~\citep{shao2026harnessr1} for clinical workflows. It follows the same recipe, an SFT warm start on harnesses that a stronger LLM writes and then GRPO for a frozen solver, with skills in place of harness edits.
\item \textbf{Weight-level}: the \mname{GRPO Solver} uses the multi-turn GRPO update and the hyperparameters of the \solver{} of \method{} in Table~\ref{tab:hparams}, with 13 training workflows per round and 8 attempts per task with $\mathcal{H}_0$. \mname{STaR} trains the solver on its own successful attempts as in STaR~\citep{zelikman2022star}, also with 8 attempts per task. It runs five iterations and is tested after the last, except on MedAgentGym, where its fifth iteration degrades and the fourth is used.
\end{itemize}

\section{Results with \texorpdfstring{\llmname{Qwen3.5-4B}}{Qwen3.5-4B}}\label{app:mab4b}
Table~\ref{tab:mab4b} reports the results on MedAgentBench with \llmname{Qwen3.5-4B} as the backbone, in the same setting as Table~\ref{tab:main}.

\providecommand{\frozen}{\textcolor{myblue}{\faSnowflake}}
\providecommand{\trained}{\textcolor{pyorange}{\faFire}}
\ifcsname NC@find@Y\endcsname\else\newcolumntype{Y}{>{\centering\arraybackslash}X}\fi
\begin{table*}[t!]
\centering
\footnotesize
\setlength{\tabcolsep}{1.8pt}
\renewcommand{\arraystretch}{1}
\caption{\textbf{Results on MedAgentBench} with \llmname{Qwen3.5-4B}. \method{} is best in every column. Outcome is the share of tasks solved, and Process also gives partial credit for passed grader checks, both in \% with bootstrap standard deviations. Out-of-domain workflows are held out from training. \frozen{} frozen, \trained{} trained. \hlfirst{Best} and \hlsecond{second best} per column. Baselines are described in Appendix~\ref{app:setup}.}
\label{tab:mab4b}
\vspace{-3pt}
\begin{tabularx}{\textwidth}{cl|cc|YY|YY!{\vrule width 0.8pt}YY}
\Xhline{1.2pt}
\rowcolor{headcolor}
 & & & & \multicolumn{2}{c|}{\hgroup{In-Domain}} & \multicolumn{2}{c!{\vrule width 0.8pt}}{\hgroup{Out-of-Domain}} & \multicolumn{2}{c}{\hgroup{Overall}} \\
\rowcolor{headcolor}
\hgroup{Model} & \quad\hgroup{Method} & \hcol{Improver} & \hcol{Solver} & \hcol{Process} & \hcol{Outcome} & \hcol{Process} & \hcol{Outcome} & \hcol{Process} & \hcol{Outcome} \\
\Xhline{1pt}
\multirow{18}{*}{\mdl{Qwen3.5}{4B}}
 & \multicolumn{9}{c}{\cellcolor{catcolor}\textit{No Self-Improvement}} \\
  & \quad \mname{Base Agent} & -- & \frozen & 61.0\textsubscript{\tiny$\pm$2.3} & 45.8\textsubscript{\tiny$\pm$3.1} & 49.0\textsubscript{\tiny$\pm$4.6} & 31.7\textsubscript{\tiny$\pm$6.2} & 58.8\textsubscript{\tiny$\pm$1.9} & 43.1\textsubscript{\tiny$\pm$2.8} \\
 & \multicolumn{9}{c}{\cellcolor{catcolor}\textit{Context-Level Self-Improvement}} \\
  & \quad \mname{Reflexion} & \frozen & \frozen & 65.8\textsubscript{\tiny$\pm$2.3} & 53.5\textsubscript{\tiny$\pm$3.0} & 51.9\textsubscript{\tiny$\pm$4.7} & 35.0\textsubscript{\tiny$\pm$6.3} & 63.2\textsubscript{\tiny$\pm$2.0} & 50.0\textsubscript{\tiny$\pm$2.8} \\
  & \quad \mname{ExpeL} & \frozen & \frozen & 57.8\textsubscript{\tiny$\pm$2.3} & 43.5\textsubscript{\tiny$\pm$3.0} & 47.3\textsubscript{\tiny$\pm$4.4} & 28.3\textsubscript{\tiny$\pm$5.9} & 55.9\textsubscript{\tiny$\pm$2.0} & 40.6\textsubscript{\tiny$\pm$2.7} \\
  & \quad \mname{GEPA} & \frozen & \frozen & 64.5\textsubscript{\tiny$\pm$2.5} & 53.5\textsubscript{\tiny$\pm$3.0} & 47.6\textsubscript{\tiny$\pm$4.6} & 30.0\textsubscript{\tiny$\pm$5.9} & 61.4\textsubscript{\tiny$\pm$2.1} & 49.1\textsubscript{\tiny$\pm$2.9} \\
  & \quad \mname{EvoMedAgent} & \frozen & \frozen & 59.5\textsubscript{\tiny$\pm$2.3} & 45.0\textsubscript{\tiny$\pm$3.1} & 48.9\textsubscript{\tiny$\pm$4.6} & 31.7\textsubscript{\tiny$\pm$6.2} & 57.6\textsubscript{\tiny$\pm$2.0} & 42.5\textsubscript{\tiny$\pm$2.8} \\
  & \quad \mname{SkillX} & \frozen & \frozen & 53.2\textsubscript{\tiny$\pm$2.2} & 35.4\textsubscript{\tiny$\pm$3.0} & 34.9\textsubscript{\tiny$\pm$3.2} & 11.7\textsubscript{\tiny$\pm$4.1} & 49.7\textsubscript{\tiny$\pm$1.9} & 30.9\textsubscript{\tiny$\pm$2.6} \\
  & \quad \mname{GRASP} & \frozen & \frozen & 68.8\textsubscript{\tiny$\pm$2.4} & 58.5\textsubscript{\tiny$\pm$3.0} & 49.4\textsubscript{\tiny$\pm$4.6} & 31.7\textsubscript{\tiny$\pm$6.1} & 65.2\textsubscript{\tiny$\pm$2.1} & 53.4\textsubscript{\tiny$\pm$2.9} \\
  & \quad \mname{Fixed Improver} & \frozen & \frozen & 62.0\textsubscript{\tiny$\pm$2.2} & 48.0\textsubscript{\tiny$\pm$2.7} & 48.8\textsubscript{\tiny$\pm$4.1} & 30.7\textsubscript{\tiny$\pm$5.5} & 59.6\textsubscript{\tiny$\pm$1.8} & 44.7\textsubscript{\tiny$\pm$2.6} \\
  & \quad \mname{Fixed Improver (GPT)} & \frozen & \frozen & 65.9\textsubscript{\tiny$\pm$2.0} & 51.6\textsubscript{\tiny$\pm$2.8} & \hlsecond{63.1\textsubscript{\tiny$\pm$4.2}} & \hlsecond{45.6\textsubscript{\tiny$\pm$5.9}} & 65.4\textsubscript{\tiny$\pm$1.7} & 50.4\textsubscript{\tiny$\pm$2.6} \\
  & \quad \mname{SFT Improver} & \trained & \frozen & 65.2\textsubscript{\tiny$\pm$1.8} & 50.0\textsubscript{\tiny$\pm$2.6} & 53.1\textsubscript{\tiny$\pm$2.4} & 31.1\textsubscript{\tiny$\pm$3.2} & 62.9\textsubscript{\tiny$\pm$1.5} & 46.4\textsubscript{\tiny$\pm$2.3} \\
  & \quad \mname{GRPO Improver} & \trained & \frozen & 70.7\textsubscript{\tiny$\pm$1.9} & 58.6\textsubscript{\tiny$\pm$2.6} & 56.9\textsubscript{\tiny$\pm$3.3} & 36.0\textsubscript{\tiny$\pm$4.8} & 68.1\textsubscript{\tiny$\pm$1.6} & 54.4\textsubscript{\tiny$\pm$2.4} \\
 & \multicolumn{9}{c}{\cellcolor{catcolor}\textit{Weight-Level Self-Improvement}} \\
  & \quad \mname{STaR} & -- & \trained & 66.4\textsubscript{\tiny$\pm$2.4} & 55.8\textsubscript{\tiny$\pm$3.1} & 53.7\textsubscript{\tiny$\pm$4.7} & 36.7\textsubscript{\tiny$\pm$6.3} & 64.0\textsubscript{\tiny$\pm$2.1} & 52.2\textsubscript{\tiny$\pm$2.9} \\
  & \quad \mname{GRPO Solver} & -- & \trained & \hlsecond{80.1\textsubscript{\tiny$\pm$2.0}} & \hlsecond{72.3\textsubscript{\tiny$\pm$2.9}} & 47.7\textsubscript{\tiny$\pm$4.3} & 26.7\textsubscript{\tiny$\pm$5.8} & \hlsecond{74.1\textsubscript{\tiny$\pm$2.0}} & \hlsecond{63.7\textsubscript{\tiny$\pm$2.8}} \\
 & \multicolumn{9}{c}{\cellcolor{catcolor}\textit{Ours}} \\
  & \quad \method{} & \trained & \trained & \hlfirst{91.0\textsubscript{\tiny$\pm$1.4}} & \hlfirst{85.8\textsubscript{\tiny$\pm$2.1}} & \hlfirst{72.5\textsubscript{\tiny$\pm$3.5}} & \hlfirst{57.5\textsubscript{\tiny$\pm$5.3}} & \hlfirst{87.6\textsubscript{\tiny$\pm$1.4}} & \hlfirst{80.5\textsubscript{\tiny$\pm$2.1}} \\
\Xhline{1.2pt}
\end{tabularx}
\vspace{-0.3cm}
\end{table*}

\section{Variation across Training Runs}\label{app:runs}
Table~\ref{tab:runs} reports three training runs of \method{} on \llmname{Qwen3.5-9B}. Their overall outcome is 84.1\% on average with a standard deviation of 2.7 points, and run 1, which Table~\ref{tab:main} reports, scores 85.1\%. Every run is at least 18 points above the strongest baseline in Table~\ref{tab:main}, the GRPO Solver at 62.8\%.

\ifcsname NC@find@Y\endcsname\else\newcolumntype{Y}{>{\centering\arraybackslash}X}\fi
\begin{table*}[t!]
\centering
\small
\setlength{\tabcolsep}{3pt}
\caption{\textbf{Three Training Runs of \method{}.} All three runs of \method{} are at least 18 points above the strongest baseline of Table~\ref{tab:main}. The runs share all settings and the SFT warm start and differ only in the randomness of sampling. Run 1 is the run in Table~\ref{tab:main}. Process and Outcome in \%, with bootstrap standard deviations over test tasks as subscripts. Mean gives the mean over the three runs, with the standard deviation over the runs as subscripts.}
\label{tab:runs}
\begin{tabularx}{\textwidth}{cl|YY|YY!{\vrule width 0.8pt}YY}
\Xhline{1.2pt}
\rowcolor{headcolor}
 & & \multicolumn{2}{c|}{\hgroup{In-Domain}} & \multicolumn{2}{c!{\vrule width 0.8pt}}{\hgroup{Out-of-Domain}} & \multicolumn{2}{c}{\hgroup{Overall}} \\
\rowcolor{headcolor}
\hgroup{Model} & \hgroup{Run} & \hcol{Process} & \hcol{Outcome} & \hcol{Process} & \hcol{Outcome} & \hcol{Process} & \hcol{Outcome} \\
\Xhline{1pt}
\multirow{4}{*}{\mdl{Qwen3.5}{9B}}
 & Run 1 & 93.0\textsubscript{\tiny$\pm$1.3} & 89.5\textsubscript{\tiny$\pm$1.9} & 78.7\textsubscript{\tiny$\pm$3.3} & 66.1\textsubscript{\tiny$\pm$5.1} & 90.3\textsubscript{\tiny$\pm$1.2} & 85.1\textsubscript{\tiny$\pm$1.9} \\
 & Run 2 & 89.9\textsubscript{\tiny$\pm$1.6} & 85.2\textsubscript{\tiny$\pm$2.2} & 76.3\textsubscript{\tiny$\pm$3.5} & 63.2\textsubscript{\tiny$\pm$5.4} & 87.3\textsubscript{\tiny$\pm$1.4} & 81.1\textsubscript{\tiny$\pm$2.0} \\
 & Run 3 & 93.3\textsubscript{\tiny$\pm$1.3} & 89.8\textsubscript{\tiny$\pm$1.9} & 81.9\textsubscript{\tiny$\pm$3.2} & 70.8\textsubscript{\tiny$\pm$5.1} & 91.1\textsubscript{\tiny$\pm$1.1} & 86.2\textsubscript{\tiny$\pm$1.9} \\
\cline{2-8}
 & Mean & \textbf{92.0}\textsubscript{\tiny$\pm$1.9} & \textbf{88.2}\textsubscript{\tiny$\pm$2.6} & \textbf{79.0}\textsubscript{\tiny$\pm$2.8} & \textbf{66.7}\textsubscript{\tiny$\pm$3.8} & \textbf{89.6}\textsubscript{\tiny$\pm$2.0} & \textbf{84.1}\textsubscript{\tiny$\pm$2.7} \\
\Xhline{1.2pt}
\end{tabularx}
\end{table*}

\section{Results on MedAgentGym}\label{app:ehrseqsql}
Table~\ref{tab:ehrseqsql} reports the results on MedAgentGym (Appendix~\ref{app:data}).

\providecommand{\frozen}{\textcolor{myblue}{\faSnowflake}}
\providecommand{\trained}{\textcolor{pyorange}{\faFire}}
\ifcsname NC@find@Y\endcsname\else\newcolumntype{Y}{>{\centering\arraybackslash}X}\fi
\begin{table*}[t!]
\centering
\footnotesize
\setlength{\tabcolsep}{1.8pt}
\renewcommand{\arraystretch}{1}
\caption{\textbf{Results on MedAgentGym.} \method{} is best in every column. Outcome is the share of tasks solved, and Process also gives partial credit for passed grader checks, both in \% with bootstrap standard deviations. Out-of-domain workflows are held out from training. \frozen{} frozen, \trained{} trained. \hlfirst{Best} and \hlsecond{second best} per column. Baselines are described in Appendix~\ref{app:setup}.}
\label{tab:ehrseqsql}
\vspace{-3pt}
\begin{tabularx}{\textwidth}{cl|cc|YY|YY!{\vrule width 0.8pt}YY}
\Xhline{1.2pt}
\rowcolor{headcolor}
 & & & & \multicolumn{2}{c|}{\hgroup{In-Domain}} & \multicolumn{2}{c!{\vrule width 0.8pt}}{\hgroup{Out-of-Domain}} & \multicolumn{2}{c}{\hgroup{Overall}} \\
\rowcolor{headcolor}
\hgroup{Model} & \quad\hgroup{Method} & \hcol{Improver} & \hcol{Solver} & \hcol{Process} & \hcol{Outcome} & \hcol{Process} & \hcol{Outcome} & \hcol{Process} & \hcol{Outcome} \\
\Xhline{1pt}
\multirow{18}{*}{\mdl{Qwen3.5}{9B}}
 & \multicolumn{9}{c}{\cellcolor{catcolor}\textit{No Self-Improvement}} \\
  & \quad \mname{Base Agent} & -- & \frozen & 43.7\textsubscript{\tiny$\pm$2.3} & 28.1\textsubscript{\tiny$\pm$2.7} & 36.0\textsubscript{\tiny$\pm$3.3} & 13.3\textsubscript{\tiny$\pm$4.4} & 42.3\textsubscript{\tiny$\pm$2.0} & 25.3\textsubscript{\tiny$\pm$2.6} \\
 & \multicolumn{9}{c}{\cellcolor{catcolor}\textit{Context-Level Self-Improvement}} \\
  & \quad \mname{Reflexion} & \frozen & \frozen & 49.6\textsubscript{\tiny$\pm$2.5} & 37.7\textsubscript{\tiny$\pm$2.9} & 42.1\textsubscript{\tiny$\pm$4.4} & 25.0\textsubscript{\tiny$\pm$5.6} & 48.2\textsubscript{\tiny$\pm$2.3} & 35.3\textsubscript{\tiny$\pm$2.8} \\
  & \quad \mname{ExpeL} & \frozen & \frozen & 41.2\textsubscript{\tiny$\pm$2.4} & 26.9\textsubscript{\tiny$\pm$2.7} & 25.0\textsubscript{\tiny$\pm$2.2} & 3.3\textsubscript{\tiny$\pm$2.4} & 38.1\textsubscript{\tiny$\pm$2.0} & 22.5\textsubscript{\tiny$\pm$2.4} \\
  & \quad \mname{GEPA} & \frozen & \frozen & 44.8\textsubscript{\tiny$\pm$2.3} & 30.8\textsubscript{\tiny$\pm$2.9} & 26.5\textsubscript{\tiny$\pm$3.0} & 8.3\textsubscript{\tiny$\pm$3.6} & 41.3\textsubscript{\tiny$\pm$2.0} & 26.6\textsubscript{\tiny$\pm$2.5} \\
  & \quad \mname{EvoMedAgent} & \frozen & \frozen & 34.0\textsubscript{\tiny$\pm$2.6} & 26.9\textsubscript{\tiny$\pm$2.7} & 24.8\textsubscript{\tiny$\pm$4.0} & 13.3\textsubscript{\tiny$\pm$4.3} & 32.3\textsubscript{\tiny$\pm$2.3} & 24.4\textsubscript{\tiny$\pm$2.5} \\
  & \quad \mname{SkillX} & \frozen & \frozen & 38.2\textsubscript{\tiny$\pm$2.6} & 30.8\textsubscript{\tiny$\pm$2.8} & 25.8\textsubscript{\tiny$\pm$4.0} & 13.3\textsubscript{\tiny$\pm$4.3} & 35.9\textsubscript{\tiny$\pm$2.3} & 27.5\textsubscript{\tiny$\pm$2.5} \\
  & \quad \mname{GRASP} & \frozen & \frozen & 37.3\textsubscript{\tiny$\pm$2.5} & 28.1\textsubscript{\tiny$\pm$2.7} & 31.0\textsubscript{\tiny$\pm$4.6} & 18.3\textsubscript{\tiny$\pm$4.9} & 36.1\textsubscript{\tiny$\pm$2.2} & 26.2\textsubscript{\tiny$\pm$2.5} \\
  & \quad \mname{Fixed Improver} & \frozen & \frozen & 49.8\textsubscript{\tiny$\pm$1.8} & 34.3\textsubscript{\tiny$\pm$2.3} & 39.3\textsubscript{\tiny$\pm$2.2} & 21.6\textsubscript{\tiny$\pm$3.0} & 47.9\textsubscript{\tiny$\pm$1.6} & 31.9\textsubscript{\tiny$\pm$2.0} \\
  & \quad \mname{Fixed Improver (GPT)} & \frozen & \frozen & \hlsecond{79.5\textsubscript{\tiny$\pm$1.6}} & \hlsecond{72.7\textsubscript{\tiny$\pm$2.1}} & \hlsecond{68.2\textsubscript{\tiny$\pm$3.7}} & \hlsecond{55.8\textsubscript{\tiny$\pm$4.9}} & \hlsecond{77.4\textsubscript{\tiny$\pm$1.5}} & \hlsecond{69.5\textsubscript{\tiny$\pm$2.1}} \\
  & \quad \mname{SFT Improver} & \trained & \frozen & 70.4\textsubscript{\tiny$\pm$1.9} & 61.0\textsubscript{\tiny$\pm$2.3} & 58.1\textsubscript{\tiny$\pm$3.6} & 46.2\textsubscript{\tiny$\pm$4.4} & 68.1\textsubscript{\tiny$\pm$1.7} & 58.3\textsubscript{\tiny$\pm$2.2} \\
  & \quad \mname{GRPO Improver} & \trained & \frozen & 73.3\textsubscript{\tiny$\pm$1.9} & 64.5\textsubscript{\tiny$\pm$2.4} & 58.8\textsubscript{\tiny$\pm$3.8} & 45.0\textsubscript{\tiny$\pm$5.0} & 70.6\textsubscript{\tiny$\pm$1.7} & 60.8\textsubscript{\tiny$\pm$2.3} \\
 & \multicolumn{9}{c}{\cellcolor{catcolor}\textit{Weight-Level Self-Improvement}} \\
  & \quad \mname{STaR} & -- & \trained & 42.5\textsubscript{\tiny$\pm$2.6} & 34.2\textsubscript{\tiny$\pm$2.8} & 31.5\textsubscript{\tiny$\pm$4.7} & 21.7\textsubscript{\tiny$\pm$5.2} & 40.4\textsubscript{\tiny$\pm$2.4} & 31.9\textsubscript{\tiny$\pm$2.6} \\
  & \quad \mname{GRPO Solver} & -- & \trained & 74.1\textsubscript{\tiny$\pm$2.3} & 66.2\textsubscript{\tiny$\pm$2.9} & 50.6\textsubscript{\tiny$\pm$4.8} & 36.7\textsubscript{\tiny$\pm$6.1} & 69.7\textsubscript{\tiny$\pm$2.1} & 60.6\textsubscript{\tiny$\pm$2.8} \\
 & \multicolumn{9}{c}{\cellcolor{catcolor}\textit{Ours}} \\
  & \quad \method{} & \trained & \trained & \hlfirst{96.3\textsubscript{\tiny$\pm$0.9}} & \hlfirst{94.5\textsubscript{\tiny$\pm$1.4}} & \hlfirst{73.7\textsubscript{\tiny$\pm$3.5}} & \hlfirst{65.2\textsubscript{\tiny$\pm$4.7}} & \hlfirst{92.0\textsubscript{\tiny$\pm$1.2}} & \hlfirst{89.0\textsubscript{\tiny$\pm$1.6}} \\
\Xhline{1.2pt}
\end{tabularx}
\vspace{-0.3cm}
\end{table*}

\section{Additional Analyses}\label{app:analyses}
\paragraph{Without outcome feedback at test time.}
Table~\ref{tab:nofeedback} compares every method on \llmname{Qwen3.5-9B} and MedAgentBench with and without the outcome feedback of the test batch. Without the feedback, \method{} stays best in every column and within 0.6 points of its overall Outcome with the feedback. Two parts of \method{} make this possible. (1)~Its trained \improver{} can tell from the trajectories alone where the attempts went wrong and write the harness that fixes them. (2)~On most workflows seen in training, memory supplies the skill or tool stored during training, which needs no feedback.
\begin{table*}[t!]
\centering
\footnotesize
\setlength{\tabcolsep}{2.2pt}
\renewcommand{\arraystretch}{1}
\caption{\textbf{Results with and without Outcome Feedback at Test Time} on MedAgentBench. \method{} is best in every column, with and without the feedback. Outcome in \%. \emph{With} repeats Table~\ref{tab:main}. \emph{Without} gives every method that writes a harness or reflects at test time the same attempts of the test batch without their outcome feedback, and $\Delta$ is the change. \textsuperscript{\textdagger} marks the methods that use no outcome feedback at test time, whose results do not change. \hlfirst{Best} and \hlsecond{second best} per column.}
\label{tab:nofeedback}
\vspace{-3pt}
\begin{tabularx}{\textwidth}{l|YYY|YYY!{\vrule width 0.8pt}YYY}
\Xhline{1.2pt}
\rowcolor{headcolor}
 & \multicolumn{3}{c|}{\hgroup{In-Domain}} & \multicolumn{3}{c!{\vrule width 0.8pt}}{\hgroup{Out-of-Domain}} & \multicolumn{3}{c}{\hgroup{Overall}} \\
\rowcolor{headcolor}
\quad\hgroup{Method} & \hcol{With} & \hcol{Without} & \hcol{$\Delta$} & \hcol{With} & \hcol{Without} & \hcol{$\Delta$} & \hcol{With} & \hcol{Without} & \hcol{$\Delta$} \\
\Xhline{1pt}
\multicolumn{10}{c}{\cellcolor{catcolor}\textit{No Self-Improvement}} \\
\quad \mname{Base Agent}\textsuperscript{\textdagger} & 40.8 & 40.8 & \textcolor{black!40}{0.0} & 20.0 & 20.0 & \textcolor{black!40}{0.0} & 36.9 & 36.9 & \textcolor{black!40}{0.0} \\
\multicolumn{10}{c}{\cellcolor{catcolor}\textit{Context-Level Self-Improvement}} \\
\quad \mname{Reflexion} & 51.2 & 41.2 & $-$10.0 & 31.7 & 21.7 & $-$10.0 & 47.5 & 37.5 & $-$10.0 \\
\quad \mname{ExpeL}\textsuperscript{\textdagger} & 36.5 & 36.5 & \textcolor{black!40}{0.0} & 13.3 & 13.3 & \textcolor{black!40}{0.0} & 32.2 & 32.2 & \textcolor{black!40}{0.0} \\
\quad \mname{GEPA}\textsuperscript{\textdagger} & 48.8 & 48.8 & \textcolor{black!40}{0.0} & 41.7 & 41.7 & \textcolor{black!40}{0.0} & 47.5 & 47.5 & \textcolor{black!40}{0.0} \\
\quad \mname{EvoMedAgent}\textsuperscript{\textdagger} & 40.4 & 40.4 & \textcolor{black!40}{0.0} & 13.3 & 13.3 & \textcolor{black!40}{0.0} & 35.3 & 35.3 & \textcolor{black!40}{0.0} \\
\quad \mname{SkillX}\textsuperscript{\textdagger} & 41.9 & 41.9 & \textcolor{black!40}{0.0} & 26.7 & 26.7 & \textcolor{black!40}{0.0} & 39.1 & 39.1 & \textcolor{black!40}{0.0} \\
\quad \mname{GRASP}\textsuperscript{\textdagger} & 54.6 & 54.6 & \textcolor{black!40}{0.0} & 31.7 & 31.7 & \textcolor{black!40}{0.0} & 50.3 & 50.3 & \textcolor{black!40}{0.0} \\
\quad \mname{Fixed Improver} & 49.5 & 45.7 & $-$3.8 & 18.3 & 16.0 & $-$2.3 & 43.6 & 40.1 & $-$3.5 \\
\quad \mname{Fixed Improver (GPT)} & 53.6 & 49.2 & $-$4.4 & 43.1 & 43.0 & $-$0.1 & 51.6 & 48.0 & $-$3.6 \\
\quad \mname{SFT Improver} & 56.3 & 53.3 & $-$3.0 & 49.0 & 42.4 & $-$6.6 & 54.9 & 51.2 & $-$3.7 \\
\quad \mname{GRPO Improver} & 63.1 & 60.8 & $-$2.3 & \hlsecond{52.7} & \hlsecond{54.4} & +1.7 & 61.2 & 59.6 & $-$1.6 \\
\multicolumn{10}{c}{\cellcolor{catcolor}\textit{Weight-Level Self-Improvement}} \\
\quad \mname{STaR}\textsuperscript{\textdagger} & 61.2 & 61.2 & \textcolor{black!40}{0.0} & 30.0 & 30.0 & \textcolor{black!40}{0.0} & 55.3 & 55.3 & \textcolor{black!40}{0.0} \\
\quad \mname{GRPO Solver}\textsuperscript{\textdagger} & \hlsecond{68.5} & \hlsecond{68.5} & \textcolor{black!40}{0.0} & 38.3 & 38.3 & \textcolor{black!40}{0.0} & \hlsecond{62.8} & \hlsecond{62.8} & \textcolor{black!40}{0.0} \\
\multicolumn{10}{c}{\cellcolor{catcolor}\textit{Ours}} \\
\quad \method{} & \hlfirst{89.5} & \hlfirst{89.4} & $-$0.1 & \hlfirst{66.1} & \hlfirst{63.0} & $-$3.1 & \hlfirst{85.1} & \hlfirst{84.5} & $-$0.6 \\
\Xhline{1.2pt}
\end{tabularx}
\end{table*}

\Needspace*{15\baselineskip}
\begin{wrapfigure}{r}{0.5\textwidth}
\centering
\vspace{-10pt}
\includegraphics[width=0.5\textwidth]{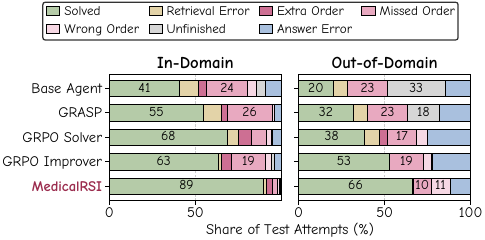}
\vspace{-4pt}
\caption{\textbf{Where Agents Fail.} \method{} solves the most test attempts and makes the fewest order errors. Each attempt counts under the first step it gets wrong, with the error types of Figure~\ref{fig:failures}.}
\label{fig:failures-bars}
\end{wrapfigure}
\paragraph{Where agents fail.}
Figure~\ref{fig:failures-bars} puts each test attempt on MedAgentBench with \llmname{Qwen3.5-9B} under the first step it fails~\citep{liu2026sciagentarena}, with the error types of Figure~\ref{fig:failures}: retrieving the needed records, placing the right orders, or reporting the answer. \method{} makes the fewest order errors on both training and unseen workflows, on 7.7\% and 21.8\% of the attempts, and misses the fewest orders. The \mname{GRPO Solver} reports a wrong answer on 5.4\% of the attempts on training workflows but on 25.0\% of those on unseen workflows, more than any other method.
\FinishWrap

\paragraph{Paired comparison.}
Every method attempts the same test tasks, so Table~\ref{tab:paired} compares \method{} with each baseline task by task on MedAgentBench with \llmname{Qwen3.5-9B}. \method{} leads every baseline on the training workflows, on the unseen workflows and overall, and no 95\% confidence interval includes zero. Its smallest lead is 13.4 points, over the \mname{GRPO Improver} on the 60 tasks of the unseen workflows, with an interval from 6.9 to 19.8.
\ifcsname NC@find@Y\endcsname\else\newcolumntype{Y}{>{\centering\arraybackslash}X}\fi
\begin{table*}[t!]
\centering
\footnotesize
\setlength{\tabcolsep}{2.2pt}
\renewcommand{\arraystretch}{1}
\caption{\textbf{Paired Comparison with Each Baseline} on MedAgentBench with \llmname{Qwen3.5-9B}. \method{} leads every baseline in every column, and no confidence interval includes zero. $\Delta$ is the outcome of \method{} minus that of the baseline in points. Its 95\% confidence interval comes from 10,000 bootstrap resamples of the per-task differences, and a sign-flip test gives $p < 0.001$ for every $\Delta$.}
\label{tab:paired}
\vspace{-3pt}
\begin{tabularx}{\textwidth}{l|cY|cY!{\vrule width 0.8pt}cY}
\Xhline{1.2pt}
\rowcolor{headcolor}
 & \multicolumn{2}{c|}{\hgroup{In-Domain}} & \multicolumn{2}{c!{\vrule width 0.8pt}}{\hgroup{Out-of-Domain}} & \multicolumn{2}{c}{\hgroup{Overall}} \\
\rowcolor{headcolor}
\quad\hgroup{Baseline} & \hcol{$\Delta$} & \hcol{95\% CI} & \hcol{$\Delta$} & \hcol{95\% CI} & \hcol{$\Delta$} & \hcol{95\% CI} \\
\Xhline{1pt}
\multicolumn{7}{c}{\cellcolor{catcolor}\textit{No Self-Improvement}} \\
\quad \mname{Base Agent} & +48.7 & [42.1, 55.3] & +46.1 & [35.3, 56.8] & +48.2 & [42.7, 53.9] \\
\multicolumn{7}{c}{\cellcolor{catcolor}\textit{Context-Level Self-Improvement}} \\
\quad \mname{Reflexion} & +38.3 & [31.5, 44.9] & +34.5 & [23.9, 45.3] & +37.6 & [31.8, 43.3] \\
\quad \mname{ExpeL} & +53.0 & [46.2, 59.6] & +52.8 & [41.8, 63.9] & +52.9 & [47.0, 58.6] \\
\quad \mname{GEPA} & +40.6 & [33.8, 47.3] & +24.5 & [13.9, 35.9] & +37.6 & [31.6, 43.6] \\
\quad \mname{EvoMedAgent} & +49.1 & [42.2, 55.8] & +52.8 & [41.4, 64.0] & +49.8 & [43.9, 55.6] \\
\quad \mname{SkillX} & +47.6 & [40.6, 54.2] & +39.5 & [29.0, 50.2] & +46.1 & [40.1, 52.0] \\
\quad \mname{GRASP} & +34.9 & [28.5, 41.5] & +34.5 & [24.1, 45.2] & +34.8 & [29.1, 40.5] \\
\quad \mname{Fixed Improver} & +40.0 & [34.2, 45.8] & +47.8 & [38.1, 57.4] & +41.5 & [36.3, 46.6] \\
\quad \mname{Fixed Improver (GPT)} & +35.9 & [30.2, 41.8] & +23.0 & [14.3, 32.3] & +33.5 & [28.6, 38.5] \\
\quad \mname{SFT Improver} & +33.2 & [27.9, 38.6] & +17.2 & [9.3, 25.4] & +30.2 & [25.5, 34.9] \\
\quad \mname{GRPO Improver} & +26.4 & [21.7, 31.2] & +13.4 & [6.9, 19.8] & +24.0 & [19.9, 28.2] \\
\multicolumn{7}{c}{\cellcolor{catcolor}\textit{Weight-Level Self-Improvement}} \\
\quad \mname{STaR} & +28.3 & [21.9, 34.6] & +36.1 & [26.0, 46.4] & +29.8 & [24.5, 35.2] \\
\quad \mname{GRPO Solver} & +21.0 & [15.5, 26.9] & +27.8 & [16.9, 38.9] & +22.3 & [17.3, 27.3] \\
\Xhline{1.2pt}
\end{tabularx}
\end{table*}

\paragraph{Writing from training failures.}
In Table~\ref{tab:main}, each harness is written from the failures on the test tasks of its workflow and scored on the same tasks, as in Harness-R1~\citep{shao2026harnessr1}. Table~\ref{tab:train} instead writes every harness from the failures on the 20 training tasks of each workflow and scores it only on the test tasks. No training round uses the training tasks of the six unseen workflows. Every method changes by at most 1.2 points overall, and \method{} stays far ahead of the \mname{GRPO Improver}, with 85.0\% against 61.7\% overall and 65.0\% against 55.1\% on the unseen workflows.
\ifcsname NC@find@Y\endcsname\else\newcolumntype{Y}{>{\centering\arraybackslash}X}\fi
\begin{table*}[t!]
\centering
\footnotesize
\setlength{\tabcolsep}{2.2pt}
\renewcommand{\arraystretch}{1}
\caption{\textbf{Harnesses Written from Training Failures} on MedAgentBench with \llmname{Qwen3.5-9B}. Every method stays within 1.2 points of its overall result in Table~\ref{tab:main}, and \method{} remains best in every column. Outcome in \%. \emph{Test Batch} writes each harness from the failures on the test tasks of its workflow, as in Table~\ref{tab:main}. \emph{Training} writes it from the failures on the 20 training tasks of the workflow, so no test task is seen before scoring. Where \method{} has a stored skill or tool, both use it. \hlfirst{Best} and \hlsecond{second best} per column.}
\label{tab:train}
\vspace{-3pt}
\begin{tabularx}{\textwidth}{l|YY|YY!{\vrule width 0.8pt}YY}
\Xhline{1.2pt}
\rowcolor{headcolor}
 & \multicolumn{2}{c|}{\hgroup{In-Domain}} & \multicolumn{2}{c!{\vrule width 0.8pt}}{\hgroup{Out-of-Domain}} & \multicolumn{2}{c}{\hgroup{Overall}} \\
\rowcolor{headcolor}
\quad\hgroup{Method} & \hcol{Test Batch} & \hcol{Training} & \hcol{Test Batch} & \hcol{Training} & \hcol{Test Batch} & \hcol{Training} \\
\Xhline{1pt}
\multicolumn{7}{c}{\cellcolor{catcolor}\textit{Context-Level Self-Improvement}} \\
\quad \mname{Fixed Improver} & 49.5 & 49.3 & 18.3 & 18.3 & 43.6 & 43.5 \\
\quad \mname{Fixed Improver (GPT)} & 53.6 & 51.3 & 43.1 & 46.9 & 51.6 & 50.4 \\
\quad \mname{SFT Improver} & 56.3 & 56.4 & 49.0 & 51.9 & 54.9 & 55.5 \\
\quad \mname{GRPO Improver} & \hlsecond{63.1} & \hlsecond{63.2} & \hlsecond{52.7} & \hlsecond{55.1} & \hlsecond{61.2} & \hlsecond{61.7} \\
\multicolumn{7}{c}{\cellcolor{catcolor}\textit{Ours}} \\
\quad \method{} & \hlfirst{89.5} & \hlfirst{89.6} & \hlfirst{66.1} & \hlfirst{65.0} & \hlfirst{85.1} & \hlfirst{85.0} \\
\Xhline{1.2pt}
\end{tabularx}
\end{table*}

\paragraph{Additional unseen workflows.}
The six unseen workflows of MedAgentBench give 60 test tasks. To test on more of them, we write fifteen further workflows of the same kinds on the same patient records, with graders in the same style and 30 tasks each. Five read or compute a value from the record, four order tests after a laboratory result, three review medication orders, and three place a referral or record vital signs. No method trains on these workflows. Each is tested with the protocol of Table~\ref{tab:main}, ten tasks at a time. On the 450 tasks, \method{} reaches an Outcome of 74.1\% against 59.8\% for the \mname{GRPO Improver} and leads every method on both metrics (Table~\ref{tab:unseen15}).
\providecommand{\frozen}{\textcolor{myblue}{\faSnowflake}}
\providecommand{\trained}{\textcolor{pyorange}{\faFire}}
\ifcsname NC@find@Y\endcsname\else\newcolumntype{Y}{>{\centering\arraybackslash}X}\fi
\begin{table*}[t!]
\centering
\footnotesize
\setlength{\tabcolsep}{2.2pt}
\renewcommand{\arraystretch}{1}
\caption{\textbf{Fifteen Additional Unseen Workflows} on MedAgentBench with \llmname{Qwen3.5-9B}. \method{} is best on both metrics, and no confidence interval includes zero. The fifteen workflows are new, with 30 tasks each, and no method trains on them. Process and Outcome are in \% with bootstrap standard deviations. $\Delta$ is the result of \method{} minus that of the baseline in points. Its 95\% confidence interval comes from 10,000 bootstrap resamples of the per-task differences, and a sign-flip test gives $p < 0.001$ for every $\Delta$. \frozen{} frozen, \trained{} trained. \hlfirst{Best} and \hlsecond{second best} per column.}
\label{tab:unseen15}
\vspace{-3pt}
\begin{tabularx}{\textwidth}{l|cc|YcY!{\vrule width 0.8pt}YcY}
\Xhline{1.2pt}
\rowcolor{headcolor}
 & & & \multicolumn{3}{c!{\vrule width 0.8pt}}{\hgroup{Process}} & \multicolumn{3}{c}{\hgroup{Outcome}} \\
\rowcolor{headcolor}
\quad\hgroup{Method} & \hcol{Improver} & \hcol{Solver} & \hcol{Mean} & \hcol{$\Delta$} & \hcol{95\% CI} & \hcol{Mean} & \hcol{$\Delta$} & \hcol{95\% CI} \\
\Xhline{1pt}
\multicolumn{9}{c}{\cellcolor{catcolor}\textit{No Self-Improvement}} \\
\quad \mname{Base Agent} & -- & \frozen & 56.1\textsubscript{\tiny$\pm$1.8} & +27.3 & [23.8, 30.7] & 38.4\textsubscript{\tiny$\pm$2.4} & +35.6 & [31.1, 40.4] \\
\multicolumn{9}{c}{\cellcolor{catcolor}\textit{Context-Level Self-Improvement}} \\
\quad \mname{Reflexion} & \frozen & \frozen & 66.1\textsubscript{\tiny$\pm$1.6} & +17.4 & [14.3, 20.5] & 48.2\textsubscript{\tiny$\pm$2.3} & +25.8 & [21.1, 30.6] \\
\quad \mname{Fixed Improver} & \frozen & \frozen & 62.5\textsubscript{\tiny$\pm$1.4} & +20.9 & [18.2, 23.8] & 42.0\textsubscript{\tiny$\pm$2.1} & +32.1 & [27.8, 36.3] \\
\quad \mname{Fixed Improver (GPT)} & \frozen & \frozen & 73.5\textsubscript{\tiny$\pm$1.4} & +9.9 & [7.4, 12.4] & 59.7\textsubscript{\tiny$\pm$2.1} & +14.4 & [10.7, 18.2] \\
\quad \mname{SFT Improver} & \trained & \frozen & 70.3\textsubscript{\tiny$\pm$1.1} & +13.1 & [11.1, 15.2] & 53.2\textsubscript{\tiny$\pm$1.8} & +20.9 & [17.7, 24.1] \\
\quad \mname{GRPO Improver} & \trained & \frozen & \hlsecond{74.8\textsubscript{\tiny$\pm$1.1}} & +8.6 & [6.6, 10.6] & \hlsecond{59.8\textsubscript{\tiny$\pm$1.8}} & +14.2 & [11.1, 17.4] \\
\multicolumn{9}{c}{\cellcolor{catcolor}\textit{Weight-Level Self-Improvement}} \\
\quad \mname{STaR} & -- & \trained & 69.9\textsubscript{\tiny$\pm$1.6} & +13.6 & [10.8, 16.5] & 54.4\textsubscript{\tiny$\pm$2.4} & +19.6 & [15.3, 24.0] \\
\quad \mname{GRPO Solver} & -- & \trained & 69.3\textsubscript{\tiny$\pm$1.5} & +14.1 & [11.2, 17.0] & 51.6\textsubscript{\tiny$\pm$2.4} & +22.5 & [18.1, 27.1] \\
\multicolumn{9}{c}{\cellcolor{catcolor}\textit{Ours}} \\
\quad \method{} & \trained & \trained & \hlfirst{83.4\textsubscript{\tiny$\pm$1.1}} & -- & -- & \hlfirst{74.1\textsubscript{\tiny$\pm$1.7}} & -- & -- \\
\Xhline{1.2pt}
\end{tabularx}
\end{table*}

\paragraph{Using the candidates without a grader.}
Table~\ref{tab:main} reports the mean over the $K$ candidates that a method writes for a workflow. This mean is the expected outcome of deploying one candidate chosen at random, which needs no grader and attempts each task once. Table~\ref{tab:deploy} adds two other rules that need no grader, a vote over all candidates on each task and a pick of the candidate that agrees most with the majority. \method{} stays best under every rule. On the unseen workflows, it leads the \mname{GRPO Improver} by 13.4 points at random, by 12.5 with the vote and by 10.5 with the pick.
\ifcsname NC@find@Y\endcsname\else\newcolumntype{Y}{>{\centering\arraybackslash}X}\fi
\begin{table*}[t!]
\centering
\footnotesize
\setlength{\tabcolsep}{2.2pt}
\renewcommand{\arraystretch}{1}
\caption{\textbf{Using the Candidates without a Grader} on MedAgentBench with \llmname{Qwen3.5-9B}. \method{} is best under every rule. Outcome in \%. \emph{Random} deploys one candidate chosen at random and repeats Table~\ref{tab:main}. \emph{Vote} runs every candidate on each task and keeps the most common result, the orders placed and the final answer. \emph{Pick} keeps the candidate whose results agree most with this majority on half of the tasks of a workflow and uses it on the other half, averaged over 200 random splits. Where \method{} has a stored skill or tool, every rule uses it. \hlfirst{Best} and \hlsecond{second best} per column.}
\label{tab:deploy}
\vspace{-3pt}
\begin{tabularx}{\textwidth}{l|YYY|YYY!{\vrule width 0.8pt}YYY}
\Xhline{1.2pt}
\rowcolor{headcolor}
 & \multicolumn{3}{c|}{\hgroup{In-Domain}} & \multicolumn{3}{c!{\vrule width 0.8pt}}{\hgroup{Out-of-Domain}} & \multicolumn{3}{c}{\hgroup{Overall}} \\
\rowcolor{headcolor}
\quad\hgroup{Method} & \hcol{Random} & \hcol{Vote} & \hcol{Pick} & \hcol{Random} & \hcol{Vote} & \hcol{Pick} & \hcol{Random} & \hcol{Vote} & \hcol{Pick} \\
\Xhline{1pt}
\multicolumn{10}{c}{\cellcolor{catcolor}\textit{Context-Level Self-Improvement}} \\
\quad \mname{Fixed Improver} & 49.5 & 46.7 & 47.1 & 18.3 & 11.7 & 11.3 & 43.6 & 40.1 & 40.4 \\
\quad \mname{Fixed Improver (GPT)} & 53.6 & 51.9 & 52.5 & 43.1 & 49.7 & 47.7 & 51.6 & 51.5 & 51.6 \\
\quad \mname{SFT Improver} & 56.3 & 53.1 & 55.2 & 49.0 & 58.1 & 58.3 & 54.9 & 54.0 & 55.8 \\
\quad \mname{GRPO Improver} & \hlsecond{63.1} & \hlsecond{64.2} & \hlsecond{63.0} & \hlsecond{52.7} & \hlsecond{60.8} & \hlsecond{59.1} & \hlsecond{61.2} & \hlsecond{63.6} & \hlsecond{62.3} \\
\multicolumn{10}{c}{\cellcolor{catcolor}\textit{Ours}} \\
\quad \method{} & \hlfirst{89.5} & \hlfirst{89.6} & \hlfirst{89.5} & \hlfirst{66.1} & \hlfirst{73.3} & \hlfirst{69.6} & \hlfirst{85.1} & \hlfirst{86.6} & \hlfirst{85.8} \\
\Xhline{1.2pt}
\end{tabularx}
\end{table*}

\paragraph{Picking and missed orders.}
A rule that looks for agreement could favor candidates that never place an order, since attempts that do nothing agree more easily than orders that differ in a detail. In Table~\ref{tab:inaction}, the baselines place no order at all in 28.6\% to 57.9\% of the workflows that need one when they use a random candidate, and the pick raises this share for the \mname{Fixed Improver} and the \mname{GRPO Improver}. For \method{}, this share is only 0.7\%, and the pick leaves its outcome on the tasks that need an order almost unchanged, 82.7\% against 82.8\% at random. A missed order can delay needed care, so this matters in clinical use.
\ifcsname NC@find@Y\endcsname\else\newcolumntype{Y}{>{\centering\arraybackslash}X}\fi
\begin{table*}[t!]
\centering
\footnotesize
\setlength{\tabcolsep}{2.2pt}
\renewcommand{\arraystretch}{1}
\caption{\textbf{Candidates That Never Place an Order} on MedAgentBench with \llmname{Qwen3.5-9B}. Almost no candidate of \method{} avoids orders altogether, and picking one without a grader keeps its outcome on the tasks that need an order. In \%. \emph{Never Order} is how often the candidate used for a workflow in which some task needs an order places no order on any of its tasks. \emph{Order Needed} and \emph{No Order Needed} give the outcome on the 110 test tasks that need at least one order and on the 143 that need none. Each group compares a random candidate with the pick of Table~\ref{tab:deploy}. \hlfirst{Best} and \hlsecond{second best} per column.}
\label{tab:inaction}
\vspace{-3pt}
\begin{tabularx}{\textwidth}{l|YY!{\vrule width 0.8pt}YY|YY}
\Xhline{1.2pt}
\rowcolor{headcolor}
 & \multicolumn{2}{c!{\vrule width 0.8pt}}{\hgroup{Never Order $\downarrow$}} & \multicolumn{2}{c|}{\hgroup{Order Needed $\uparrow$}} & \multicolumn{2}{c}{\hgroup{No Order Needed $\uparrow$}} \\
\rowcolor{headcolor}
\quad\hgroup{Method} & \hcol{Random} & \hcol{Pick} & \hcol{Random} & \hcol{Pick} & \hcol{Random} & \hcol{Pick} \\
\Xhline{1pt}
\multicolumn{7}{c}{\cellcolor{catcolor}\textit{Context-Level Self-Improvement}} \\
\quad \mname{Fixed Improver} & 57.9 & 71.6 & 23.0 & 19.6 & 67.3 & 63.3 \\
\quad \mname{Fixed Improver (GPT)} & 44.7 & 40.4 & 29.8 & 30.7 & 77.2 & 75.6 \\
\quad \mname{SFT Improver} & 43.4 & 40.0 & 29.5 & 27.4 & 79.8 & 76.4 \\
\quad \mname{GRPO Improver} & \hlsecond{28.6} & \hlsecond{34.4} & \hlsecond{42.4} & \hlsecond{36.3} & \hlsecond{82.0} & \hlsecond{82.7} \\
\multicolumn{7}{c}{\cellcolor{catcolor}\textit{Ours}} \\
\quad \method{} & \hlfirst{0.7} & \hlfirst{0.1} & \hlfirst{82.8} & \hlfirst{82.7} & \hlfirst{91.3} & \hlfirst{91.2} \\
\Xhline{1.2pt}
\end{tabularx}
\end{table*}

\begin{figure*}[t]
\centering
\includegraphics[width=\textwidth]{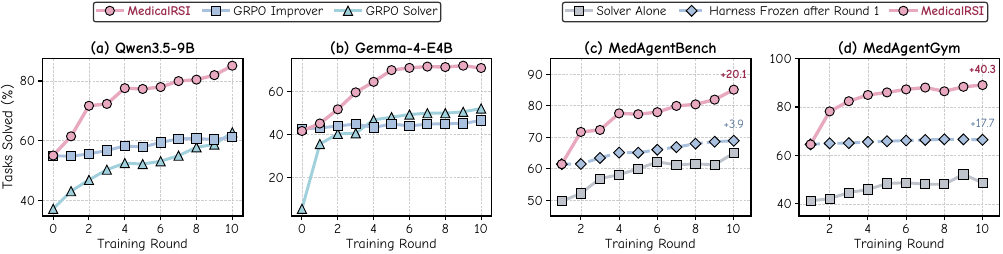}
\caption{\textbf{Training Curves and the Frozen Harness.} (a, b) Training both roles together gains far more than training either one. \method{} ends far above the \mname{GRPO Improver}, which trains only the improver, and the \mname{GRPO Solver}, which trains only the model. (c, d) The harness must evolve with the model. A harness frozen after round 1 adds less and less as the model learns, while the harness of each round keeps adding the most. Numbers above the last points give the final gain over the \solver{} alone.}
\label{fig:curves-appx}
\end{figure*}

\paragraph{Cost of a new workflow.}
Figure~\ref{fig:cost} counts the tokens of one attempt. On a workflow without a stored skill or tool, \method{} also spends tokens once, on a first pass of the \solver{} over the tasks of the workflow and on writing the harness from its failures. Table~\ref{tab:cost} adds this one-time cost. Writing one harness takes about 10k tokens and the first pass about 127k. With both counted, \method{} spends fewer tokens in total than the \mname{GRPO Solver} after about 140 tasks of the workflow. Per solved task on unseen workflows, it spends 20.8k tokens against 40.4k for the \mname{GRPO Solver}, since it solves more tasks with fewer tokens per attempt.
\ifcsname NC@find@Y\endcsname\else\newcolumntype{Y}{>{\centering\arraybackslash}X}\fi
\begin{table}[t!]
\centering
\footnotesize
\setlength{\tabcolsep}{4pt}
\renewcommand{\arraystretch}{1}
\caption{\textbf{Cost of a New Workflow} on MedAgentBench with \llmname{Qwen3.5-9B}. Writing a harness costs few tokens, and per solved task on unseen workflows \method{} spends about half the tokens of the \mname{GRPO Solver}. Tokens are input plus output, counted as in Figure~\ref{fig:cost}. The 11 workflows are those without a stored skill or tool, 6 unseen and 5 seen in training, for which the \improver{} writes the harness at test time. The one-time cost counts the first pass of the \solver{} over the tasks of a workflow, whose failures the harness is written from, since these tasks are attempted again with the harness. Tokens per solved task divide the tokens per task by the share of tasks solved.}
\label{tab:cost}
\vspace{-3pt}
\begin{tabularx}{0.8\textwidth}{l|YY}
\Xhline{1.2pt}
\rowcolor{headcolor}
\quad\hgroup{Tokens} & \hgroup{11 Workflows} & \hgroup{6 Unseen Workflows} \\
\Xhline{1pt}
\multicolumn{3}{c}{\cellcolor{catcolor}\textit{One-Time Cost of a Workflow}} \\
\quad First pass of the \solver{} on its 10 tasks & 127,024 & 138,505 \\
\quad Writing one harness & 10,409 & 12,066 \\
\quad Writing 16 candidates & 15,783 & 17,693 \\
\quad Running 16 candidates on the 10 tasks & 2,074,474 & 2,202,151 \\
\multicolumn{3}{c}{\cellcolor{catcolor}\textit{Tokens per Task}} \\
\quad \method{} & \textbf{12,965} & \textbf{13,763} \\
\quad \mname{GRPO Solver} & 13,947 & 15,483 \\
\quad \mname{Base Agent} & 21,544 & 24,835 \\
\multicolumn{3}{c}{\cellcolor{catcolor}\textit{Tokens per Solved Task}} \\
\quad \method{} & \textbf{16,138} & \textbf{20,808} \\
\quad \mname{GRPO Solver} & 22,235 & 40,389 \\
\quad \mname{Base Agent} & 56,424 & 124,176 \\
\multicolumn{3}{c}{\cellcolor{catcolor}\textit{Tasks Until \method{} Has Spent Fewer Tokens}} \\
\quad Against the \mname{GRPO Solver}, one harness & 140 & 88 \\
\quad Against the \mname{GRPO Solver}, 16 candidates run & 2,258 & 1,372 \\
\quad Against the \mname{Base Agent}, one harness & 16 & 14 \\
\quad Against the \mname{Base Agent}, 16 candidates run & 258 & 213 \\
\Xhline{1.2pt}
\end{tabularx}
\end{table}

\paragraph{Training curves.}
Figure~\ref{fig:curves-appx} follows the three training routes over the ten rounds (a, b) and the harness frozen after round 1 (c, d). \method{} and the \mname{GRPO Improver} start from the same warm start, and the \mname{GRPO Solver} from the \mname{Base Agent}. Training both roles together gains 30 points on \llmname{Qwen3.5-9B} and 29 on \llmname{Gemma-4-E4B}, most of it in the first five rounds, while training only the improver for a frozen solver gains 6 and 4 points. Training only the model rises steadily from the lower start of the \mname{Base Agent} but ends 22.3 and 18.9 points below \method{}. In panels (c, d), the harness frozen after round 1 adds less and less as the solver learns. In the last round, it adds 3.9 points over the \solver{} alone on MedAgentBench and 17.7 points on MedAgentGym, against 20.1 and 40.3 points for the harness that evolves with the model.

\FloatBarrier
\section{Case Studies}\label{app:cases}
Figure~\ref{fig:case-v3-task8} gives the full trajectories of the case in Figure~\ref{fig:case}. Figures~\ref{fig:case-v1-task6} to~\ref{fig:case-v3-task11} show more tasks that the \solver{} of \method{} on \llmname{Qwen3.5-9B} fails without a harness and solves with its harness, deployed as in Table~\ref{tab:main}: the skill that memory stored for the workflow during training where there is one, and otherwise one of the 16 skills written from the failures of the test batch. Each caption says which. Each figure shows the attempt without a harness, the steps of the skill and the attempt with the skill. Patient identifiers are masked and order payloads are shortened. Figure~\ref{fig:case-gym-task3} shows a MedAgentGym task that the \solver{} fails without a harness and solves with the tool that \method{} stored for its workflow during training. Appendix~\ref{app:skills} gives each skill and tool in full.

{\setlength{\intextsep}{6pt plus 2pt minus 2pt}
\begin{figure}[H]
\centering
\prerendered{case_app1}{\includegraphics{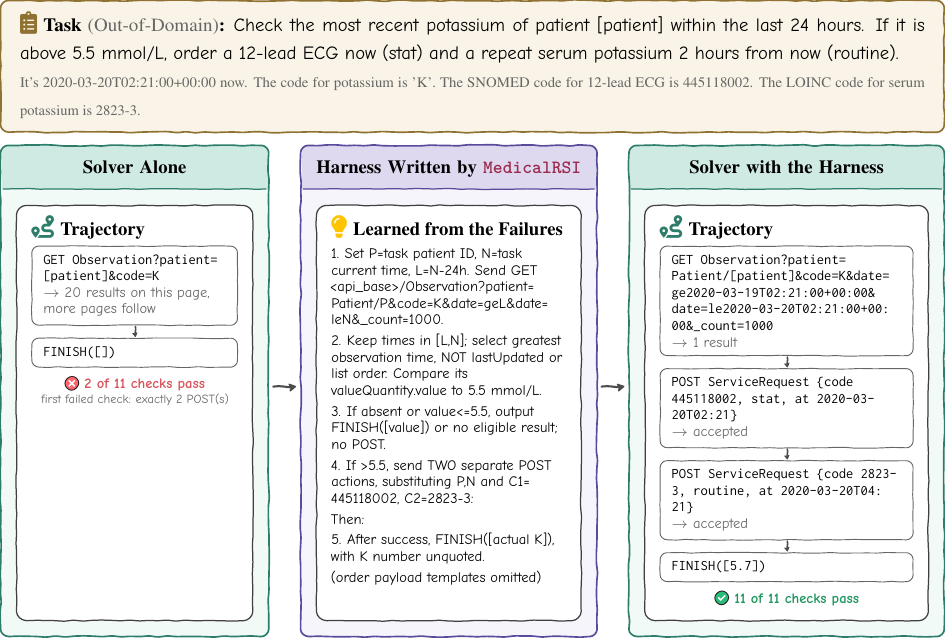}}
\vspace{-8pt}
\caption{\textbf{Potassium Check with Conditional ECG and Repeat Lab.} The full trajectories of the case in Figure~\ref{fig:case}. The harness is written from the failures of the test batch.}
\label{fig:case-v3-task8}
\end{figure}

\begin{figure}[H]
\centering
\prerendered{case_app2}{\includegraphics{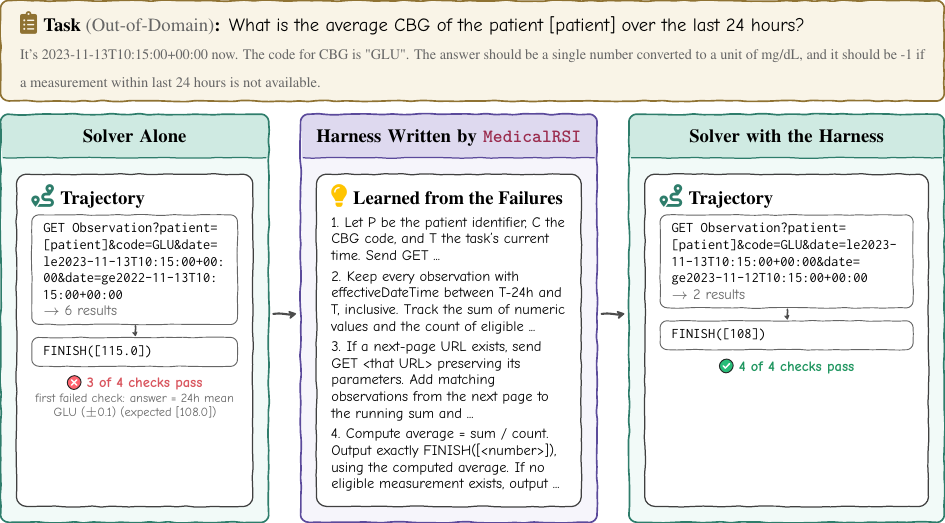}}
\vspace{-8pt}
\caption{\textbf{Average Glucose over the Last 24 Hours.} Without a harness, the \solver{} averages the glucose values of the past year instead of the past 24 hours. The harness sets the 24-hour window, and the \solver{} then reports the right average. It is written from the failures of the test batch.}
\label{fig:case-v1-task6}
\end{figure}

\begin{figure}[H]
\centering
\prerendered{case_app3}{\includegraphics{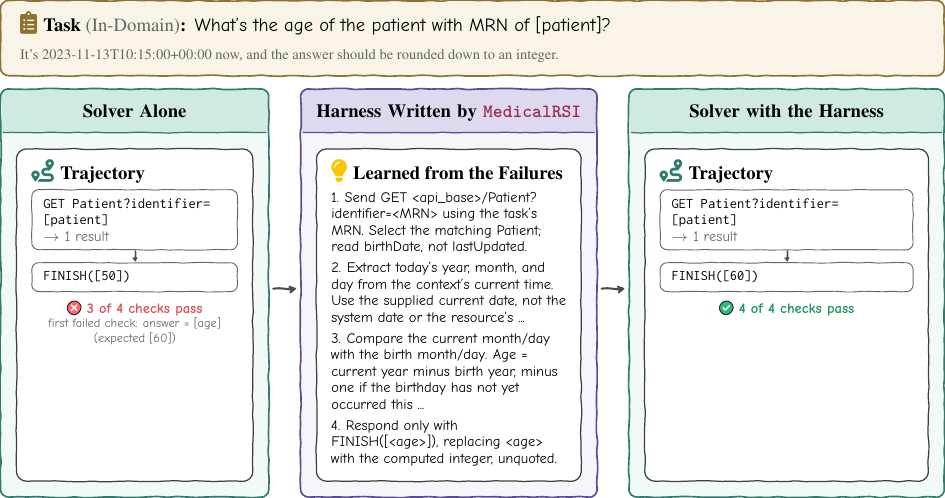}}
\vspace{-8pt}
\caption{\textbf{Patient Age.} Without a harness, the \solver{} reports a wrong age. The harness computes the age from the birth date and the current date given in the task, and the \solver{} then reports the right age. It is the one stored for this workflow during training.}
\label{fig:case-v1-task2}
\end{figure}

\begin{figure}[H]
\centering
\prerendered{case_app4}{\includegraphics{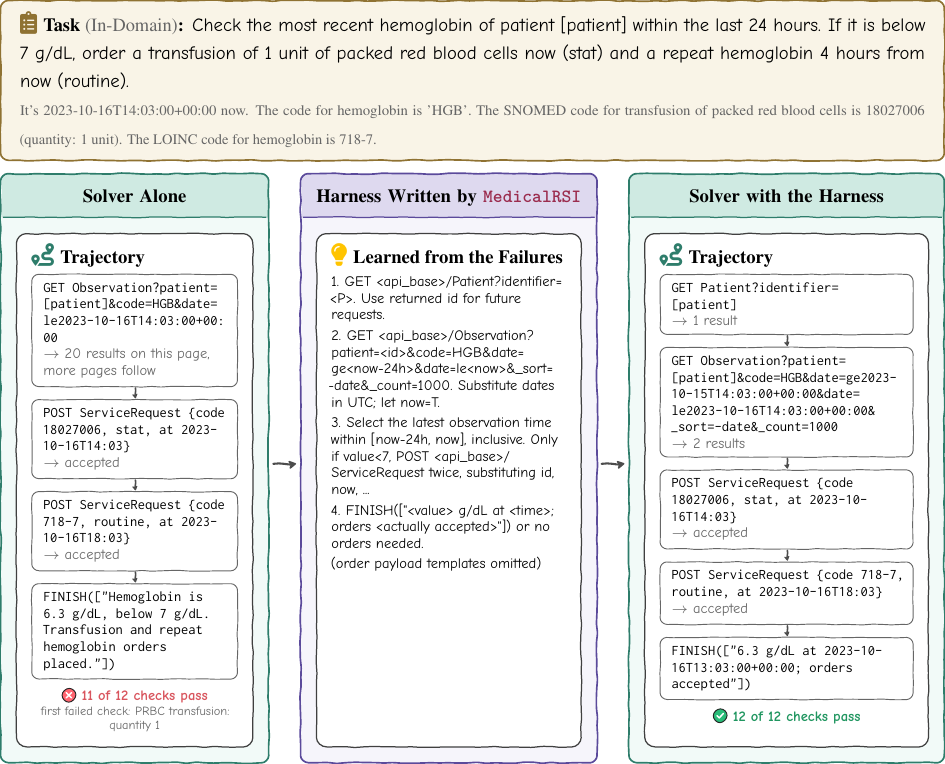}}
\vspace{-8pt}
\caption{\textbf{Hemoglobin Check with Conditional Transfusion.} Without a harness, the \solver{} orders the transfusion without its quantity of one unit. The harness gives the order templates, and the \solver{} then passes every check. It is the one stored for this workflow during training.}
\label{fig:case-v3-task7}
\end{figure}

\begin{figure}[H]
\centering
\prerendered{case_app5}{\includegraphics{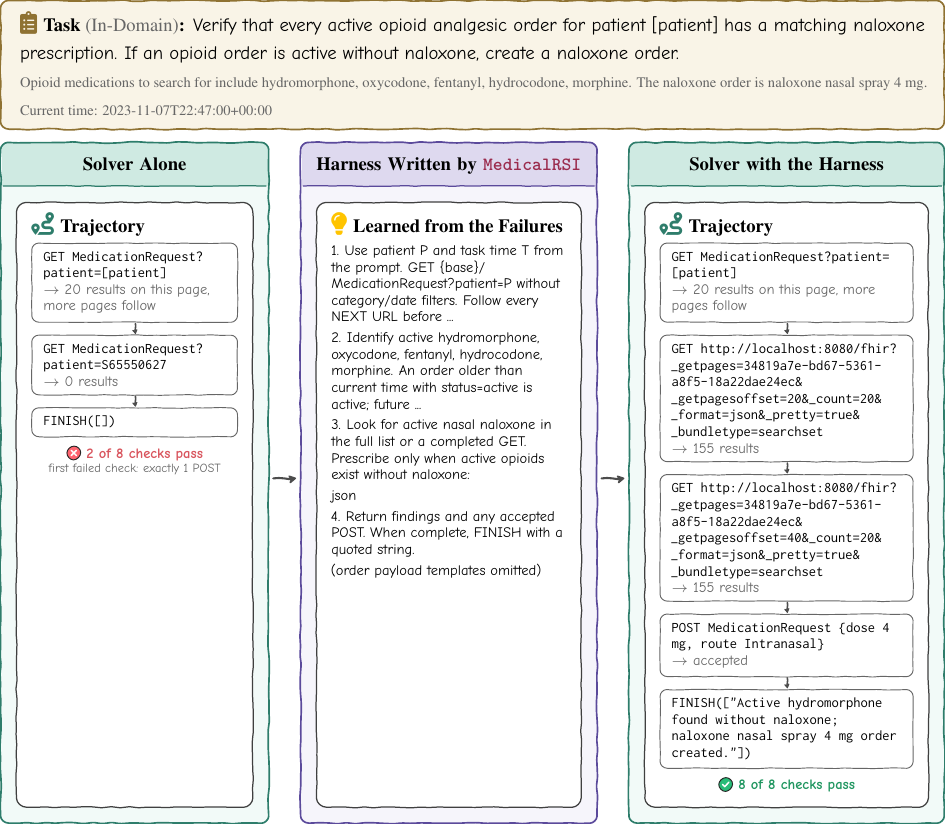}}
\vspace{-8pt}
\caption{\textbf{Naloxone for Active Opioid Orders.} Without a harness, the \solver{} reads only the first page of orders and places no naloxone order. The harness reads every page, and the \solver{} then finds the active opioid order without naloxone and orders it. It is the one stored for this workflow during training.}
\label{fig:case-v2-task8}
\end{figure}

\begin{figure}[H]
\centering
\prerendered{case_app6}{\includegraphics{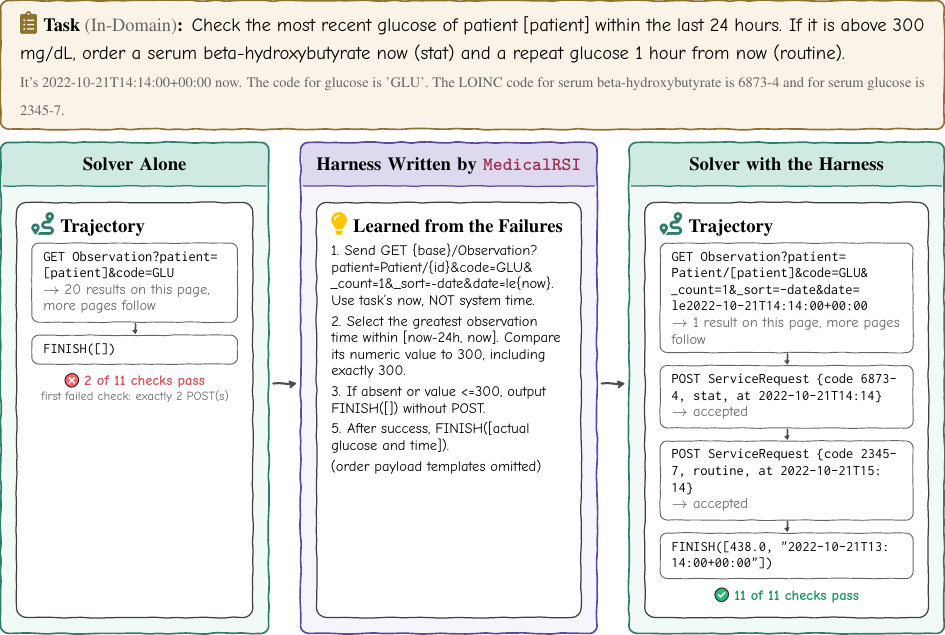}}
\vspace{-8pt}
\caption{\textbf{Glucose Check with Conditional Ketone and Repeat Lab.} Without a harness, the \solver{} stops after the first page of glucose values and places no order. The harness asks for the latest value directly, and the \solver{} then places both orders. It is the one stored for this workflow during training.}
\label{fig:case-v3-task11}
\end{figure}

\begin{figure}[H]
\centering
\prerendered{case_tool1}{\includegraphics{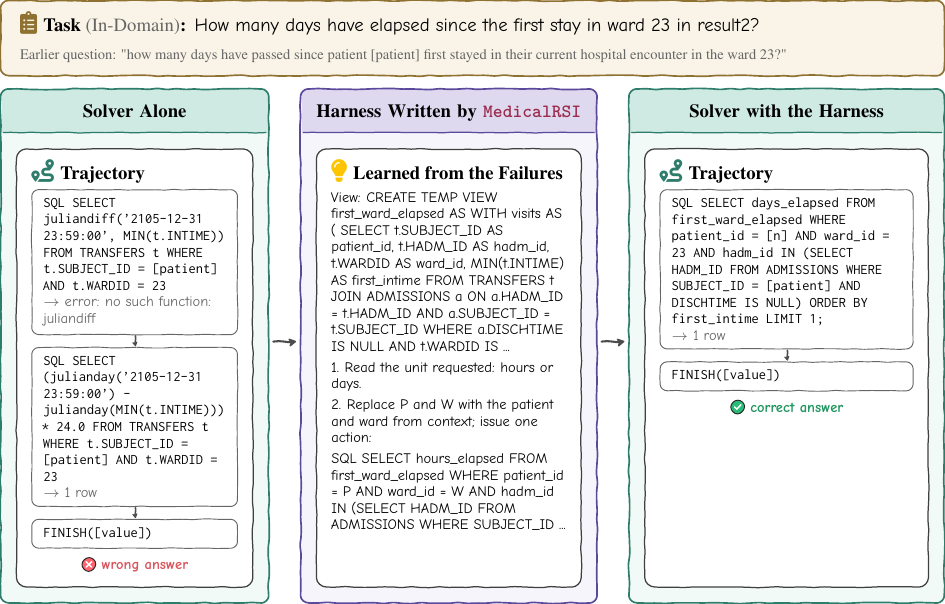}}
\vspace{-8pt}
\caption{\textbf{Time Since the First Ward Stay of the Current Admission (MedAgentGym).} Without a tool, the \solver{} counts hours over all admissions instead of days in the current one. The tool gives the days since the first stay in the ward during the current admission, and the \solver{} then answers with one query. It is the one stored for this workflow during training. Patient identifiers, dates and answer values are masked, and each query result is shown by its number of rows.}
\label{fig:case-gym-task3}
\end{figure}
}

\clearpage

\section{Skills and Tools Written by \method{}}\label{app:skills}
\paragraph{Skills on MedAgentBench.} The skills below are outputs of the trained model of \method{} on \llmname{Qwen3.5-9B}, deployed as in Table~\ref{tab:main}: the skill stored for the workflow during training where there is one, and otherwise one of the 16 written from the attempts of the test batch and their outcome feedback, without the reference answers. They are shown as written apart from masked patient identifiers. They are the skills of the cases above. Each title gives where the skill comes from and the success rate of the \solver{} alone and with that skill on the ten test tasks of the workflow.

\begin{skillbox}{Potassium Check with Conditional ECG and Repeat Lab (out-of-domain, written at test time): Solver alone 70\%, with~this~skill~90\%}{impcolor}
---
name: potassium_stat_ecg
description: "Use when asked to check potassium within 24 hours and order a stat ECG and routine repeat potassium only if above 5.5 mmol/L."
tags: [potassium, ECG, orders]
---
## When to use
- Latest K check with conditional ECG and repeat lab.
## Steps
1. Set P=task patient ID, N=task current time, L=N-24h. Send `GET <api_base>/Observation?patient=Patient/P&code=K&date=geL&date=leN&_count=1000`.
2. Keep times in [L,N]; select greatest observation time, NOT lastUpdated or list order. Compare its valueQuantity.value to 5.5 mmol/L.
3. If absent or value<=5.5, output `FINISH([value])` or no eligible result; no POST.
4. If >5.5, send TWO separate POST actions, substituting P,N and C1=445118002, C2=2823-3:
   `POST <api_base>/ServiceRequest`
   `{"resourceType":"ServiceRequest","status":"active","intent":"order","subject":{"reference":"Patient/P"},"authoredOn":"N","priority":"stat","code":{"coding":[{"system":"http://snomed.info/sct","code":"C1"}]},"occurrenceDateTime":"N"}`
   Then:
   `POST <api_base>/ServiceRequest`
   `{"resourceType":"ServiceRequest","status":"active","intent":"order","subject":{"reference":"Patient/P"},"authoredOn":"N","priority":"routine","code":{"coding":[{"system":"http://loinc.org","code":"C2"}]},"occurrenceDateTime":"N plus 2 hours"}`
5. After success, `FINISH([actual K])`, with K number unquoted.
## Pitfalls
- One plain action per turn; POST JSON on the next line with no trailing text.
- Do not repeat rejected POSTs or claim orders in FINISH. FINISH by turn 8.
\end{skillbox}

\begin{skillbox}{Average Glucose over the Last 24 Hours (out-of-domain, written at test time): Solver alone 60\%, with~this~skill~90\%}{impcolor}
---
name: cbg_average
description: "Use when asked for the average CBG or glucose over the last 24 hours in mg/dL."
tags: [clinical, glucose, retrieval]
---
## When to use
- The task asks for the average CBG over the last 24 hours.

## Steps
1. Let P be the patient identifier, C the CBG code, and T the task's current time. Send `GET <api_base>/Observation?patient=P&code=C&date=leT&date=geT-24h`.
2. Keep every observation with effectiveDateTime between T-24h and T, inclusive. Track the sum of numeric values and the count of eligible observations; ignore lastUpdated, flags, and values outside the window.
3. If a next-page URL exists, send `GET <that URL>` preserving its parameters. Add matching observations from the next page to the running sum and count; do not terminate after the first page.
4. Compute average = sum / count. Output exactly `FINISH([<number>])`, using the computed average. If no eligible measurement exists, output `FINISH([-1])`. This is read-only: no POST.

## Pitfalls
- An incomplete page or an out-of-window result is not proof of absence; follow every next link.
- Do not average only the first results or assume the newest value represents the average.
- One plain action per turn; no surrounding prose.
\end{skillbox}

\begin{skillbox}{Patient Age (in-domain, stored during training): Solver alone 40\%, with~this~skill~90\%}{impcolor}
---
name: patient_age
description: "Use when asked for a patient's age in completed years from their MRN and a supplied current date."
tags: [patient, age, demographics]
---
## When to use
- The task asks "What's the age" of a patient identified by MRN, rounded down to an integer.

## Steps
1. Send `GET <api_base>/Patient?identifier=<MRN>` using the task's MRN. Select the matching Patient; read `birthDate`, not `lastUpdated`.
2. Extract today's year, month, and day from the context's current time. Use the supplied current date, not the system date or the resource's `lastUpdated`.
3. Compare the current month/day with the birth month/day. Age = current year minus birth year, minus one if the birthday has not yet occurred this year; otherwise no subtraction. Ignore leap years: simply check whether today's month/day is earlier than the birth month/day.
4. Respond only with `FINISH([<age>])`, replacing `<age>` with the computed integer, unquoted.

## Pitfalls
- Do not copy the birth year or an unrelated number from the record as the age.
- Do not output a range of years, an incomplete list, or prose; the action must be exactly one JSON-compatible FINISH call.
- Do not query Observations, MedicationRequest, or other endpoints; age comes from Patient alone.
\end{skillbox}

\begin{skillbox}{Hemoglobin Check with Conditional Transfusion (in-domain, stored during training): Solver alone 50\%, with~this~skill~100\%}{impcolor}
---
name: hemoglobin_transfusion
description: "Use when asked to check hemoglobin within 24 hours and conditionally order PRBC transfusion and repeat HGB."
tags: [hemoglobin, transfusion, orders]
---
## When to use
- Latest HGB in last 24 hours; if below 7 g/dL, transfuse and repeat.
## Steps
1. GET <api_base>/Patient?identifier=<P>. Use returned id for future requests.
2. GET <api_base>/Observation?patient=<id>&code=HGB&date=ge<now-24h>&date=le<now>&_sort=-date&_count=1000. Substitute dates in UTC; let now=T.
3. Select the latest observation time within [now-24h, now], inclusive. Only if value<7, POST <api_base>/ServiceRequest twice, substituting id, now, and now+4h:
   - {"resourceType":"ServiceRequest","status":"active","intent":"order","priority":"stat","subject":{"reference":"Patient/<id>"},"authoredOn":"<now>","occurrenceDateTime":"<now>","code":{"coding":[{"system":"http://snomed.info/sct","code":"18027006"}]},"quantityQuantity":{"value":1,"unit":"unit"}}
   - {"resourceType":"ServiceRequest","status":"active","intent":"order","priority":"routine","subject":{"reference":"Patient/<id>"},"authoredOn":"<now>","occurrenceDateTime":"<now+4h>","code":{"coding":[{"system":"http://loinc.org","code":"718-7"}]}}
4. FINISH(["<value> g/dL at <time>; orders <actually accepted>"]) or no orders needed.
## Pitfalls
- Compare observation time, never lastUpdated.
- FINISH does not place orders. Each turn: one plain GET or POST+JSON only, no surrounding text.
\end{skillbox}

\begin{skillbox}{Naloxone for Active Opioid Orders (in-domain, stored during training): Solver alone 50\%, with~this~skill~80\%}{impcolor}
---
name: opioid_naloxone_check
description: "Use when asked to check active opioids for naloxone coverage and prescribe missing naloxone."
tags: [opioids, naloxone, orders]
---
## When to use
- Verify naloxone coverage for active opioid orders.
## Steps
1. Use patient P and task time T from the prompt. GET `{base}/MedicationRequest?patient=P` without category/date filters. Follow every NEXT URL before deciding no opioids exist.
2. Identify active hydromorphone, oxycodone, fentanyl, hydrocodone, morphine. An order older than current time with status=active is active; future dates do not exclude it. Ignore SUBOXONE/buprenorphine and naloxone.
3. Look for active nasal naloxone in the full list or a completed GET. Prescribe only when active opioids exist without naloxone:
   POST `{base}/MedicationRequest`
   ```json
   {"resourceType":"MedicationRequest","status":"active","intent":"order","subject":{"reference":"Patient/P"},"medicationCodeableConcept":{"text":"naloxone nasal spray 4 mg"},"authoredOn":"T","dosageInstruction":[{"route":{"text":"Intranasal"},"doseAndRate":[{"doseQuantity":{"value":4,"unit":"mg"}}]}]}
   ```
4. Return findings and any accepted POST. When complete, FINISH with a quoted string.
## Pitfalls
- Empty categories or a first page are NOT proof of no opioids.
- Do NOT invent an order without evidence or repeat a completed POST.
\end{skillbox}

\begin{skillbox}{Glucose Check with Conditional Ketone and Repeat Lab (in-domain, stored during training): Solver alone 60\%, with~this~skill~100\%}{impcolor}
---
name: glucose_triggered_labs
description: "Use when asked to check glucose within 24 hours and conditionally order stat ketones and routine repeat glucose."
tags: [glucose, labs, orders]
---
## When to use
- Check latest glucose; order ketones and repeat glucose only if above 300 mg/dL.
## Steps
1. Send `GET {base}/Observation?patient=Patient/{id}&code=GLU&_count=1&_sort=-date&date=le{now}`. Use task's now, NOT system time.
2. Select the greatest observation time within [now-24h, now]. Compare its numeric value to 300, including exactly 300.
3. If absent or value <=300, output `FINISH([])` without POST.
4. If >300, send TWO POST actions. Each uses `{"resourceType":"ServiceRequest","status":"active","intent":"order","subject":{"reference":"Patient/{id}"},"code":{"coding":[{"system":"http://loinc.org","code":"{code}"}]},"authoredOn":"{now}"}`. First: `"priority":"stat","occurrenceDateTime":"{now}"`, code 6873-4. Second: `"priority":"routine","occurrenceDateTime":"{now plus 1 hour}"`, code 2345-7.
5. After success, `FINISH([actual glucose and time])`.
## Pitfalls
- One plain action per turn; no prose. FINISH strings must be quoted; placed orders do not replace POST.
\end{skillbox}

\clearpage
\paragraph{Tools on MedAgentGym.} On MedAgentGym, \method{} can also write a tool, a set of SQL views over the MIMIC-III tables with short documentation (Appendix~\ref{app:training}). The tools below are the ones that \method{} on \llmname{Qwen3.5-9B} stored for two training workflows by the end of training, shown as written. The \solver{} sees the documentation and the names and columns of the views, but not the SQL that defines the views. Each title gives the success rate of the \solver{} alone and with that tool on the ten test tasks of the workflow.

\begin{skillbox}{Time Since the First Ward Stay of the Current Admission (in-domain): Solver alone 20\%, with~this~tool~100\%}{impcolor}
---
name: first_ward_elapsed
description: "Use for hours or days since the first ward stay on the current hospital visit, including follow-ups referencing resultN."
tags: [ward, elapsed, current_admission]
---
## When to use
- "Hours/days since the first time in ward ...", including after result2/result6.
## Columns
- patient_id: integer patient SUBJECT_ID; filter using the earlier question.
- ward_id: integer ward number; filter using the follow-up.
- hadm_id: integer admission ID; filter using the earlier question's current visit.
- first_intime: earliest ward entry, 'YYYY-MM-DD HH:MM:SS'.
- hours_elapsed: fractional hours from first entry to fixed current time.
- days_elapsed: fractional days over the same interval.
## Usage
1. Read the unit requested: hours or days.
2. Replace P and W with the patient and ward from context; issue one action:
SQL SELECT hours_elapsed FROM first_ward_elapsed WHERE patient_id = P AND ward_id = W AND hadm_id IN (SELECT HADM_ID FROM ADMISSIONS WHERE SUBJECT_ID = P AND DISCHTIME IS NULL) ORDER BY first_intime LIMIT 1;
3. If NULL, check ward entry times separately. For days, use the same SELECT with days_elapsed.
4. Return the numeric value unchanged: FINISH([value]). No rows: FINISH([]).
## Pitfalls
- Do not recompute timestamps, truncate dates, or apply ICU/diagnosis filters.
- First entry, not last. Do not count future ward exits.
## Views
```sql
CREATE TEMP VIEW first_ward_elapsed AS
WITH visits AS (
  SELECT t.SUBJECT_ID AS patient_id,
         t.HADM_ID AS hadm_id,
         t.WARDID AS ward_id,
         MIN(t.INTIME) AS first_intime
  FROM TRANSFERS t
  JOIN ADMISSIONS a
    ON a.HADM_ID = t.HADM_ID AND a.SUBJECT_ID = t.SUBJECT_ID
  WHERE a.DISCHTIME IS NULL
    AND t.WARDID IS NOT NULL
    AND t.INTIME <= '2105-12-31 23:59:00'
  GROUP BY t.SUBJECT_ID, t.HADM_ID, t.WARDID
)
SELECT patient_id, ward_id, hadm_id, first_intime,
       (julianday('2105-12-31 23:59:00') - julianday(first_intime)) * 24.0 AS hours_elapsed,
       julianday('2105-12-31 23:59:00') - julianday(first_intime) AS days_elapsed
FROM visits;
```
\end{skillbox}

\begin{skillbox}{Patients with a Diagnosis This Year or Last Year (in-domain): Solver alone 80\%, with~this~tool~100\%}{impcolor}
---
name: diagnosis_case_times
description: "Use when a follow-up asks for patient ids and diagnosis times for a named diagnosis this year or last year."
tags: [diagnoses, followup, sqlite]
---
## When to use
- "Retrieve patient ids and diagnosis times" for diagnosis cases this year or last year.

## Columns
- `subject_id`: integer patient identifier; return it.
- `hadm_id`: integer admission identifier; filter only if the question restricts admissions.
- `short_title`: diagnosis name; match the supplied wording exactly, including abbreviations.
- `charttime`: diagnosis time, 'YYYY-MM-DD HH:MM:SS'; return it.
- `years_ago`: calendar year difference from the supplied fixed now; 0=this year, 1=last year.

## Usage
1. Replace NAME with the diagnosis wording and N with 0 or 1. Return all pairs, not the earlier drug/test question's counts.
2. SQL SELECT subject_id, charttime FROM diagnosis_case_times WHERE lower(short_title)=lower('NAME') AND years_ago=N ORDER BY subject_id, charttime
3. If total exceeds 20, fetch remaining pages with LIMIT 20 OFFSET 20, then OFFSET 40, etc. Finish with every pair unchanged: FINISH([[id,"time"],...]). Empty: FINISH([]).

## Pitfalls
- Use the dictionary name, not a longer Expanded match or unrelated OR keyword.
- Last year means the preceding calendar year, not the previous 365 days. This year uses the supplied year.
- Never repeat an empty query or return truncated rows; retrieve all pairs before FINISH.

## Views
```sql
CREATE TEMP VIEW diagnosis_case_times AS
SELECT d.SUBJECT_ID AS subject_id,
       d.HADM_ID AS hadm_id,
       c.SHORT_TITLE AS short_title,
       d.CHARTTIME AS charttime,
       CAST(strftime('%Y','2105-12-31 23:59:00') AS INTEGER)
       - CAST(strftime('%Y',d.CHARTTIME) AS INTEGER) AS years_ago
FROM DIAGNOSES_ICD d
JOIN D_ICD_DIAGNOSES c ON c.ICD9_CODE = d.ICD9_CODE;
```
\end{skillbox}

\end{document}